\documentclass{article}

\PassOptionsToPackage{numbers,sort&compress}{natbib}

\usepackage[final]{neurips_2022}

\usepackage[utf8]{inputenc} 
\usepackage[T1]{fontenc}    
\usepackage{hyperref}       
\usepackage{url}            
\usepackage{booktabs}       
\usepackage{amsfonts}       
\usepackage{nicefrac}       
\usepackage{microtype}      
\usepackage{xcolor}         
\usepackage{graphicx}
\usepackage{subcaption}
\usepackage{textcomp, gensymb}
\usepackage{colortbl}
\usepackage{multirow}
\usepackage{booktabs}
\usepackage{mathtools}
\usepackage{algorithm}
\usepackage{algpseudocode}
\definecolor{mygreen}{RGB}{0, 204, 51}
\definecolor{myorange}{RGB}{255, 140, 0}
\usepackage{comment}
\usepackage{makecell}
\usepackage{bbding}

\usepackage{amsmath}
\DeclareMathOperator*{\argmax}{argmax} 

\title{Does Self-supervised Learning Really Improve Reinforcement Learning from Pixels?}

\author{%
  Xiang~Li$^1$~\quad~Jinghuan~Shang$^1$~\quad~Srijan~Das$^{1,2}$~\quad~Michael~S.~Ryoo$^1$\\
  Department of Computer Science\\
  $^1$Stony Brook University, $^2$University of North Carolina at Charlotte\\
  $^1$\texttt{\{xiangli8, jishang, mryoo\}@cs.stonybrook.edu, $^2$sdas24@uncc.edu}\\
}

\begin{document}

\maketitle

\begin{abstract}

We investigate whether self-supervised learning (SSL) can improve online reinforcement learning (RL) from pixels.
We extend the contrastive reinforcement learning framework (e.g., CURL) that jointly optimizes SSL and RL losses and conduct an extensive amount of experiments with various self-supervised losses.
Our observations suggest that the existing SSL framework for RL fails to bring meaningful improvement over the baselines only taking advantage of image augmentation when the same amount of data and augmentation is used.
We further perform evolutionary searches to find the optimal combination of multiple self-supervised losses for RL, but find that even such a loss combination fails to meaningfully outperform the methods that only utilize carefully designed image augmentations.
After evaluating these approaches together in multiple different environments including a real-world robot environment, 
we confirm that no single self-supervised loss or image augmentation method can dominate all environments and that the current framework for joint optimization of SSL and RL is limited.
Finally, we conduct the ablation study on multiple factors and demonstrate the properties of representations learned with different approaches.

\end{abstract}

\section{Introduction}

Learning to act from image observations is crucial in many real-world applications.
One popular approach is online reinforcement learning (RL), which requires no human demonstration or expert trajectories.
Since all training samples are collected by the agent during policy learning in online RL, the collected data often has strong correlations and high variance, challenging the policy learning.
Meanwhile, the cost of interacting with environments requires the RL algorithms to have higher sample efficiency.
Compared to RL using state-based features, pixel-based RL continuously takes images as inputs, which usually come with a much higher dimensionality than numerical states.
Such properties pose serious challenges to image representation learning in RL.

Several recent works studied such challenges from various directions, including:
(1) Inspired by the great success of self-supervised learning (SSL) with images and videos (e.g., \cite{kingma2013auto, doersch2015unsupervised, noroozi2016unsupervised, gidaris2018unsupervised, oord2018representation, chen2020simple, he2020momentum, chen2021exploring, caron2021emerging, caron2020unsupervised, he2021masked, pan2021videomoco, tong2022videomae, kahatapitiya2021self, das2021viewclr, das2021stc, ranasinghe2021self}), some RL methods~\cite{yarats2019improving, laskin2020curl, stooke2021decoupling,   schwarzer2020data, agarwal2021contrastive, lee2020predictive, raileanu2021decoupling, zhu2020masked} take advantage of self-supervised learning. 
This is typically done by applying both self-supervised loss and reinforcement learning loss in one batch. 
In this paper, we dub such joint optimization of the self-supervised loss and the RL loss as the \emph{joint learning framework}.
(2) On the other hand, many papers~\cite{laskin2020reinforcement, yarats2020image, yarats2021mastering, raileanu2021automatic, wang2020improving, pitis2020counterfactual, lin2020invariant, hansen2021stabilizing} investigate how online RL can take advantage of image augmentations.
Among them, RAD~\cite{laskin2020reinforcement} and DrQ~\cite{yarats2020image, yarats2021mastering} show significant improvements by applying relatively simple image augmentations to observations of RL agents.

Our objective is to study how well a single or combination of self-supervised losses and augmentations work under the current \emph{joint learning framework} and to empirically identify their impact on RL systems.
In this paper, we extend such joint (SSL + RL) learning framework, conduct experiments comparing multiple self-supervised losses with augmentations, and empirically evaluate them in many environments from different benchmarks. 
We confirm that a single self-supervised loss under such a joint learning framework typically fails to bring meaningful improvements to existing image augmentation-only methods. 
We also computationally search for a better combination of losses and image augmentations for RL with the joint learning framework. 
The experiments in different environments and tasks show inconsistency in self-supervised learning's capability to improve reinforcement learning. 
Given a sufficient amount of image augmentations, under the current framework, self-supervision failed to show benefits over augmentation-only methods regardless how many self-supervised losses are used. 



With all our findings, we present this work as a thorough reference for investigating better frameworks and losses for SSL + RL and inspiring future research.
Our contributions can be summarized as follows:
\begin{enumerate}
  \item We conduct an extensive comparison of various self-supervised losses under the existing joint learning framework for pixel-based reinforcement learning in many environments from different benchmarks, including one real-world environment. 
  \item We perform evolutionary searches for the optimal combination of multiple self-supervised losses and the magnitudes of image augmentation, and confirm its limitations. 
  \item We conduct the ablation study on multiple factors and demonstrate the properties of representations learned by different methods. 
\end{enumerate}



\section{Preliminaries}
\label{sec:prelim}

\subsection{Reinforcement Learning}
In this paper, we extend the configurations of previous work~\cite{yarats2019improving, laskin2020curl} and exploit SAC~(\textbf{S}oft \textbf{A}ctor \textbf{C}ritic)~\cite{haarnoja2018softa, haarnoja2018softb} and Rainbow DQN~\cite{hessel2018rainbow} for the environments with continuous action space and discrete action space respectively.

\paragraph{Soft Actor Critic} \cite{haarnoja2018softa, haarnoja2018softb} is an off-policy actor-critic algorithm that takes advantage of the maximum entropy to encourage the agent to explore more states during the training. 
It maintains a policy network $\pi_\psi$ and two critic networks $Q_{\phi_1}$ and $Q_{\phi_2}$.
The goal of $\pi_\psi$ is to maximize the expected sum of rewards and a $\gamma$-discounted entropy simultaneously, where the entropy encourages the agent to explore during learning.

\paragraph{Rainbow DQN} \cite{hessel2018rainbow} is a variant of DQN~\cite{mnih2015human} with a bag of improvements such as double Q-learning~\cite{hasselt2010double, van2016deep}, prioritized sampling~\cite{schaul2015prioritized}, noisy net~\cite{fortunato2017noisy}, distributional RL~\cite{bellemare2017distributional}, dueling networks~\cite{wang2016dueling} and multi-step reward.


\subsection{Pairwise Learning}
We coin the term ``pairwise'' learning for the frameworks that learn visual representations based on semantic invariance between dual-stream encoder representations.
A general pairwise learning method first generates multiple augmented views by applying a series of random image augmentations to the input sample, then clusters views with the same semantics in the representation space.
Optionally in such frameworks, methods using contrastive losses repel samples with different semantics.
In this paper, we focus on four representative pairwise learning methods, MoCo~\cite{he2020momentum, chen2020improved, mocov3}, BYOL~\cite{grill2020bootstrap}, SimSiam~\cite{chen2021exploring} and DINO~\cite{caron2021emerging}.
We have a detailed explanation and comparison of these methods in Appendix~\ref{sec:pairwise}.

\subsection{Representation Learning for Pixel-based RL}
Previous works explore the possibility of learning better visual representation which may finally benefit policy learning.
One direction is using image augmentation for policy learning~\cite{laskin2020reinforcement, yarats2020image, yarats2021mastering, raileanu2021automatic, wang2020improving, pitis2020counterfactual, lin2020invariant, hansen2021stabilizing}, where RAD~\cite{laskin2020reinforcement} and DrQ~\cite{yarats2020image, yarats2021mastering} achieve significant performance using simple image augmentation.
Another direction is to combine SSL with RL~\cite{yarats2019improving, laskin2020curl, stooke2021decoupling,   schwarzer2020data, agarwal2021contrastive, lee2020predictive, raileanu2021decoupling, zhu2020masked}, in which there are two representative methods, SAC+AE~\cite{yarats2019improving} and CURL~\cite{laskin2020curl}.

\paragraph{RAD} (\textbf{R}einforcement Learning with \textbf{A}ugmented \textbf{D}ata)~\cite{laskin2020reinforcement} investigates the impact of different types of image augmentations for both image and state inputs. By applying random translation or random crop to the input image, RAD significantly improves data efficiency solely through image augmentation without any auxiliary losses.

\paragraph{DrQ} (\textbf{D}ata-\textbf{r}egularized \textbf{Q})~\cite{yarats2020image} further investigates the possibilities of utilizing image augmentation. 
DrQ applies image augmentation twice on the input images and averages the Q value over two augmented images which is assigned as the Q value of the input images.
DrQ~v2~\cite{yarats2021mastering}, which is the successor of DrQ, switches to DDPG (Deep Deterministic Policy Gradient)~\cite{lillicrap2015continuous} as the RL method, brings scheduled exploration noise to control the levels of exploration at different learning stages, and introduces faster implementations of the image augmentation and the replay buffer.

\paragraph{SAC+AE}~\cite{yarats2019improving} takes advantage of a RAE (deterministic \textbf{R}egularized \textbf{A}uto\textbf{E}ncoder)~\cite{ghosh2019variational}, in replacement of $\beta-$VAE~\cite{higgins2016beta} to improve learning stability.
The RAE is jointly trained with SAC by performing both SAC update and RAE update alternatingly in one batch.

\paragraph{CURL} (\textbf{C}ontrastive \textbf{U}nsupervised Representations for \textbf{R}einforcement \textbf{L}earning)~\cite{laskin2020curl} combines contrastive learning with an online RL algorithm by introducing an additional contrastive learning head at the end of the image encoder.
Similar to the aforementioned SAC+AE, here the contrastive loss and reinforcement learning loss are applied alternatively at training.

\section{Self-supervision for Reinforcement Learning}
\label{sec:method}
To effectively evaluate different self-supervised losses, we extend the well-known \emph{joint learning framework} widely used in previous papers~\cite{yarats2019improving, laskin2020curl,  schwarzer2020data, agarwal2021contrastive, lee2020predictive} by adding a general self-supervised learning head to the RL framework. 
We keep the same RL method in CURL~\cite{laskin2020curl}: we use SAC~\cite{haarnoja2018softb} in tasks with continuous action space and use Rainbow DQN~\cite{hessel2018rainbow} in tasks with discrete action space.
\subsection{General Joint Learning Framework}
\label{sec:general}
\begin{figure}
  \centering
  \begin{subfigure}{0.43\linewidth}
    \includegraphics[width=1\linewidth]{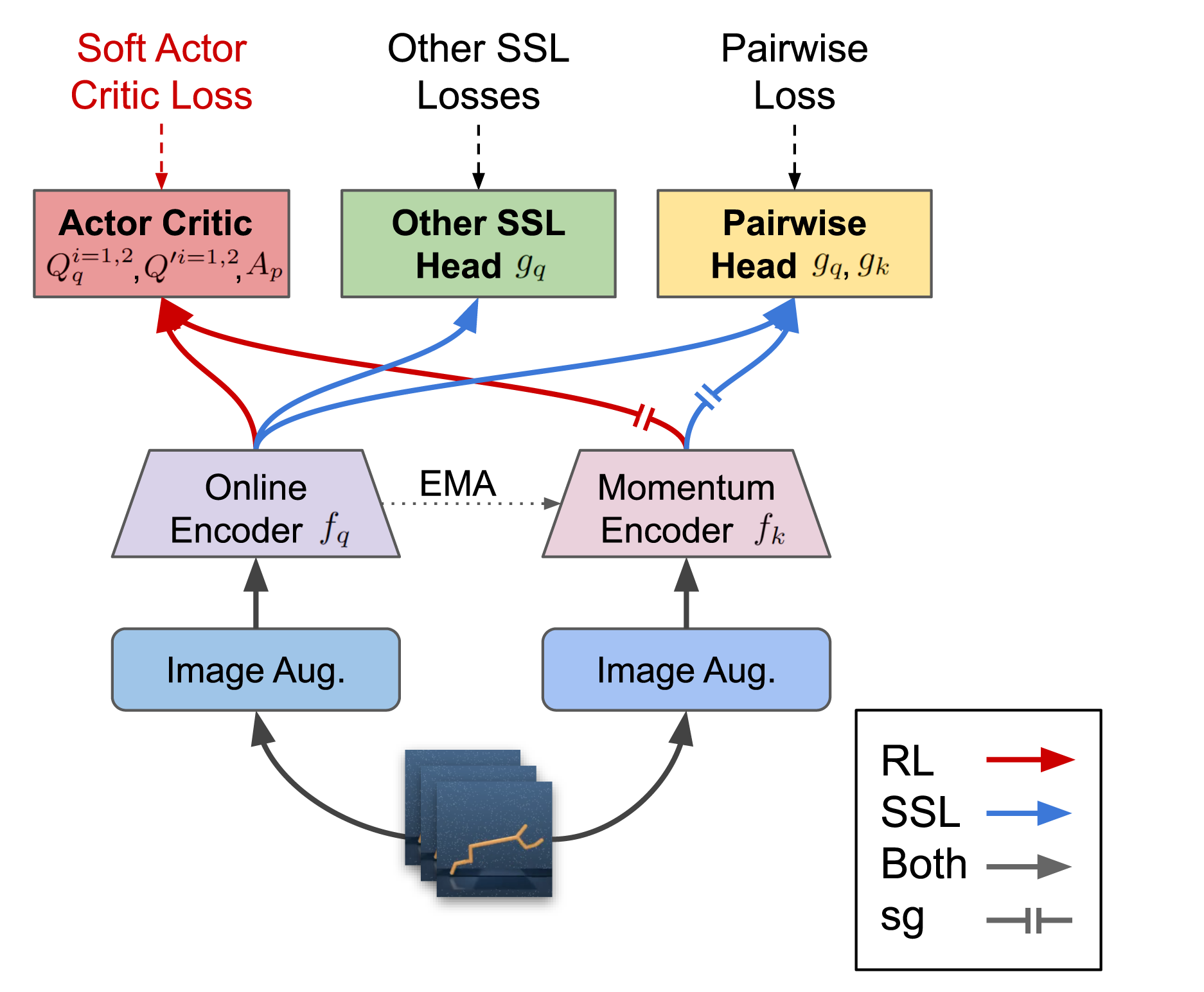}
    \caption{With SAC as the RL method}
    \label{fig:sac}
  \end{subfigure}
  \hspace{1cm}
  \begin{subfigure}{0.43\linewidth}
    \includegraphics[width=1\linewidth]{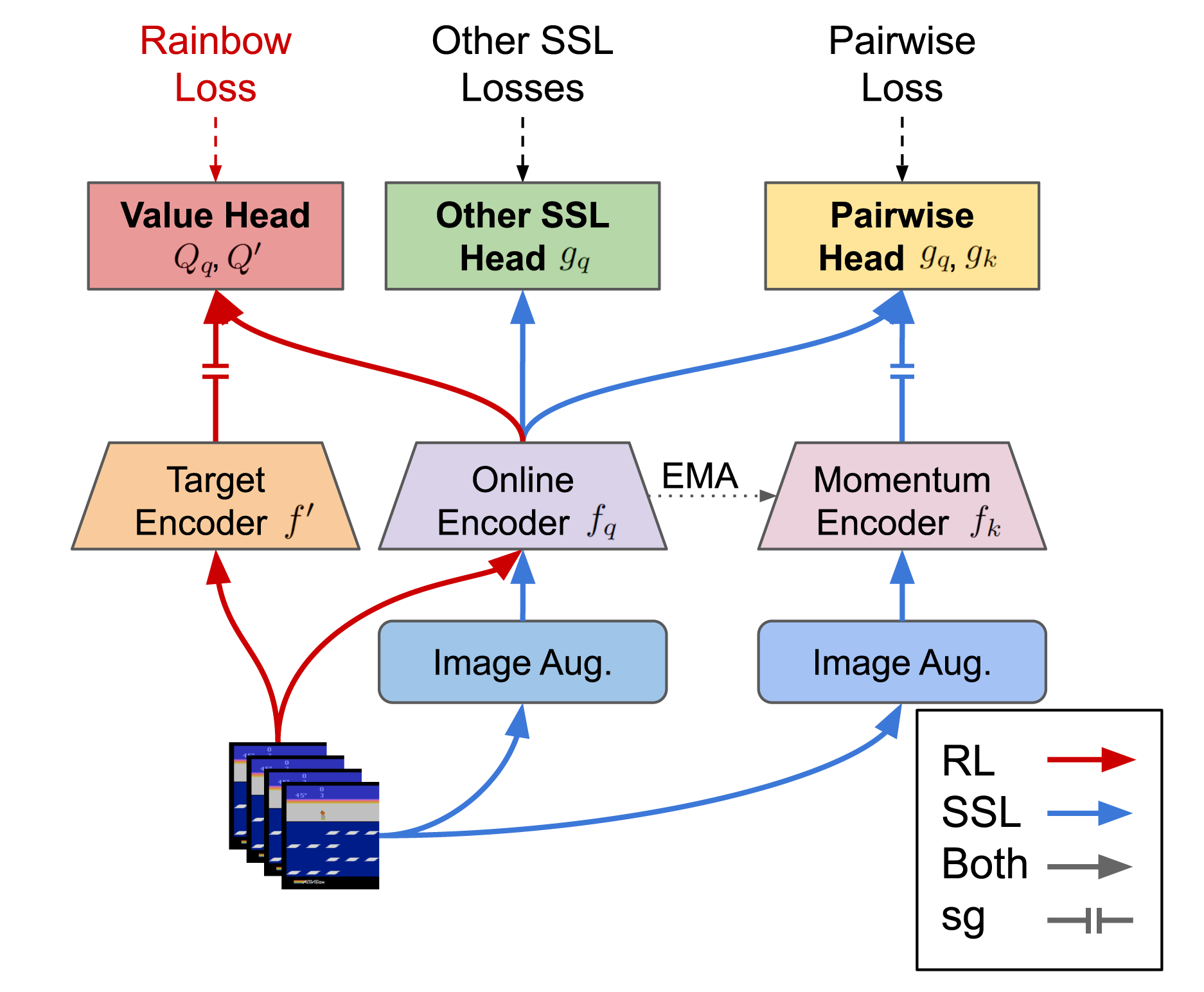}
    \caption{With Rainbow as the RL method}
    \label{fig:rainbow}
    \end{subfigure}
  \caption{General joint learning framework for SSL + RL. The red solid arrow represents the RL flow; the blue one represents the SSL flow and the black one means a shared flow for both; sg stands for stop gradients.}
\end{figure}

\paragraph{With SAC} Fig.~\ref{fig:sac} shows a general joint learning framework, using SAC as the RL method.
The unmodified SAC contains an online encoder~$f_q$, a target (or momentum) encoder~$f_k$, and an actor head~$A_p$.
Each encoder is also followed by two critic heads.
Besides that, we attach an additional self-supervised head $g_q$ after the online encoder.
For pairwise learning losses, we concatenate a momentum SSL head $g_k$ after the target encoder when needed.

For every sampled batch of transitions, we first apply image augmentation to both the current state $s$ and the next state $s'$ and update the SAC model ($f_q, Q_q^{i=1,2}, A_p$) using the augmented images.
Note that for stability concerns, we do not update the parameters of the image encoder when updating the actor head $A_p$.
Then, the target networks are updated by Exponential Moving Average (EMA).
This is followed by also performing an EMA update of the SSL head if required.
Finally, the online encoder $f_q$ and the self-supervised head $g_q$ are updated by the self-supervised loss. By alternatingly performing RL and SSL in every batch, we jointly train all the components in the framework.
The pseudo-code of SAC update alternating RL and SSL is provided in Algorithm~\ref{alg:sacshort}.
\begin{algorithm}
    \centering
    \caption{Update SAC with Self-supervised Losses\\
    \textcolor{mygreen}{Green: additional operations for SSL}; 
    \textcolor{orange}{Orange: only for BYOL and DINO}.
    }\label{alg:sacshort}
    \footnotesize
    \begin{algorithmic}
        \Procedure{UpdateSACwithSSL}{$s$: current state, $s'$: next state, $a$: action, $r$: reward, $d$: done signal, step: model update step counter, $f_q$: online encoder, $f_k$:  target/momentum encoder, $A_p$: actor head, $Q_q^i$: online critic head, $Q'^i$:  target critic head, $\tau:$ target/momentum network update rate, \textcolor{mygreen}{$g_q$: online SSL head}, \textcolor{orange}{$g_k$: momentum SSL head}}
        \State {\textcolor{mygreen}{$s_a, s_a'\gets \textsc{ImageAugmentation}(s),        \textsc{ImageAugmentation}(s')$}}
        \State {\textcolor{mygreen}{$s_p, s_p'\gets \textsc{ImageAugmentation}(s),        \textsc{ImageAugmentation}(s')$}}
        \State {$f_q, Q_q^{i=1, 2}, A_p \gets \textsc{UpdateSoftActorCritic}(\textcolor{mygreen}{s_a}, \textcolor{mygreen}{s_a'}, a, r, d)$}
        \State {$f_k, Q'^{i=1, 2} \gets \tau (f_q, Q_q^{i=1, 2}) + (1 - \tau) (f_k, Q'^{i=1, 2}$)}\Comment{EMA update of SAC}
        \State {\textcolor{orange}{$g_k \gets \tau g_q + (1 - \tau)g_k$}}\Comment{\textcolor{orange}{EMA update of the momentum SSL head}}
        \State {\textcolor{mygreen}{$f_q, g_q \gets \textsc{UpdateSSL}(s_a, s_a', s_p, s_p', a, r)$}}
        \EndProcedure
    \end{algorithmic}
\end{algorithm}

\paragraph{With Rainbow DQN} 
Fig.~\ref{fig:rainbow} demonstrates how to jointly apply SSL to Rainbow DQN.
The unmodified Rainbow DQN maintains an online encoder~$f_q$ and a target encoder~$f'$, followed by two state value heads $Q_q$ and $Q'$.
We introduce an additional momentum encoder $f_k$ and self-supervised heads $g_q$ and $g_k$ as suggested in CURL.
For each batch, the self-supervised losses are computed using augmented images, while the RL loss is computed using the original data.
Finally, the online encoder $f_q$ and the self-supervised head $g_q$  are updated by the self-supervised loss. 
The pseudo-code of Rainbow DQN update can be found at Algorithm~\ref{alg:rainbowshort}.

\begin{algorithm}
    \centering
    \caption{Update Rainbow with Self-supervised Losses\\
    \textcolor{mygreen}{Green: additional operations for SSL}; 
    \textcolor{orange}{Orange: only for BYOL and DINO}.
    }\label{alg:rainbowshort}
    \footnotesize
    \begin{algorithmic}
        \Procedure{UpdateRainbowDQNwithSSL}{$s$: current state, $s'$: next state, $a$: action, $r$: reward, $d$: done, step: model update step counter, $f_q$: online encoder, $f'$: target encoder, $Q_q$: online value head, $Q'$:  target value head, \textcolor{orange}{$f_k$: momentum networks}, \textcolor{orange}{$\tau:$ momentum network update rate},
        \textcolor{mygreen}{$g_q$: online SSL head}, \textcolor{orange}{$g_k$: momentum SSL head}, \textcolor{mygreen}{$w_{SSL}$: weights of self-supervised losses}}
            \State {\textcolor{mygreen}{$s_a, s_a'\gets \textsc{ImageAugmentation}(s), \textsc{ImageAugmentation}(s')$}}
            \State {\textcolor{mygreen}{$s_p, s_p'\gets \textsc{ImageAugmentation}(s),        \textsc{ImageAugmentation}(s')$}}
            \State {\textcolor{mygreen}{$\mathcal{L}_{SSL} \gets \textsc{CalculateSSLoss}(s_a, s_a', s_p, s_p', a, r)$}}
            \State {$\mathcal{L}_{\text{Rainbow}} \gets \textsc{CalculateRainbowLoss}(s, s', a, r, d)$}
            \State {$\mathcal{L} \gets \mathcal{L}_{\text{Rainbow}}$ \textcolor{mygreen}{+ $w_{SSL} \mathcal{L}_{SSL}$}}
            \State {$f_q, Q_q, \textcolor{mygreen}{g_q} \gets $ \textsc{OnlineNetworksUpdate($\mathcal{L}$)}}
            \State {$f', Q' \gets f_q, Q_q$} \Comment{Copy parameters from online networks to target networks}
            \State {\textcolor{orange}{$f_k, g_k \gets \tau (f_q, g_q) + (1 - \tau)(f_k, g_k)$}}\Comment{\textcolor{orange}{EMA update of momentum networks and SSL head}}
        \EndProcedure
    \end{algorithmic}
\end{algorithm}

\subsection{Losses for Self-supervised Learning}
\label{sec:sslopt}
The self-supervised losses we investigated can be categorized into four classes: pairwise learning, transformation awareness, reconstruction, and reinforcement learning context prediction.

\paragraph{Pairwise Learning}
We investigate three representative pairwise learning methods: BYOL~\cite{grill2020bootstrap}, DINO~\cite{caron2021emerging} and SimSiam~\cite{chen2021exploring}, along with existing CURL whose framework is similar to MoCo~\cite{he2020momentum}. 
BYOL, DINO, and SimSiam only explicitly pull positive samples closer without the need for a large number of negative samples. 
CURL uses a contrastive loss taking both positive and negative samples into consideration.

Given the general joint learning framework described in Sec.~\ref{sec:general}, by substituting the self-supervised head and loss, we can easily formulate different agents w.r.t. self-supervised losses.
For BYOL, as shown in Fig.~\ref{fig:byol}, a projector and a predictor are appended to the online encoder sequentially while a momentum projector is attached on top of the target/momentum encoder.
DINO (Fig.~\ref{fig:dino}) maintains only projector in both online and target branches.
Similar to BYOL, the momentum projector in DINO is also updated by EMA.
The two encoders in BYOL and DINO operate on two augmented views of the data respectively whereas SimSiam (see Fig.~\ref{fig:simsiam}), uses only the online network and a projector for processing both the augmented views.

We also test two methods that introduce RL-specific variables to this pairwise learning framework, \emph{CURL-w-Action} and \emph{CURL-w-Critic}.
\emph{CURL-w-Action} is based on CURL while the contrastive loss is applied to the concatenation of image representation and output of the actor network, instead of the image representation only.
Similarly, \emph{CURL-w-Critic} concatenates the critic network output with the existing image representation for contrastive loss.

\paragraph{Transformation Awareness}
Recent works (e.g., \cite{noroozi2016unsupervised, kim2019self, gidaris2018unsupervised, jing2018self, dangovski2021equivariant, lee2021improving}) have shown that the awareness of transformations (like rotation, Jigsaw puzzle, and temporal ordering) improves many downstream tasks in computer vision like image classification and action recognition.
Typically such awareness can be acquired by explicitly asking a classifier to identify the applied transformation from the pixel representation.
Therefore, we investigate two simple classification losses, rotation classification (\emph{RotationCLS}) and shuffle classification (\emph{ShuffleCLS}), and set a two-layer MLP classifier as the self-supervised head in the joint learning framework.

\emph{RotationCLS} represents the methods that encourage spatial transformation awareness.
Inspired by RotNet~\cite{gidaris2018unsupervised} and E-SSL~\cite{dangovski2021equivariant}, we rotate the input image after augmentation by 0\textdegree, 90\textdegree, 180\textdegree and 270\textdegree. 
The classifier predicts the rotation angle from the visual representation and it is trained by cross-entropy loss.

\begin{figure}
    \centering
    \begin{minipage}{0.33\textwidth}
        \includegraphics[width=1\linewidth]{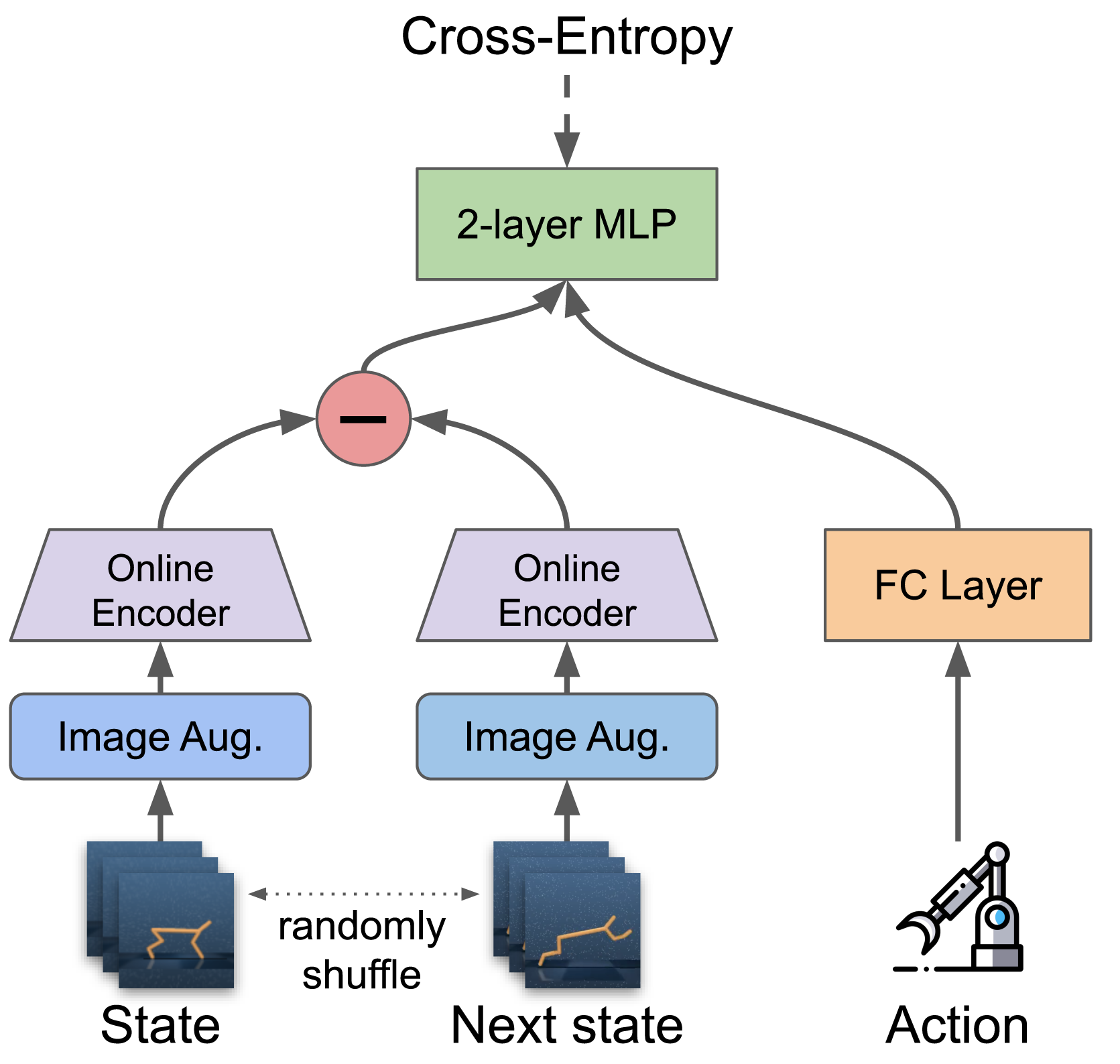}
        \caption{ShuffleCLS}
        \label{fig:shufflecls}
        \end{minipage}
    \hspace{1cm}
    \begin{minipage}{0.45\textwidth}
        \includegraphics[width=1\linewidth]{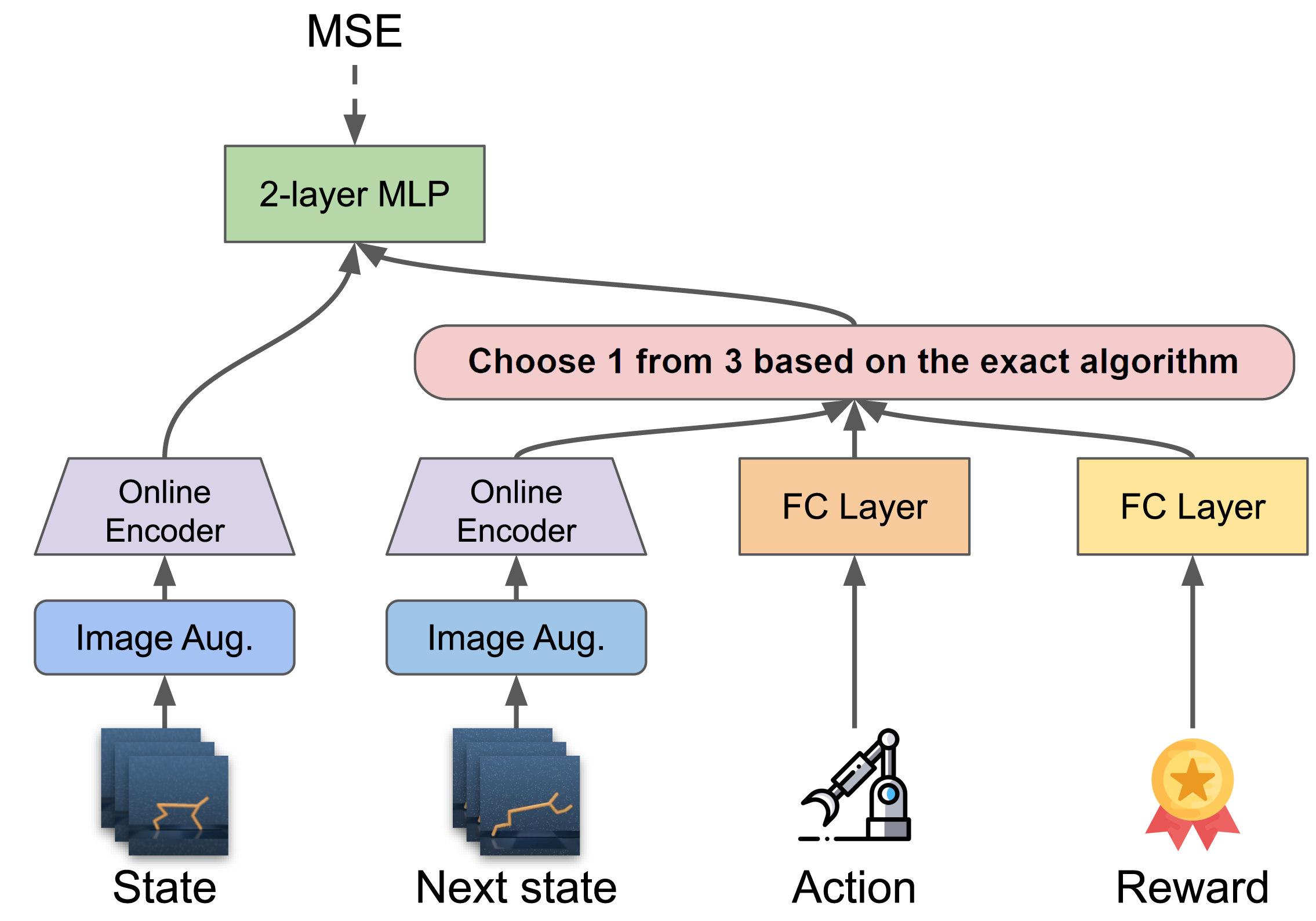}
        \caption{General RL context prediction}
        \label{fig:regression}
    \end{minipage}
\end{figure}

Shuffle~Tuple~\cite{misra2016shuffle} encourages the encoder to develop an awareness of action causality by predicting if two frames appear in order.
We adapt Shuffle~Tuple by randomly shuffling the current state image and next state image in a state transition tuple and predicting whether it is shuffled or not.
The classifier also takes action into consideration because some of the transitions are reversible.
The overall architecture of \emph{ShuffleCLS} is shown in Fig.~\ref{fig:shufflecls}.

\paragraph{Reconstruction}
Reconstructing the input image with an hourglass architecture has been shown to be an effective way to learn image representation~\cite{kingma2013auto, higgins2016beta, ghosh2019variational}.
We simply extend SAC+AE by changing the input and reconstruction target to be augmented images.
The reconstruction loss and regularization from RAE~\cite{ghosh2019variational} are left untouched.

Recent study on \textbf{M}asked \textbf{A}uto\textbf{E}ncoder~\cite{he2021masked} (MAE) adapts the reconstruction task for patch-based Vision Transformers~\cite{dosovitskiy2020image}.
The objective in MAE includes reconstructing the entire image from input masked image patches.
Inspired by this, we adapt SAC+AE into SAC+MAE by replacing the augmented input image with its masked version and only penalizing the reconstruction error for the masked patches.


\paragraph{RL Context Prediction}
Besides the self-supervised learning methods that are specifically designed for pixels, we investigate the losses using attributes naturally collected during the RL process.
For any state transition that is not the end of a trajectory, it contains four components: current state $s$, next state $s'$, action $a$ and reward $r$, with the trajectory termination signal omitted.
Inspired by \citet{shelhamer2016loss}, we concatenate the visual representation of the current state $s$ and another representation $h$ as the input.
Without loss of generality, the second input representation $h$ can be any of these three representations of $s'$, $a$, and $r$.
Then, we predict the remaining components using a two-layer MLP.
For continuous outputs, mean-squared error (MSE) loss is applied, while for the discrete target (e.g., action in discrete action space), we use cross-entropy loss.
The architecture of this group of self-supervised losses is shown in Fig.~\ref{fig:regression}.
From the combination of inputs and outputs, we define nine losses whose I/O specifications are provided in Table~\ref{tab:regression}.
For those losses whose outputs include two components, two target prediction networks share the same SSL head except the last task-specific layer.

\begin{table}[t]
    \caption{I/O of RL context prediction losses}
    \label{tab:regression}
    \centering
\resizebox{\textwidth}{!}{
  \begin{tabular}{@{}cccccccccc@{}}
    \toprule
    & Extract-A & Extract-R & Guess-A & Guess-F & Predict-F & Predict-R & Extract-AR & Guess-AF & Predict-FR\\
    \midrule
   Rep. of $s'$ & Input&Input&-&Output&Output&-&Input&Output&Output\\
    Action $a$&Output&-&Output&-&Input&Input&Output&Output&Input\\
    Reward $r$&-&Output&Input&Input&-&Output&Output&Input&Output\\
    \bottomrule
  \end{tabular}
 }
\end{table}

\subsection{Evolving Multiple Self-supervised Losses}
\label{sec:elo}
Besides a single self-supervised loss or handcrafted combination of two losses, we further investigate how multiple self-supervised losses affect the policy learning together with the joint learning framework.
In such a configuration, the agent maintains multiple SSL heads at the same time and we apply losses to their corresponding head individually.
We formulate the combination of multiple losses as a weighted sum $\mathcal{L}_{\text{Combo}} = \sum_{i=1}^{N_l}{w_i \cdot \mathcal{L}_i}$ where $w_i$ is the weight of a specific loss $\mathcal{L}_i$ and $N_l$ is the total number of losses in the search space.
In the joint learning framework, we apply both self-supervised $\mathcal{L}_{\text{Combo}}$ and RL losses jointly to the networks for every mini-batch.
Considering that the policy learning is quite sensitive to hyper-parameters, it is non-trivial to find each weight for every SSL loss. 

ELo~(\textbf{E}volving \textbf{Lo}sses)~\cite{piergiovanni2019evolving} shows promising results in unsupervised video representation learning~\cite{pan2021videomoco, tong2022videomae}, by using evolutionary search to automatically find optimal combination of many self-supervised losses.
In the spirit of ELo, we turn to evolutionary search to automatically find the optimal solution.
Assume an unknown objective function whose inputs are weights of multiple losses $w_i$ and the magnitudes of image augmentation $m_{j=1,2}$ for the online encoder and momentum encoder. 
The function output is the score achieved by the trained agent in its environment with a certain random seed: $\mathcal{R_{\text{env}}^{\text{seed}}}(m_{j=1, 2}, w_{i=1, 2, \dots N_l})$.
Essentially, the objective function maps a set of $w_i$ and $m_j$ to the reward achieved by a corresponding agent, and $w_i$ and $m_j$ stay unchanged during the agent learning process.
The optimization algorithm approaches the maximum value of the objective function by repeatedly updating $w_i$ and $m_j$ and testing the value of the objective function, which in our case is the training and evaluation of an agent with the given parameters (i.e., the input of the objective function).
We choose an off-the-shelf optimization algorithm PSO~(\textbf{P}article \textbf{S}warm \textbf{O}ptimization)~\cite{kennedy1995particle} for its simplicity.
For each set of inputs, we find it critical to run with multiple random seeds and report IQM (interquartile mean)\footnote{Mean using only the data between the first and third quartiles~\cite{wikipedia_2022}} for a stable and robust search. 
The optimization process is presented as:  
\begin{equation}
\argmax_{m_{j=1, 2}, w_{i=1, \dots, N_l}} \text{IQM}(\mathcal{R_{\text{env}}^{\text{seed=1,\dots,5}}}(m_{j=1, 2}, w_{i=1, \dots, N_l}))
\end{equation}
Note that we are also implicitly searching for the balance between the self-supervised loss and the RL loss by performing this search, as it has the capability to adjust the absolute weights of the self-supervised losses overall.
We search on DMControl~\cite{tassa2018deepmind} with SAC using three different configurations named ELo-SAC, ELo-SACv2 and ELo-SACv3 respectively.
ELo-Rainbow performs a search on Atari with Efficient Rainbow.
Please refer to Sec.~\ref{sec:implementelo} for our detailed configurations and search results.
\section{Experiments}
\label{sec:exp}
We conduct experiments in three major directions, in order to better understand how we should integrate SSL with RL. 
First, we demonstrate how different self-supervised losses affect the RL process, by trying them on multiple challenging tasks. 
Then, we dive into detailed ablations on multiple factors, and finally, we perform empirical analysis on the visual representations learned with the joint learning framework (Sec.~\ref{sec:analysis}).
In addition, we benchmark a pretraining framework as an alternative to the joint learning framework (Sec.~\ref{sec:pretraining}).

\paragraph{Evaluation Scheme}
Thorough evaluation of reinforcement learning algorithms is challenging due to the high variances between each run and the extensive requirement of computation.
Consequently, we run all experiments with multiple different random seeds and report the interquartile mean and the standard deviation of the scores as suggested by \citet{agarwal2021deep}. 
For a quantitative comparison of the different methods mentioned in Section~\ref{sec:sslopt}, in addition to the absolute scores, we assign a \emph{Relative Score} to each method.
We denote the interquartile mean of scores achieved by agent $A$ in environment $e \in E$ as $\text{IQM}^{A,e}$ and denote the collection of all interquartile mean scores achieved in environment $e$ by different agents as $\text{IQM}^e$.
The Relative Score of agent $A$ is computed as  $S_\text{Relative}^A = \sum_{e \in E}{(\text{IQM}^{A,e} - \text{mean}(\text{IQM}^e)) / \text{std}(\text{IQM}^e)}$.

\paragraph{DMControl Experiments}
DMControl (\textbf{D}eep\textbf{M}ind \textbf{C}ontrol suite)~\cite{tassa2018deepmind} contains many challenging visual continuous control tasks, which are widely utilized by recent papers.
We evaluate all the methods introduced in Sec.~\ref{sec:method}, along with two important baselines, SAC-NoAug and SAC-Aug(100), in six environments of DMControl that are commonly used in previous papers \cite{yarats2019improving, laskin2020curl, laskin2020reinforcement, yarats2020image}.
Other methods that only take advantage of image augmentation, like RAD~\cite{laskin2020reinforcement} and DrQ~\cite{yarats2020image} are also benchmarked for comparison.
In the case of SAC+AE~\cite{yarats2019improving}, we provide the augmented images for a fair comparison, which is a different configuration to the original paper.
Please refer to Appendix~\ref{sec:dataaugmentationdiff} for a detailed comparison of method variants and the exact data augmentation they applied.

\begin{figure}
\centering
\includegraphics[width=\linewidth]{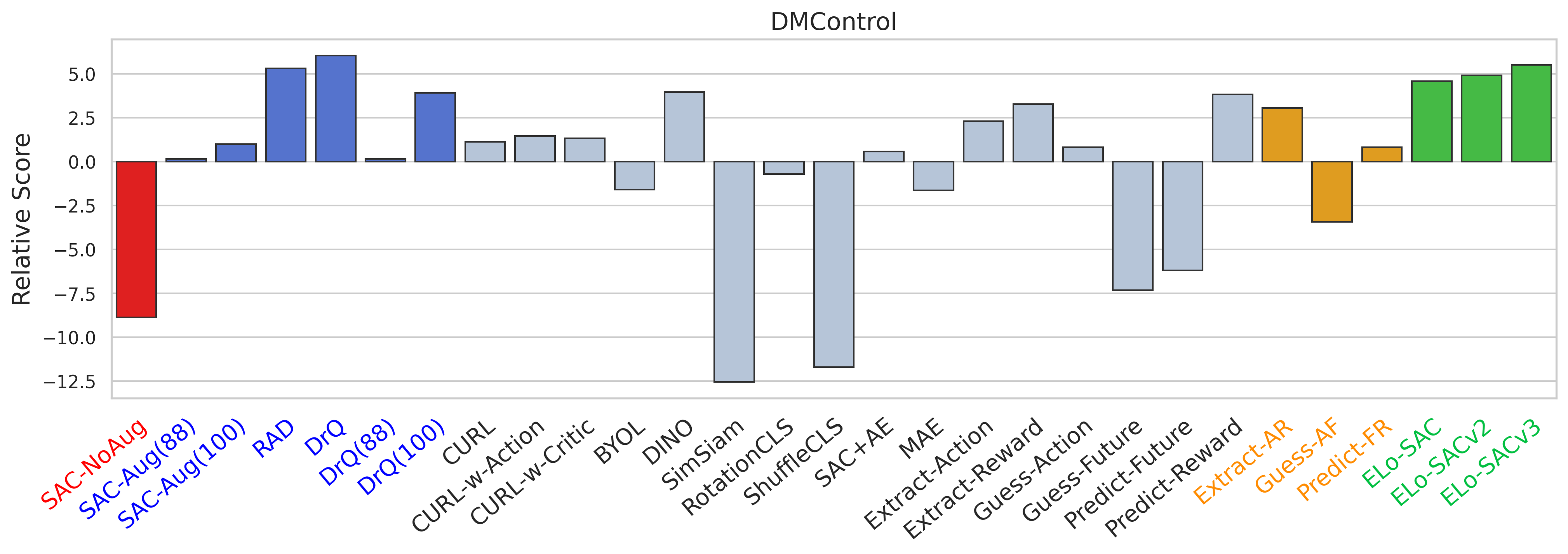}
\caption{Relative Scores on six DMControl tasks, environment step=100k, batch size=512, Number of seeds=10. \textcolor{red}{SAC-NoAug} uses no image augmentation, while all the other methods benefit from image augmentation; The methods in blue (like \textcolor{blue}{DrQ}) only take advantage of image augmentation without any SSL; the methods in black (like CURL) apply one self-supervised loss; the methods in orange (like \textcolor{myorange}{Extract-AR}) manually combines two self-supervised losses; \textcolor{mygreen}{ELo-SAC}, \textcolor{mygreen}{ELo-SACv2} and \textcolor{mygreen}{ELo-SACv3} combine multiple self-supervised losses with specific weights from an evolutionary search. From this figure, \textbf{No} existing SSL-based method with the joint learning framework achieves better performance than DrQ which only use well-designed image augmentation. ELo-SAC methods achieve higher Relative Scores than all the self-supervised methods, but it still performs worse than DrQ and RAD, with an exception of \textcolor{mygreen}{ELo-SACv3} which is marginally better than RAD. 
}
\label{fig:dmcbox}
\end{figure}

We mainly follow the hyper-parameters and the test environments reported in CURL, except that we use the same learning rate $10^{-3}$ in all environments for simplicity.
All the methods are benchmarked at 100k environment steps, with training batch size 512 under 10 random seeds, and they share the same capacity of policy network.
The relative score of each tested algorithm on DMControl is reported as Fig.~\ref{fig:dmcbox}.
We also strongly encourage readers to check full results at Table~\ref{tab:DMC} and results in two additional harder environments at Table~\ref{tab:harder} for a full picture.

From the first glance at Fig.~\ref{fig:dmcbox}, \text{no} tested SSL-based method within the joint learning framework achieves better performance than DrQ and RAD which are carefully designed to take the best advantage of specific image augmentations.
Compared to the baseline SAC-Aug(100), approaches with a self-supervised loss frequently (11 out of 19) fail to improve reinforcement learning.
Some SSL methods (like SimSiam, ShuffleCLS) ruin the policy learning resulting in performance even worse than SAC-NoAug, which suggests that improper use of self-supervised loss can damage the benefits brought by image augmentation.
Then, regarding combining losses, Guess-AF and Predict-FR, which are manually designed to combine two individual losses, are not better than the single self-supervised loss in their combinations (check Guess-Action and Predict-Reward in Fig.~\ref{fig:dmcbox}).

ELo-SAC and ELo-SACv2 find the desired combination by searching in one task.
Such combination generalizes to other environments on DMControl with better overall performance than any approach in the search space. 
In the `cheetah run' where the search was performed, they obtained the best result among the approaches with SSL (Table~\ref{tab:harder}). 
This demonstrates the feasibility of ELo-SAC and implies that the obtained combination through evolutionary search has the potential to generalize to other environments in DMControl.
However, weaker performance in `finger, spin' and `reacher, easy' made ELo-SAC relatively worse than DrQ (which does not use any self-supervision) on average.
Interestingly, there is a similar performance pattern between ELo-SAC and ELo-SACv2 though they have different search spaces.
By contrast, ELo-SACv3 finds an overall better combination by searching in six environments simultaneously. 
Though it achieves highest score in `walker, walk' and `reacher, easy', it performs worse in `cartpole, swingup' and `cheetah, run' than ELo-SAC and ELo-SACv2.
Such observations could be a clue to the properties of different tasks and self-supervised methods.

\begin{figure}
    \centering
    \begin{minipage}{0.43\textwidth}
        \includegraphics[width=1\linewidth]{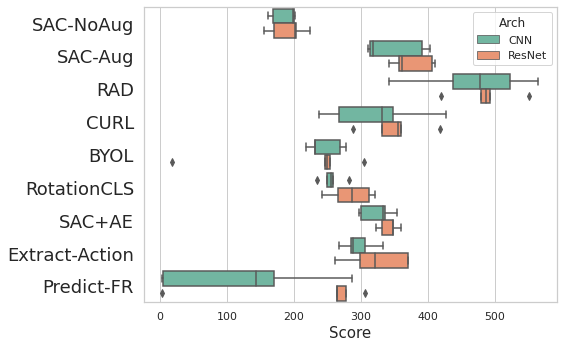}
        \caption{Ablation on encoder backbone}
        \label{fig:ablationcnn}
        \end{minipage}
    \hfill
    \begin{minipage}{0.55\textwidth}
        \includegraphics[width=1\linewidth]{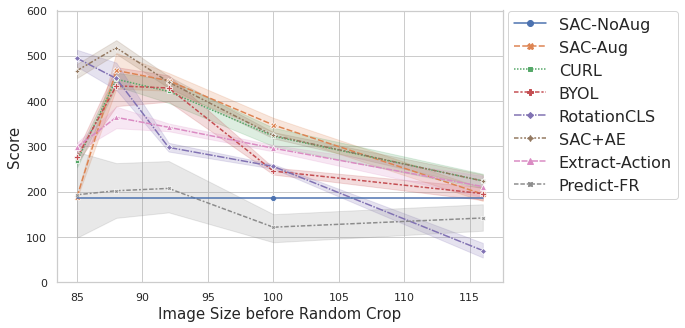}
        \caption{Ablation on random crop augmentation}
        \label{fig:ablationrandomcrop}
    \end{minipage}
\end{figure}

\paragraph{Ablations}
Our observations with SAC-Aug(88), SAC-Aug(100), and RAD suggest the importance of augmentation hyper-parmeters, given the only difference between these three methods is the augmentation applied.
We conduct an ablation study on the image augmentation random crop~\cite{laskin2020curl} in cheetah run, DMControl. 
All the hyper-parameters are as noted in Table~\ref{tab:hydmc} except that the environment step is set to $400k$ and the batch size is reduced to $128$.
Fig.~\ref{fig:ablationrandomcrop} shows how the magnitudes of random crop and translate contribute to the score that the agent achieved.
The image size before the random crop is linear to the magnitude of the random crop when using a fixed crop size: the larger the image size, the stronger the augmentation.
There is a trend that the score first increases and then decreases as the image augmentation gets stronger.
In summary, it is critical to engineering image augmentation carefully when designing an RL system with or without SSL. 

Then we investigate a different visual encoder backbone ResNet~\cite{he2016deep} by replacing the last two convolutional layers with a residual block that has the same number of layers and channels as the CNN baseline.
The ResNet backbone slightly improves all these methods (see Fig.~\ref{fig:ablationcnn}).
We also encourage the readers to check more ablations regarding image augmentation (e.g. random translate), learning rate, encoder layers, and activation function in Appendix~\ref{sec:ablation}.

\paragraph{Atari Game Experiments}
\begin{figure}[t]
\centering
\includegraphics[width=0.85\linewidth]{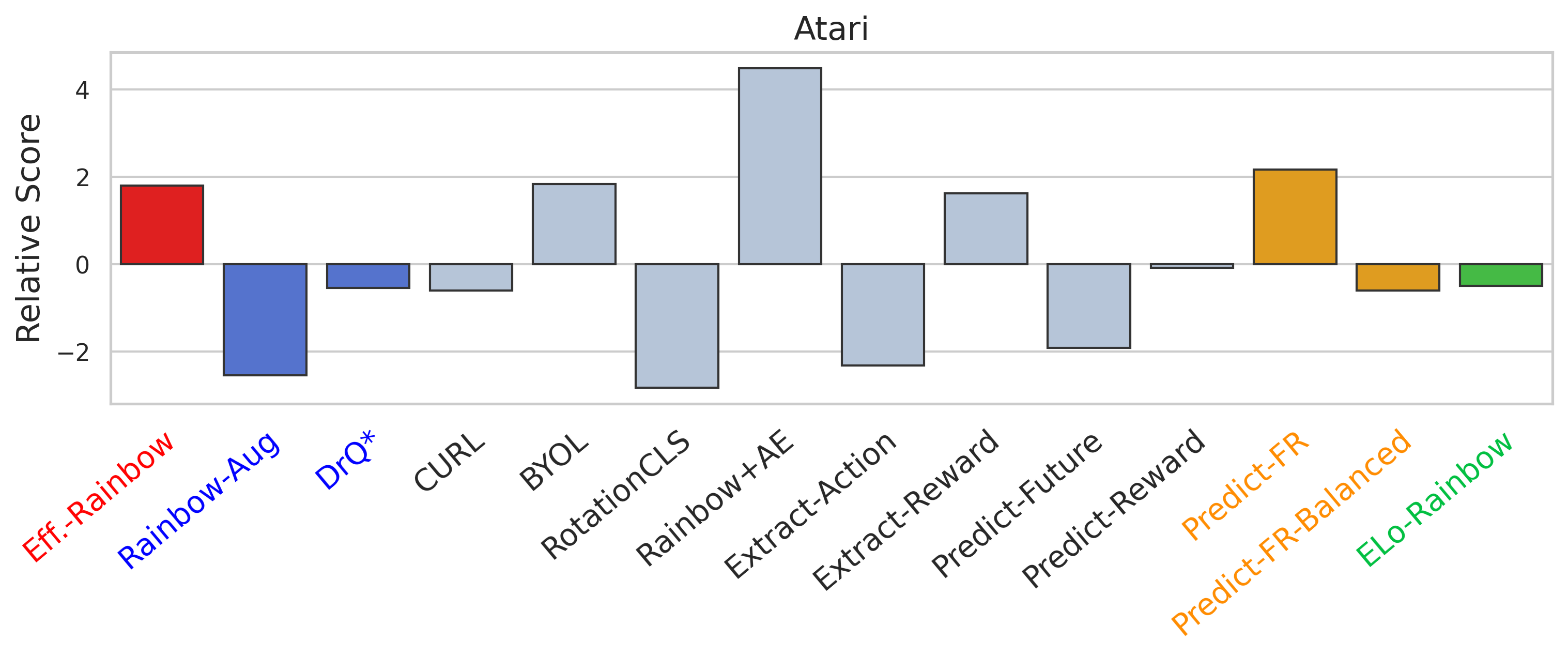}
\caption{Relative Scores on seven Atari games, environment step=400k, batch size=32, Number of seeds=20. The color of a method reflects its category same as Fig.~\ref{fig:dmcbox}. The overall results show that image augmentation for RL does not benefit policy learning on Atari which is quite different from DMControl. Most of the self-supervised losses fail to bring improvements even given more computation and extra model capacity from the SSL head. Only Rainbow+AE outperforms Efficient Rainbow, which is inconsistent with SAC+AE. ELo-Rainbow achieves worse results even than some of the SSL-based methods in the search space like BYOL and Rainbow+AE. The high variance and the image domain gap between different games make it extremely challenging for ELo-Rainbow to find the combined loss that generalizes to all environments.}
\label{fig:ataribox}
\end{figure}
\label{sec:atari}
Atari 2600 Games are also challenging benchmarks but with discrete action space~\cite{bellemare2013arcade}.
We choose seven games in this benchmark for selected methods.
All the methods use Efficient Rainbow~\cite{van2019use} as the RL method, which is a Rainbow~\cite{hessel2018rainbow} variant with modifications for better data efficiency.
Note that Efficient Rainbow, as a baseline, does not take advantage of image augmentation.
Therefore, we also benchmark Rainbow-Aug which is essentially Efficient Rainbow taking the augmented images for policy learning instead.
We use the same image augmentation and hyper-parameters reported by CURL for all applicable methods.
For a fair comparison, the augmentation for DrQ* is also adopted from CURL, which is different from what the original DrQ paper suggested.
We denote our setting as DrQ* to distinguish it from the original DrQ. 
Similarly, Rainbow+AE takes augmented images.
For each game, we run 20 random seeds and benchmark the agent at 400K environment steps (100K model steps with a frame skip of 4).
We report interquartile mean, standard deviation, and Relative Scores same as DMControl (See Table~\ref{tab:Atari}). 

Figure \ref{fig:ataribox} shows a summary of the seven different tasks in Relative Score.
Firstly, compared to vanilla baseline Efficient Rainbow which does not have any image augmentation or self-supervised learning, Rainbow-Aug performs worse overall with additional image augmentation for RL.
This suggests that the image augmentation used for self-supervised learning in CURL does not easily transfer. 
Similarly, DrQ* achieves compromised performance than Efficient Rainbow, showing that using image augmentation for Rainbow on Atari does not benefit policy learning unlike SAC on DMControl.
Based on the inconsistent impacts of image augmentation, further investigation is required when applying image augmentation to RL on Atari.

As for the self-supervised losses, BYOL, Rainbow+AE, Extract-Reward, and Predict-Reward gain better performance than CURL.
However, only Rainbow+AE shows significant improvement on Efficient Rainbow and outperforms all the other tested methods, which interestingly is inconsistent with SAC+AE on DMControl.
Predict-FR-Balanced, which shows considerable improvements on DMControl by manually combining two self-supervised losses, even fails to surpass Predict-Reward on Atari.
ELo-Rainbow, which searches in Frostbite, improves the baseline only in demon attack and frostbite.
The high variance on this benchmark made the evolutionary search extremely difficult. 
Further, there are huge image domain gaps between games, which makes it even harder for ELo-Rainbow to work across multiple games on Atari.

\begin{figure}
    \centering
    \begin{minipage}{0.4\textwidth}
    \begin{center}
        \includegraphics[width=0.8\linewidth]{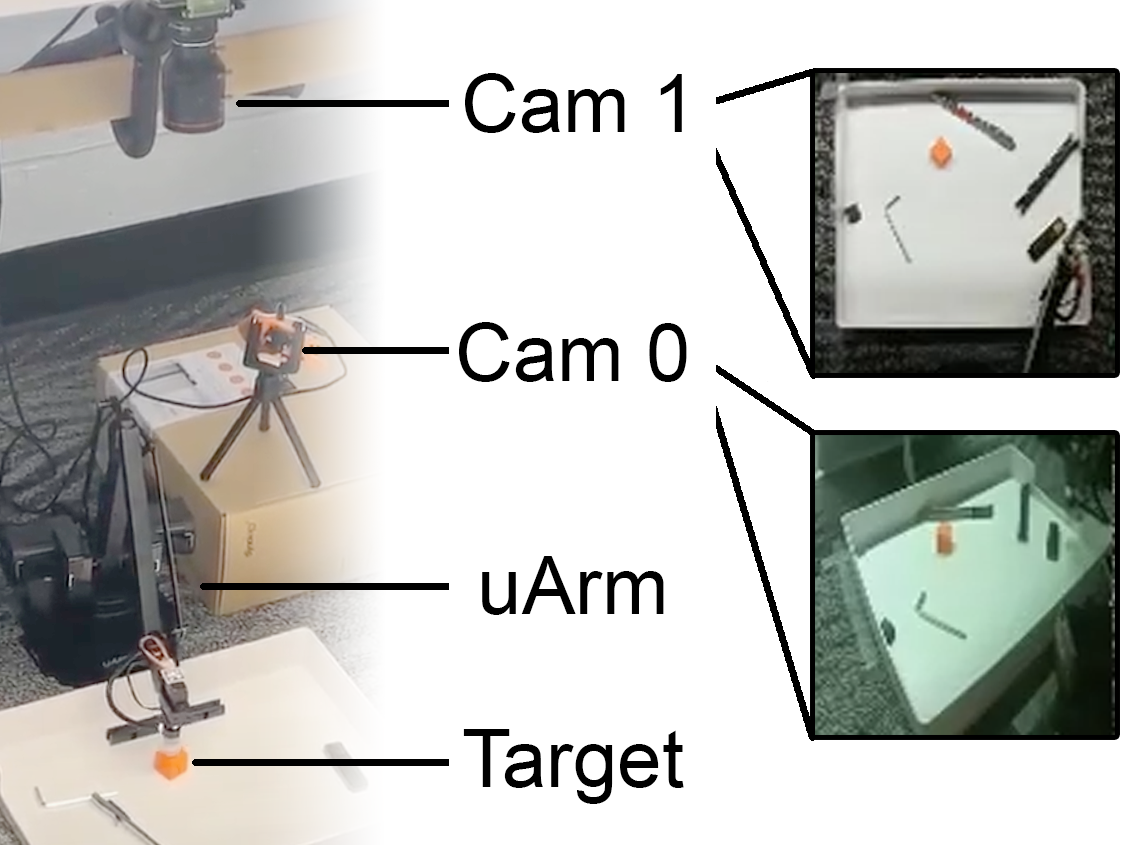}
    \end{center}
    \caption{Real robot environment setup}
    \label{fig:robotobs}
        \end{minipage}
    \hfill
    \begin{minipage}{0.55\textwidth}
    \begin{center}
        \includegraphics[width=1\linewidth]{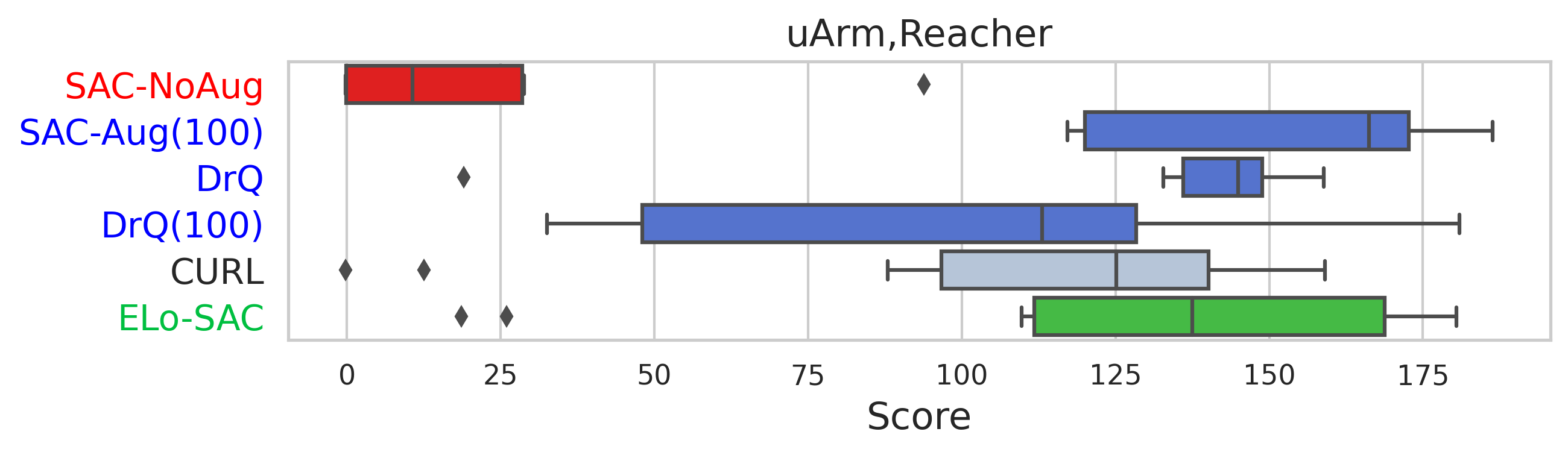}
    \end{center}
    \caption{Scores on real robot uArm, reacher, environment step=100k, batch size=512. The agent fails to learn effective policy without image augmentation.}
    \label{fig:robotres}
    \end{minipage}
\end{figure}

\paragraph{Real Robot Experiments} We further conduct experiments in a real-world robot environment, uArm reacher. 
Similar to \citet{burgerttriton}, the goal is to move the actuator close to a target object as fast as possible.
Our autonomous training environment and results are shown in Figs.~\ref{fig:robotobs} and \ref{fig:robotres} (Please check Appendix~\ref{sec:robot} for environment setup details).
We benchmark all methods with ten different random seeds, using the same hyper-parameters as DMControl experiments unless reported in Table~\ref{tab:hyrw}. 
Results are shown as Fig.~\ref{fig:robotres}, where ELo-SAC uses the optimal combination found in cheetah run shown as Table~\ref{tab:elodmc}.

Surprisingly, in this real-world environment, the agent fails to learn an effective policy without any image augmentation.
The image augmentation alone (i.e., SAC-Aug(100)) was sufficient to outperform other methods including CURL and ELo-SAC using self-supervision.
SAC-Aug(100) performs even better than DrQ, which is quite different from our previous observations on DMControl.
From all three methods only relying on image augmentation (blue in Fig.~\ref{fig:robotres}), we conclude that it requires a careful design of image augmentation that helps in a specific task/environment.


\section{Related Works}
Self-supervised learning can fit in robot policy learning in multiple fashions and at different stages.
Some works~\cite{ha2018world, sermanet2018time, zhan2020framework, shah2021rrl, stooke2021decoupling, shang2021self, wang2022vrl3, xiao2022masked} use SSL for representation learning in a pre-training stage before policy learning.
Others~\cite{oord2018representation, igl2018deep, hafner2019learning, yingjun2019learning, yarats2019improving, lee2020stochastic, laskin2020curl, zhu2020masked, mazoure2020deep, lee2020predictive, guo2020bootstrap, schwarzer2020data, zhang2020learning, yu2021playvirtual} jointly optimize the self-supervised loss with policy learning.
Specifically, Transporter~\cite{kulkarni2019unsupervised} and VAI~\cite{wang2021unsupervised} train an unsupervised keypoint detector to discover critical objects in the image for control. 
RRL~\cite{shah2021rrl} and VRL3~\cite{wang2022vrl3} also benefit from pre-training a deeper visual encoder on large datasets like ImageNet~\cite{deng2009imagenet}.
TCN~\cite{sermanet2018time} and CURL~\cite{laskin2020curl} take advantage of contrastive learning.
After the agent is deployed, SSL can be used to continuously improve the policy~\cite{hansen2020self}.
\citet{shelhamer2016loss} study several self-supervised losses within both the pretraining framework and the joint learning framework, while their selection of losses, the number of runs, and test environments are limited from a current point of view.
\citet{chen2021empirical} focus on imitation learning and test multiple SSL objectives for representation learning in various environments. 
They confirmed the critical role of image augmentation in imitation learning and showed inconsistencies in performance across environments.
Our investigation supports some of their observations, beyond that, our evolving loss, real robot environment, and representation analysis provide unique perspectives for online reinforcement learning.



\section{Discussion}
From DMControl and the real robot experiments, we empirically show that compared to the image augmentation, the role of existing self-supervised losses with the joint learning framework is usually limited, even with the help of evolutionary search.
While results on Atari show a different trend from DMControl, once again we confirm that there is no golden self-supervised loss or image augmentation that generalizes across environments. 
At the same time, it is usually challenging to conclude a consistent trend that one method is meaningfully better than others across multiple tasks.
One should cautiously decide the design choice of image augmentation or self-supervised loss for a specific RL task.
We are excited to see future works that introduce more self-supervised losses designed specifically for RL, as well as novel training frameworks that can benefit policy learning.

\section*{Acknowledgments}
We thank Kumara Kahatapitiya, Ryan Burgert and other lab members of Robotics Lab for valuable discussion. 
We thank Hanyi Yu and Rui Miao for their helpful feedback.
This work is supported by Institute of Information \& communications Technology Planning \& Evaluation (IITP) grant funded by the Ministry of Science and ICT (No.2018-0-00205, Development of Core Technology of Robot Task-Intelligence for Improvement of Labor Condition. 
This work is also supported by the National Science Foundation (IIS-2104404 and CNS-2104416). 

\newpage
\bibliographystyle{plainnat}
\bibliography{refs} 

\newpage

\begin{enumerate}

\item For all authors...
\begin{enumerate}
  \item Do the main claims made in the abstract and introduction accurately reflect the paper's contributions and scope?
    \answerYes{}
  \item Did you describe the limitations of your work?
    \answerYes{See Sec.~\ref{sec:limitation}}
  \item Did you discuss any potential negative societal impacts of your work?
    \answerNA{}
  \item Have you read the ethics review guidelines and ensured that your paper conforms to them?
    \answerYes{}
\end{enumerate}

\item If you are including theoretical results...
\begin{enumerate}
  \item Did you state the full set of assumptions of all theoretical results?
    \answerNA{}
        \item Did you include complete proofs of all theoretical results?
    \answerNA{}
\end{enumerate}

\item If you ran experiments...
\begin{enumerate}
  \item Did you include the code, data, and instructions needed to reproduce the main experimental results (either in the supplemental material or as a URL)?
    \answerYes{Please check supplemental material}
  \item Did you specify all the training details (e.g., data splits, hyper-parameters, how they were chosen)?
    \answerYes{See Sec.~\ref{sec:implementation}}
        \item Did you report error bars (e.g., with respect to the random seed after running experiments multiple times)?
    \answerYes{See Sec.~\ref{sec:figures}}
        \item Did you include the total amount of compute and the type of resources used (e.g., type of GPUs, internal cluster, or cloud provider)?
    \answerYes{See Sec.~\ref{sec:computation}}
\end{enumerate}

\item If you are using existing assets (e.g., code, data, models) or curating/releasing new assets...
\begin{enumerate}
  \item If your work uses existing assets, did you cite the creators?
    \answerYes{}
  \item Did you mention the license of the assets?
    \answerYes{}
  \item Did you include any new assets either in the supplemental material or as a URL?
    \answerYes{}
  \item Did you discuss whether and how consent was obtained from people whose data you're using/curating?
    \answerNA{}
  \item Did you discuss whether the data you are using/curating contains personally identifiable information or offensive content?
    \answerNA{}
\end{enumerate}

\item If you used crowdsourcing or conducted research with human subjects...
\begin{enumerate}
  \item Did you include the full text of instructions given to participants and screenshots, if applicable?
    \answerNA{}
  \item Did you describe any potential participant risks, with links to Institutional Review Board (IRB) approvals, if applicable?
    \answerNA{}
  \item Did you include the estimated hourly wage paid to participants and the total amount spent on participant compensation?
    \answerNA{}
\end{enumerate}

\end{enumerate}

\newpage

\appendix
\section{Appendix}

\subsection{Background on Pairwise Learning}
\label{sec:pairwise}
We coin the term ``pairwise'' learning for the frameworks that learn visual representations based on semantic invariance between dual-stream encoder representations.
A general pairwise learning method first generates multiple augmented views by applying a series of random image augmentations to the input sample, then clusters views with the same semantics in the representation space.
Optionally in such frameworks, methods using contrastive losses repel samples with different semantics.
Previous works not only have built various self-supervised tasks that benefit representation learning but also show that learned representations can benefit different downstream tasks.

In this paper, we focus on four representative pairwise learning methods, MoCo~\cite{he2020momentum, chen2020improved, mocov3}, BYOL~\cite{grill2020bootstrap}, SimSiam~\cite{chen2021exploring} and DINO~\cite{caron2021emerging}.
Specifically, MoCo takes advantage of the contrastive loss and negative samples in the mini-batch, while BYOL, SimSiam, and DINO focus on the similarity of the same image across diverse augmentations.

\textbf{MoCo}
\label{sec:moco}
\textbf{Mo}mentum \textbf{Co}ntrast (MoCo) takes advantage of a contrastive loss function InfoNCE~\cite{oord2018representation} with dot product similarity.
It starts from two identical encoder networks, an online encoder $f_q$ and a momentum encoder $f_k$.

At each training step, a mini-batch of $N$ images $x$ are uniformly sampled from a training set $D$. 
Given two distributions of image augmentations $\mathcal{T}$ and $\mathcal{T}'$, two image augmentations $t \sim \mathcal{T}$ and $t' \sim \mathcal{T}'$ are sampled respectively and applied to $x$, resulting in $2N$ samples.
Augmented images, $v=t(x)$ and $v'=t'(x)$, are called \textit{views}.
Then, $v$ and $v'$ are fed to two encoders to generate queries $q=f_q(v)$ and keys $k=f_k(v')$.

For each view $v_i$ in $v$ and its corresponding query $q_i=f_q(v_i)$, the contrastive loss is formulated as:
\begin{equation}
  \mathcal{L}_{\text{MoCo}, q_i} = - \log{\frac{\mathrm{sim}(q_i, k_i)}{\sum_{j=1}^N{\mathrm{sim}(q_i, k_j)}}}
  \label{eq:moco}
\end{equation}

where $\mathrm{sim}(q_i, k_i)=\exp(q_i \cdot \text{sg}[k_i] / \tau)$, $\text{sg}[\cdot]$ implies stop gradients and $\tau$ is a temperature hyper-parameter.
This loss encourages $q_i$ to be similar to its corresponding key $k_i$ (called \textit{positive}), but dissimilar to other keys (called \textit{negatives}) in the mini-batch.
The online encoder $f_q$ with parameters $\theta_q$ is updated by the above contrastive loss. The momentum encoder $f_k$ with parameters $\theta_k$, an Exponential Moving Average (EMA) of $f_q$, is updated by
\begin{equation}
  \theta_k\coloneqq m\theta_k + (1-m)\theta_q,
  \label{eq:ema}
\end{equation}
where $m \in [0, 1)$ is a momentum coefficient that controls how fast $\theta_k$ updates towards the online network $\theta_q$. Finally, $f_k$ will be discarded once the training completes.

\begin{figure}
  \centering
  \begin{subfigure}[b]{0.3\linewidth}
    \includegraphics[width=1\linewidth]{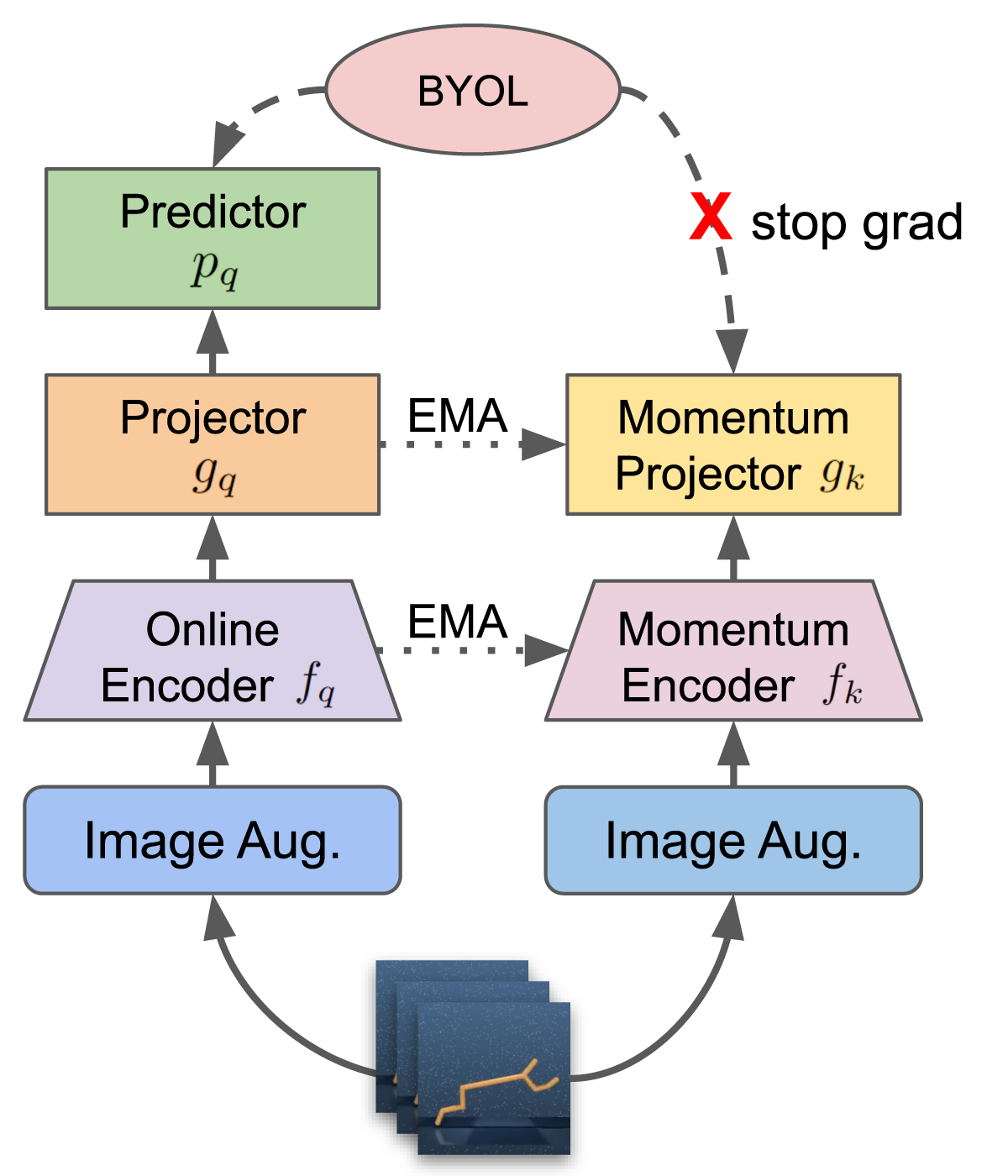}
    \caption{BYOL}
    \label{fig:byol}
  \end{subfigure}
  \hfill
  \begin{subfigure}[b]{0.3\linewidth}
    \includegraphics[width=1\linewidth]{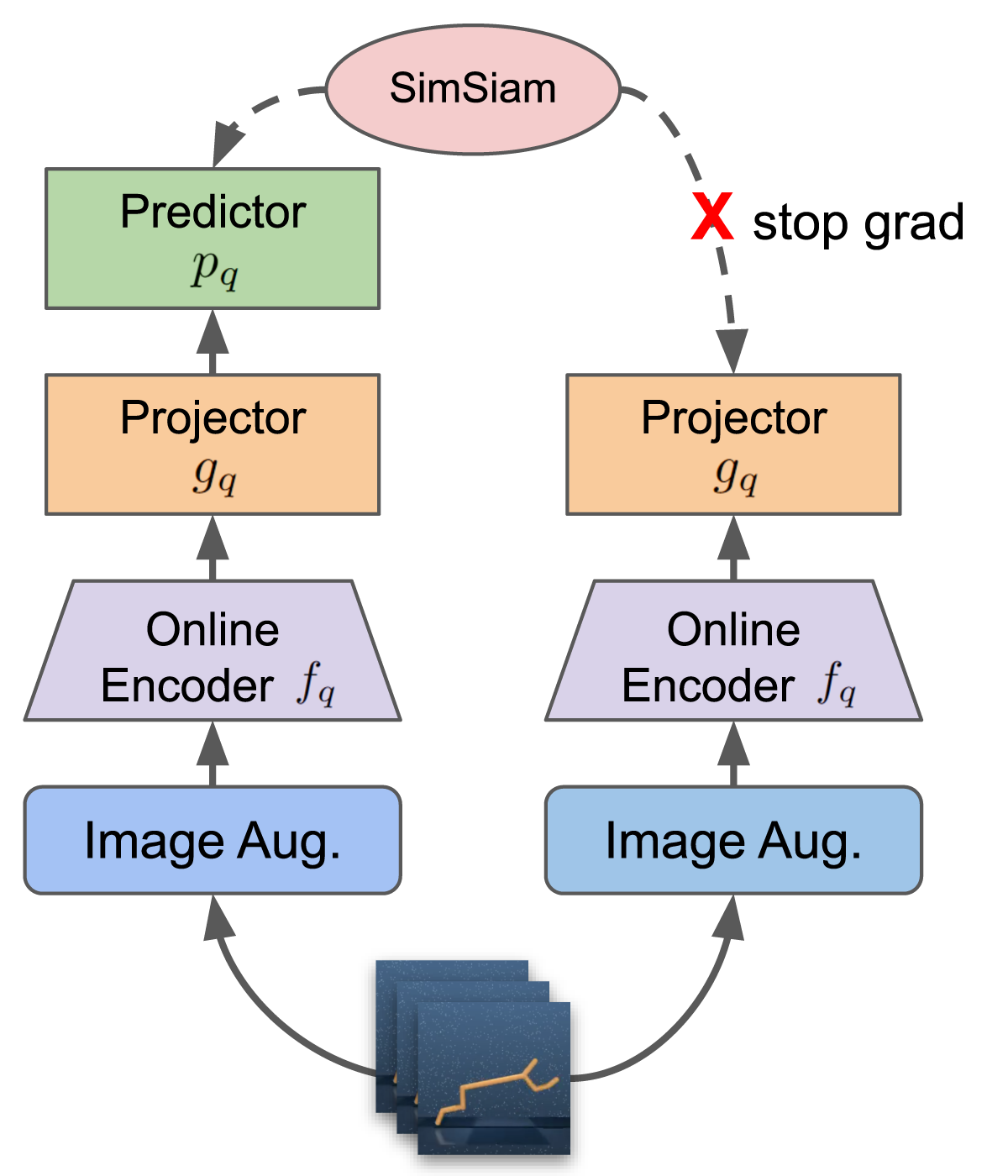}
    \caption{SimSiam}
    \label{fig:simsiam}
  \end{subfigure}
  \hfill
  \begin{subfigure}[b]{0.3\linewidth}
    \includegraphics[width=1\linewidth]{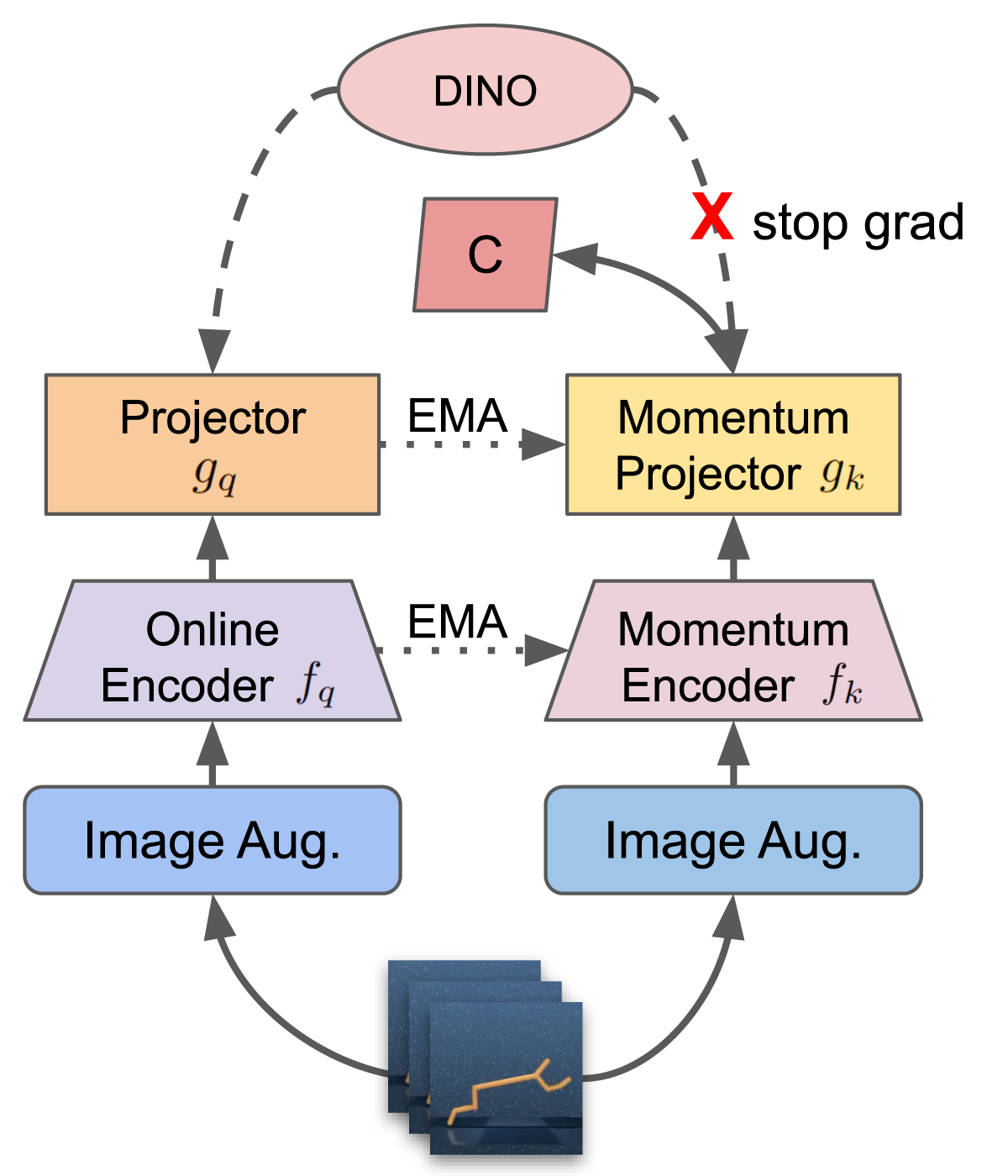}
    \caption{DINO}
    \label{fig:dino}
  \end{subfigure}
  \caption{Conceptual comparison of three pairwise learning frameworks}
  \label{fig:addachi}
\end{figure}

\textbf{BYOL}
Similar to MoCo, in addition to $f_q$ and $f_k$, BYOL maintains two identical projection networks $g_q$, $g_k$ and one prediction networks $p_q$ (See Fig.~\ref{fig:byol}).
BYOL also starts from inputs $v$ and $v'$ but calculates the projection $z_1=g_q(f_q(v))$ and $z_2=g_k(f_k(v'))$, and tries to regress $z_2$ from $z_1$ using the prediction network $p_q$.

After applying $l2$-normalization to the prediction $p_q(z_1)$ and the target projection $z_2$, a mean squared error is measured as:
\begin{equation}
  \mathcal{L}_{\text{BYOL}, 1} = \left\Vert p_q(z_1) - \text{sg}[z_2]\right\Vert^2_2 = 2 - 2 \frac{p_q(z_1) \cdot \text{sg}[z_2]}{\left\Vert p_q(z_1) \right\Vert_2 \cdot \left\Vert \text{sg}[z_2] \right\Vert_2}
\end{equation}
whose value is low when $p_q(z_1)$ is close to $z_2$.

Similarly, by swapping $v$ and $v'$, another symmetric loss can be applied on top of $z_1'=g_q(f_q(v'))$ and $z_2'=g_k(f_k(v))$, as $\mathcal{L}_{\text{BYOL}, 2} = \left\Vert p_q(z_1') - \text{sg}[z_2']\right\Vert^2_2$.
The total loss is $\mathcal{L}_{\text{BYOL}} = (\mathcal{L}_{\text{BYOL}, 1} + \mathcal{L}_{\text{BYOL}, 2}) / 2$.
The parameters of $f_k$ and $g_k$ are also the EMA of $f_q$ and $g_q$ respectively.

Finally, $f_k$, $g_q$, $g_k$ and $p_q$ will be discarded once the training completes.

\textbf{SimSiam}
SimSiam (\textbf{Sim}ple \textbf{Siam}ese) shares the same architecture as BYOL, while the parameters of the `momentum' branch of SimSiam are always tied to the `online' branch (See Fig.~\ref{fig:simsiam}).
Therefore, SimSiam only maintains one branch, including an encoder $f$, a projector $g$, and a predictor $p$.
SimSiam uses negative cosine similarity to encourage the predicted representation $h = p(g(f(v))$ to be similar to the projected representation of another view $g(f(v'))$, as follows:
\begin{equation}
  \mathcal{L}_{\text{SimSiam}, 1} =  -\frac{h}{\left\Vert h \right\Vert_2} \cdot \frac{\text{sg}[g(f(v'))]}{\left\Vert \text{sg}[g(f(v'))] \right\Vert_2}
  \label{eq:simsiam}
\end{equation}

Another symmetric loss term can also be derived as $\mathcal{L}_{\text{SimSiam}, 2}=-\frac{h'}{\left\Vert h' \right\Vert_2} \cdot \frac{\text{sg}[g(f(v))]}{\left\Vert \text{sg}[g(f(v))] \right\Vert_2}$, where $h'=p(g(f(v')))$.
The total loss is $\mathcal{L}_{\text{SimSiam}} = (\mathcal{L}_{\text{SimSiam}, 1} + \mathcal{L}_{\text{SimSiam}, 2}) / 2$.
And $g_q$ will be discarded once the model is trained.

\textbf{DINO}
DINO shares a similar overall structure as MoCo which contains two encoders $f_q$ and $f_k$, and $f_k$ is the EMA of $f_q$ (See Fig.~\ref{fig:dino}).
The outputs of both encoder networks are normalized as probability distributions over $K$ dimensions by applying softmax with a temperature parameter $\tau_t$, and $K$ is the dimension of $f_q(v)$.
DINO also maintains a centering vector $C$ with dimension $K$.
Following the formulation of knowledge distillation, a cross-entropy loss is applied to encourage the output distribution of $f_q$ to become similar to a centered distribution from $f_k$, as follows:

\begin{equation}
  \mathcal{L}_{\text{DINO}, 1} =  -P(\text{sg}[f_k(v')] - C) \cdot \log{P(f_q(v))}
  \label{eq:dino}
\end{equation}

where $P(x)=\text{softmax}(x / \tau_t)$. 
By swapping $v$ and $v'$ in Eq.~\ref{eq:dino}, another loss $\mathcal{L}_{\text{DINO}, 2}$ which is symmetric to $\mathcal{L}_{\text{DINO}, 1}$ can be derived.
And the total loss is the mean value of $\mathcal{L}_{\text{DINO}, 1}$ and $\mathcal{L}_{\text{DINO}, 2}$.

After each step of optimization, $f_k$ is updated by Eq.~\ref{eq:ema}.
$C$ also gets updated in a similar manner:
\begin{equation}
  C \coloneqq m_cC + (1-m_c) \cdot \text{mean}(f_k(v), f_k(v'))
\end{equation}
Here $m_c \in [0, 1)$ is another momentum coefficient.

\subsection{Implementation Details}
\label{sec:implementation}
Here we present the implementation details in all settings.

\subsubsection{General Joint Learning Framework}
The general joint learning framework starts from the official implementation of CURL~\cite{laskin2020curl} for DMControl and Atari.
For different self-supervised learning losses, we only replace the contrastive learning head of CURL with a different SSL head and update the loss calculation.
All the hyper-parameters are left untouched, except that we use the learning rate $10^{-3}$ for all DMControl environments.
The detailed hyper-parameters can be found at Table~\ref{tab:hydmc} (DMControl) and Table~\ref{tab:hyatari} (Atari).
We keep the most of hyper-parameters from DMControl for real-world robot experiments.
The modified configuration is listed as Table~\ref{tab:hyrw}.

We use the official implementations of DrQ~\cite{yarats2020image} and RAD~\cite{laskin2020reinforcement} for DMControl benchmark.
The re-implementation of DrQ (denoting as DrQ*) has the same joint learning framework and image augmentation from CURL.

\begin{table}[ht]
\caption{hyper-parameters used for DMControl with general joint learning framework}
\label{tab:hydmc}
\centering
\begin{tabular}{rl}
\toprule
\textbf{Hyperparameter} & \textbf{Value} \\
\midrule
Image augmentation & Random crop\\
Image size before augmentation &  (100, 100)\\
Image size after augmentation &  (84, 84)\\
Replay buffer size & 100000\\
Number of environment step & 100000\\
Initial explore steps & 1000\\
Stacked frames & 3\\
Action repeat & 2 finger, spin; walker, walk; \\
& reacher, hard; hopper, hop\\
& 8 cartpole, swingup\\
& 4 otherwise\\
\midrule
Critic target update frequency & 2\\
Actor update freq & 2\\

EMA $\tau$ for $Q', g_k$ & 0.01\\
EMA $\tau$ for $f_k$ & 0.05\\

Discount $\gamma$ & .99\\
Initial $\alpha$ & 0.1\\
\midrule
Convolutional layers in $f_q$ & 4\\
Number of filters & 32 \\
Fully connected layer in $f_q$ & 1\\
Tanh after $f_q$ & False\\
Image representation dimension & 50 \\
MLP layer of $Q_q^i, A_p$ & 3\\
MLP Hidden units & 1024 \\
MLP Non-linearity & ReLU \\

\midrule
Optimizer & Adam\\
$(\beta_1, \beta_2) \rightarrow (f_q, Q_q^i, A_p)$ & (.9, .999)\\
$(\beta_1, \beta_2) \rightarrow (\alpha)$ & (.5, .999)\\
Learning rate $(f_q, Q_q^i, A_p)$ & $10^{-3}$\\
Learning rate $(\alpha)$ & $10^{-4}$\\
Batch size & 512\\
\midrule

Evaluation episodes & 10\\
Train with random seeds & 10\\
\bottomrule

\end{tabular}
\end{table}

\begin{table}[ht]
\caption{Modified hyper-parameters for real-world robot experiments}
\label{tab:hyrw}
\centering
\begin{tabular}{rl}
\toprule
\textbf{Hyperparameter} & \textbf{Value} \\
\midrule
Stacked frames & 1\\
Action repeat & 1\\
Number of environment step & 100000\\
Number of random seeds & 10\\
\bottomrule

\end{tabular}
\end{table}

\begin{table}[ht]
\caption{hyper-parameters used for Atari with general joint learning framework}
\label{tab:hyatari}
\centering
\begin{tabular}{rl}
\toprule
\textbf{Hyperparameter} & \textbf{Value} \\
\midrule
Image augmentation pipeline with image size& Original Image (84, 84)$\rightarrow$\\
 & Random crop (80, 80) $\rightarrow$\\
 & Replication padding (88, 88) $\rightarrow$\\
 & Random crop (84, 84)\\
Replay buffer size & 100000\\
Number of environment step & 400000\\
Initial explore steps & 1600\\
Stacked frames & 4\\
Frameskip & 4\\
Action repeat & 4 \\
\midrule
Discount $\gamma$ & .99\\
Priority exponent & 0.5\\
Priority correction & $0.4 \rightarrow 1$\\
Target update frequency & 2000\\
Support of Q distribution & 51 bins\\
EMA $\tau$ for $f_k$ & 0.05\\
Reward Clipping & $[-1, 1]$\\
Max gradient norm & 10\\

\midrule
Convolutional layers in $f_q$ & 2\\
Number of filters & (32, 64) \\
Image representation dimension & 576 \\
Fully connected layer type & Noisy Nets\\
Noisy nets parameter & 0.1 \\
MLP layer of $Q_q$ & 2\\
MLP Hidden units & 256 \\
MLP Non-linearity & ReLU \\

\midrule
Optimizer & Adam\\
$(\beta_1, \beta_2) \rightarrow (f_q, Q_q)$ & (.9, .999)\\
Learning rate & $10^{-4}$\\
Batch size & 32\\
\midrule

Evaluation episodes & 10\\
Train with random seeds & 20\\
\bottomrule

\end{tabular}
\end{table}

\subsubsection{Losses for Self-supervised Learning}

\textbf{Pairwise Learning}
In this section, we replace the contrastive learning head with projectors and encoders depending on the exact loss.
All the projectors and predictors are two-layer MLPs with ReLU in the middle. 
The input dimension which is the output dimension of the encoder, ($50$ on DMControl and uArm Reacher, and $576$ on Atari).
The hidden dimension of the MLP is $256$ and the output dimension is $128$.
We use the same encoder EMA update rate ($\tau=0.05$ for SAC and $0.001$ for Rainbow) to update the projector$g_k$ (if applicable) in the target branch.
The applied losses are introduced in Sec.~\ref{sec:pairwise}.

\textbf{Transformation Awareness}
We use a two-layer MLP with ReLU as the classifier for both rotation classification and shuffle classification.
The hidden dimension of the MLP is $1024$ and the classifier is supervised by a cross-entropy loss.
The output dimension is $4$ for four-fold rotation classification and $1$ for binary shuffle classification.

\textbf{Reconstruction}
In this section, we follow the official implementation of SAC+AE~\cite{yarats2019improving} and apply the same image augmentation from CURL.
The decoder has one fully connected layer and the same number of transposed convolutional layers as the convolutional layers in the encoder.
When the output image from the decoder is smaller than the ground truth, we crop the ground truth to the size of the decoder output from the upper left corner.

For MAE we start from augmented SAC+AE and first divide the augmented image into non-overlapping patches in the spatial domain with a size of $4 \times 4$.
Then we randomly mask $50\%$ of the patches by setting the pixel value of the masked patches to zero.
Finally, the reconstruction loss is modified to calculate MSE only over the masked patches.
Other regularization losses are left untouched.

\textbf{RL Context Prediction}
For all kinds of losses, the dimensions of all the fully connected layers and hidden layers in MLPs are $1024$.

\subsubsection{Manually Balance Two Self-supervised Losses}
\label{sec:balanced}
In this section, we further explore the ways to manually combine two self-supervised losses.
Extract-AR, Guess-AF, and Predict-FR are methods manually designed to combine two individual losses
However, Guess-AF and Predict-FR are not better than the single self-supervised loss in their combinations (see Guess-Action and Predict-Reward in Table~\ref{tab:DMC} and Fig.~\ref{fig:dmcbox}).
Considering that Extract-AR, Guess-AF, and Predict-FR concatenate both the outputs and apply supervision by averaging loss per element of the output, the target with a higher dimension will naturally get more penalty due to the larger number of elements in the output. 
We further test the `Balanced' configuration, where we only modify how the supervision is applied.
Take Extract-AR as an example, in the `Balanced' setting, we first calculate loss regarding action prediction and reward prediction separately, then the total self-supervised loss is the average of both the action prediction loss and the reward prediction loss.
By adjusting the combination weights, the `Balanced' trick brings overall improvements on top of all three methods as shown in Table~\ref{tab:balanced}. 
Such observation suggests that we need to carefully design how the two losses are combined, which is getting trickier as the number of combined losses increases.

\begin{table}[ht]
\caption{Scores on DMControl improved by manually balancing two self-supervised losses, suggesting the importance of weight hyper-parameters when combining multiple losses. Methods in gray are without a self-supervised loss for reference.}
\label{tab:balanced}
\centering
\resizebox{\textwidth}{!}{

\begin{tabular}{r|llllll}
\toprule
\textbf{Agent}&ball\_in\_cup,~catch&cartpole,~swingup&cheetah,~run&finger,~spin&reacher,~easy&walker,~walk\\
\midrule
\textcolor{gray}{\textbf{SAC-Aug(100)}}&\textcolor{gray}{541.4 ± 306.2}&\textcolor{gray}{563.4 ± 235.0}&\textcolor{gray}{172.1 ± 64.0}&\textcolor{gray}{724.6 ± 154.9}&\textcolor{gray}{654.4 ± 222.1}&\textcolor{gray}{422.1 ± 250.8}\\
\textcolor{gray}{\textbf{RAD}}&\textcolor{gray}{879.9 ± 82.0}&\textcolor{gray}{786.4 ± 95.1}&\textcolor{gray}{387.9 ± 81.3}&\textcolor{gray}{910.4 ± 104.5}&\textcolor{gray}{508.8 ± 111.5}&\textcolor{gray}{522.1 ± 95.5}\\
\textcolor{gray}{\textbf{DrQ}}&\textcolor{gray}{914.9 ± 21.2}&\textcolor{gray}{692.2 ± 222.9}&\textcolor{gray}{360.4 ± 67.7}&\textcolor{gray}{935.6 ± 201.3}&\textcolor{gray}{713.7 ± 147.6}&\textcolor{gray}{523.9 ± 182.2}\\
\midrule
\textbf{Extract-AR}&822.2 ± 240.5&592.9 ± 124.7&225.8 ± 60.7&783.0 ± 112.0&678.7 ± 181.3&458.4 ± 148.9\\
\textbf{Extract-AR-Balanced}&897.9 ± 113.9(\textcolor{mygreen}{↑75.7})&582.3 ± 119.2(\textcolor{red}{↓-10.6})&232.9 ± 33.2(\textcolor{mygreen}{↑7.1})&881.5 ± 114.2(\textcolor{mygreen}{↑98.5})&720.5 ± 136.4(\textcolor{mygreen}{↑41.8})&533.8 ± 100.8(\textcolor{mygreen}{↑75.4})\\
\midrule
\textbf{Guess-AF}&329.8 ± 298.4&140.7 ± 144.0&0.9 ± 22.8&880.0 ± 59.5&382.9 ± 265.0&494.7 ± 112.7\\
\textbf{Guess-AF-Balanced}&918.4 ± 353.5(\textcolor{mygreen}{↑588.6})&536.1 ± 190.3(\textcolor{mygreen}{↑395.4})&191.2 ± 78.6(\textcolor{mygreen}{↑190.3})&842.9 ± 67.9(\textcolor{red}{↓-37.1})&462.0 ± 208.7(\textcolor{mygreen}{↑79.1})&507.0 ± 128.7(\textcolor{mygreen}{↑12.3})\\
\midrule
\textbf{Predict-FR}&750.3 ± 256.0&723.2 ± 167.5&12.4 ± 35.7&861.5 ± 49.2&636.1 ± 201.4&270.0 ± 154.9\\
\textbf{Predict-FR-Balanced}&829.6 ± 241.6(\textcolor{mygreen}{↑79.3})&751.0 ± 90.0(\textcolor{mygreen}{↑27.8})&216.2 ± 77.6(\textcolor{mygreen}{↑203.8})&864.8 ± 72.2(\textcolor{mygreen}{↑3.3})&882.1 ± 87.4(\textcolor{mygreen}{↑246.0})&472.9 ± 189.7(\textcolor{mygreen}{↑202.9})\\
\bottomrule
\end{tabular}

}

\end{table}

\subsubsection{Evolving Multiple Self-supervised Losses}
\label{sec:implementelo}

We choose PSO~(\textbf{P}article \textbf{S}warm \textbf{O}ptimization)~\cite{kennedy1995particle} for the optimal combination of hyper-parameters including $N_w$ weights of losses $w_{i=1, 2, ...,N_w}$ and two magnitudes of augmentation $m_{j=1, 2}$ for the online networks and the target networks respectively.
Each $m_j$ varies from $[84, 116]$.
We limit the range of each $w_i$ to $[0, 10]$ for ELo-SAC and ELo-Rainbow while a range of $[10^{-4}, 10^{4}]$ for ELo-SACv2 and ELo-SACv3.

During the evolutionary search, we use a batch size of $128$ for ELo-SAC and ELo-SACv2, each combination is trained with $5$ different random seeds.
As for ELo-SACv3, the batch size is set to $64$ and the number of random seeds is set to $3$ to save computation.
ELo-Rainbow also train with $5$ random seeds during the evolutionary search.
Other hyperparameters used in the search and all hyperparameters for evaluation are identical to Table~\ref{tab:hydmc} and Table~\ref{tab:hyatari}.

\paragraph{ELo-SAC}
ELo-SAC maintains a population of $50$ for DMControl and each particle evolves 15 generations in ``cheetah, run''.
Before the search, the first $i^{\text{th}}$ particles are initialized with $m_{j=1,2}=88$, and each particle only has one weight set to $1$ and other weights set to $0$.
In another word, these first $i^{\text{th}}$ particles start with the existing single self-supervised loss method in the search space.
Other particles are randomly initialized.
Table~\ref{tab:elodmc} shows the combination Elo-SAC found in cheetah run.
The columns in Table~\ref{tab:elodmc} show the search space.
The first six columns denote the optimal weight $w_i$ of its corresponding loss obtained with the evolutionary search, while the last two columns denote the original image size before random crop (image augmentation magnitude $m_{j=1,2}$).
\begin{table}[ht]
    \caption{Optimal parameters that ELo-SAC found in cheetah run}
    \label{tab:elodmc}
    \centering
    
  \resizebox{\textwidth}{!}{
  \begin{tabular}{@{}cc|cccccc|cc@{}}
    \toprule
    {Agent}  &  \thead{Searched\\Env.}& \thead{CURL\\$w_1$} & \thead{BYOL\\$w_2$}  & \thead{Predict FR\\$w_3$} & \thead{Extract AR\\$w_4$} & \thead{AutoEncoder\\$w_5$} & \thead{RotationCLS\\$w_6$} & \thead{Online Aug.\\$m_1$} & \thead{Target Aug.\\$m_2$}\\
    \midrule
    ELo-SAC & Cheetah, run & 0 & 0.288 & 0.628 & 0 & 0 & 0.009 & 87 & 86\\
    \bottomrule
  \end{tabular}
  }
  
\end{table}

\paragraph{ELo-SACv2}
Compared with ELo-SAC, ELo-SACv2 has the following major improvements:
\begin{enumerate}
\item \textbf{Initialization} Define the set of image augmentation magnitude $\mathcal{M} = \{(m_1 = t, m_2 = t) \mid t = 86, 88, 92, 100, 116\}$, 
and the set of SSL weights $\mathcal{W} = \{(w_{i = t} = 1, w_{i \neq t} = 0) \mid t = 1, 2, ..., 6\}$. 
The first $|\mathcal{M}| \times |\mathcal{W}|$ particles are initialized from the Cartesian product of $\mathcal{M}$ and $\mathcal{W}$. 
Other particles are randomly initialized.
\item \textbf{Search space} The search space of self-supervised losses is updated based on the loss performance at Table~\ref{tab:DMC}. We empirically choose the losses from different categories that have a relatively strong performance when it is applied solely to RL.

\item \textbf{Weight range} The weight of each self-supervised loss is presented on a log scale so that the search can cover a larger range.
\end{enumerate}

Besides the improvements above, ELo-SACv2 evolves $45$ generations and the optimal combination is chosen from the top 10 combinations regarding the overall performance. 
The optimal combination found by ELo-SACv2 is shown in Table~\ref{tab:elodmc2}. 
ELo-SACv2 slightly improves the results of ELo-SAC with all the modifications (see Figure~\ref{fig:dmcbox} and Table~\ref{tab:DMC}).

\begin{table}[ht]
    \caption{Optimal parameters that ELo-SACv2 found in cheetah run}
    \label{tab:elodmc2}
    \centering
    
  \resizebox{\textwidth}{!}{
  \begin{tabular}{@{}cc|cccccc|cc@{}}
    \toprule
    {Agent}  &  \thead{Searched\\Env.}& \thead{CURL\\$\log_{10}{w_1}$} & \thead{DINO\\$\log_{10}{w_2}$}  & \thead{Predict FR\\Balanced\\$\log_{10}{w_3}$} & \thead{Extract AR\\Balanced\\$\log_{10}{w_4}$} & \thead{AutoEncoder\\$\log_{10}{w_5}$} & \thead{RotationCLS\\$\log_{10}{w_6}$} & \thead{Online Aug.\\$m_1$} & \thead{Target Aug.\\$m_2$}\\
    \midrule
    ELo-SACv2 & Cheetah, run & -3.309 & -0.562 & 1.272 & -0.772 & -3.904 & 0.344 & 88 & 91\\
    \bottomrule
  \end{tabular}
  }
  
\end{table}

\paragraph{ELo-SACv3}
Since ELo-SAC and ELo-SACv2 only search in one DMControl environment, `cheetah run', and both the found solutions perform weaker on `finger, spin' and `reacher, easy', we further extend ELo-SACv2 to search in multiple environments at the same time, named ELo-SACv3.
The optimization process of ELo-SACv3 is presented as:  
\begin{equation}
\argmax_{m_{j=1, 2}, w_{i=1, \dots,
N_l}} \text{mean}(\mathcal{\hat{R}_{\text{envs}}^{\text{seed=1,2,3}}}(m_{j=1, 2}, w_{i=1, \dots, N_l}))
\end{equation}
where $\hat{R}=R / R_{\text{DrQ}}$ is the original agent reward $R$ normalized by the score of DrQ $R_{\text{DrQ}}$ reported in \cite{yarats2020image}, and $\text{envs}$ is the set of six DMControl environments listed in Table~\ref{tab:DMC}.

We let ELo-SACv3 evolve for 25 generations and chose the loss combination with best performance among the top 10 records. The found parameters are listed in Table~\ref{tab:elodmc3}.

\begin{table}[ht]
    \caption{Optimal parameters that ELo-SACv3 found in six DMControl environments}
    \label{tab:elodmc3}
    \centering
    
  \resizebox{\textwidth}{!}{
  \begin{tabular}{@{}cc|cccccc|cc@{}}
    \toprule
    {Agent}  &  \thead{Searched\\Env.}& \thead{CURL\\$\log_{10}{w_1}$} & \thead{DINO\\$\log_{10}{w_2}$}  & \thead{Predict FR\\Balanced\\$\log_{10}{w_3}$} & \thead{Extract AR\\Balanced\\$\log_{10}{w_4}$} & \thead{AutoEncoder\\$\log_{10}{w_5}$} & \thead{RotationCLS\\$\log_{10}{w_6}$} & \thead{Online Aug.\\$m_1$} & \thead{Target Aug.\\$m_2$}\\
    \midrule
    ELo-SACv3 & 6 environments & -2.304 & -4.0 & -2.989 & 0.103 & -1.722 &  -3.481 & 88 & 89\\
    \bottomrule
  \end{tabular}
  }
  
\end{table}

\paragraph{ELo-Rainbow}
ELo-Rainbow has a population of $30$ and the initialization is similar to ELo-SAC.
The search is performed on Frostbite only for 10 generations and the found combination is shown in Table~\ref{tab:eloatari}.

\begin{table}[ht]
    \caption{Parameters of ELo-Rainbow found in Frostbite}
    \label{tab:eloatari}
    \centering
    \resizebox{\textwidth}{!}{
  \begin{tabular}{@{}cc|ccccc@{}}
    \toprule
    {Agent}  &  {Searched Env.}& \thead{BYOL\\$w_1$} & \thead{Predict Future\\$w_2$} & \thead{Extract Reward\\$w_3$}& \thead{AutoEncoder\\$w_4$} & \thead{Rotation CLS\\$w_5$} \\
    \midrule
    ELo-Rainbow & Frostbite & 0.250 & 1.054 & 2.280 & 0.953 & 0.591 \\
    \bottomrule
  \end{tabular}
  }
\end{table}

Interestingly, we find that the optimal combination found by ELo-SAC is relatively sparse, where BYOL and Predict FR are the only two major losses.
Similarly, ELo-SACv2 relies more on Predict-FR-Balanced and RotationCLS, while ELo-SACv3 relies on Extract-AR-Balanced mostly.
However, for ELo-Rainbow, the magnitudes of all the weights are relatively similar.
The difference between the found results reflects the different properties of different environments.
Our further experiments in DMControl confirm the generalization ability to evolve losses; i.e., the obtained solution of weights in one environment achieves relatively good performance on other environments in the same benchmark.
However, results on Atari are much inconsistent with DMControl.
We cover detailed observations and discussions in Section~\ref{sec:exp} and Appendix~\ref{sec:figures}.

The code for ELo-SACv3 is available at \url{https://github.com/LostXine/elo-sac}, and the code for ELo-Rainbow is available at \url{https://github.com/LostXine/elo-rainbow}.


\subsubsection{Comparison of method variants}
\label{sec:dataaugmentationdiff}
In Section~\ref{sec:exp}, several variants of the existing methods are introduced.
The difference between these methods, especially on image augmentation, can be summarized as follows: 
SAC-NoAug is the original pixel-based SAC~\cite{haarnoja2018softa, haarnoja2018softb}.
SAC-Aug(88) and SAC-Aug(100) use the random crop as the only image augmentation, where (88) means the original image has a size of $88 \times 88$ before randomly cropping to $84 \times 84$ and (100) means the original image has a size of $100 \times 100$.
These two methods should be regarded as variants of RAD with different augmentation choices.
The random crop from $100 \times 100$ to $84 \times 84$ is the default image augmentation method for all the methods introduced in Sec.~\ref{sec:sslopt}, including SAC+AE. 
Essentially, if we remove their self-supervised loss, they will fall back to SAC-Aug(100).
Similarly, we test DrQ variants by replacing its default random shift augmentation with the random crop, reported as DrQ(88) and DrQ(100).
Meanwhile, RAD uses random translate by default except on walker walk;
ELo-SAC and ELo-SACv2 first crop the center of the input image to the found optimal sizes.
Then two central patches are randomly cropped to $84 \times 84$ as the inputs for the online networks and the target networks respectively.

For the policy learning part, all the methods share the same model. 
However, DrQ, DrQ(88), DrQ(100), and SAC+AE apply an additional $\tanh$ activation after the visual encoder.
We also study the effect of the activation function in the coming Section~\ref{sec:ablationaf}.

\subsection{Ablations}
\label{sec:ablation}
Besides random crop and encoder backbone investigated in Section~\ref{sec:exp}, we further perform detailed ablations on more image augmentation, learning rate, encoder architecture, and activation function in this section.
The default test environment is identical to ablations in Section~\ref{sec:exp}.

\subsubsection{Image Augmentation}
\begin{figure}[ht]
  \centering
  \includegraphics[width=.7\linewidth]{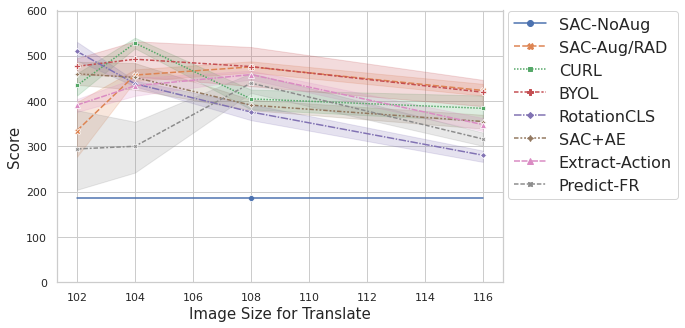}
  \caption{Ablations on translate image augmentation.}
  \label{fig:ablationtranslate}
\end{figure}

We study the effect of random translate, an image augmentation method which is widely used in RAD~\cite{laskin2020reinforcement}.
Similar to random crop, the image size for translate is linear to the magnitude of the translate, when using a fixed crop size: the larger the image size, the stronger the augmentation.
As shown in Fig.~\ref{fig:ablationtranslate}, image size for translate has a similar pattern for most tested methods (with an exception of RotationCLS).
In summary, it is critical to engineering image augmentation carefully when designing an RL system with or without SSL. 

\subsubsection{Learning Rate}
\begin{figure}[ht]
  \centering
  \includegraphics[width=.7\linewidth]{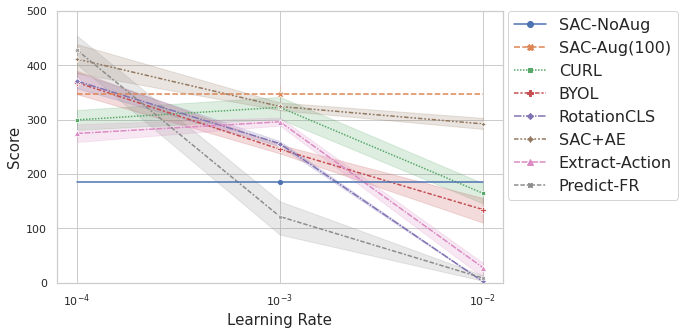}
  \caption{Ablations on the self-supervised learning rate.}
  \label{fig:ablationlr}
\end{figure}

Fig.~\ref{fig:ablationlr} shows how the learning rate of self-supervised loss contributes to the performance.
In this group of ablations, we only change the learning rate for SSL and leave the RL part untouched.
SAC-NoAug and SAC-Aug(100) are both baselines for reference without any self-supervised losses.
The results suggest that a smaller learning rate for SSL may improve performance.
Therefore, it is necessary to search for the absolute weights of losses like ELo~\cite{piergiovanni2019evolving}, which is equivalent to searching for the learning rate.

\subsubsection{Encoder Architecture}
\begin{figure}[ht]
  \centering
  \includegraphics[width=.7\linewidth]{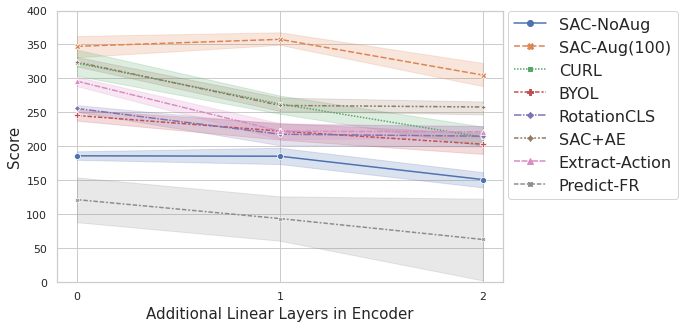}
  \caption{Ablations on additional linear layers after the visual encoder.}
  \label{fig:ablationencl}
\end{figure}

We further investigate the effect of additional linear layers in the visual encoder. 
Additional linear layers with ReLU activation are appended to the end of the visual encoder.
All additional layers have a latent dimension of $128$.
Fig.~\ref{fig:ablationencl} shows that additional layers usually bring downgraded performance, which could be caused by limited data.

\begin{figure}[ht]
  \centering
  \begin{minipage}{0.49\textwidth}
  \includegraphics[width=1\linewidth]{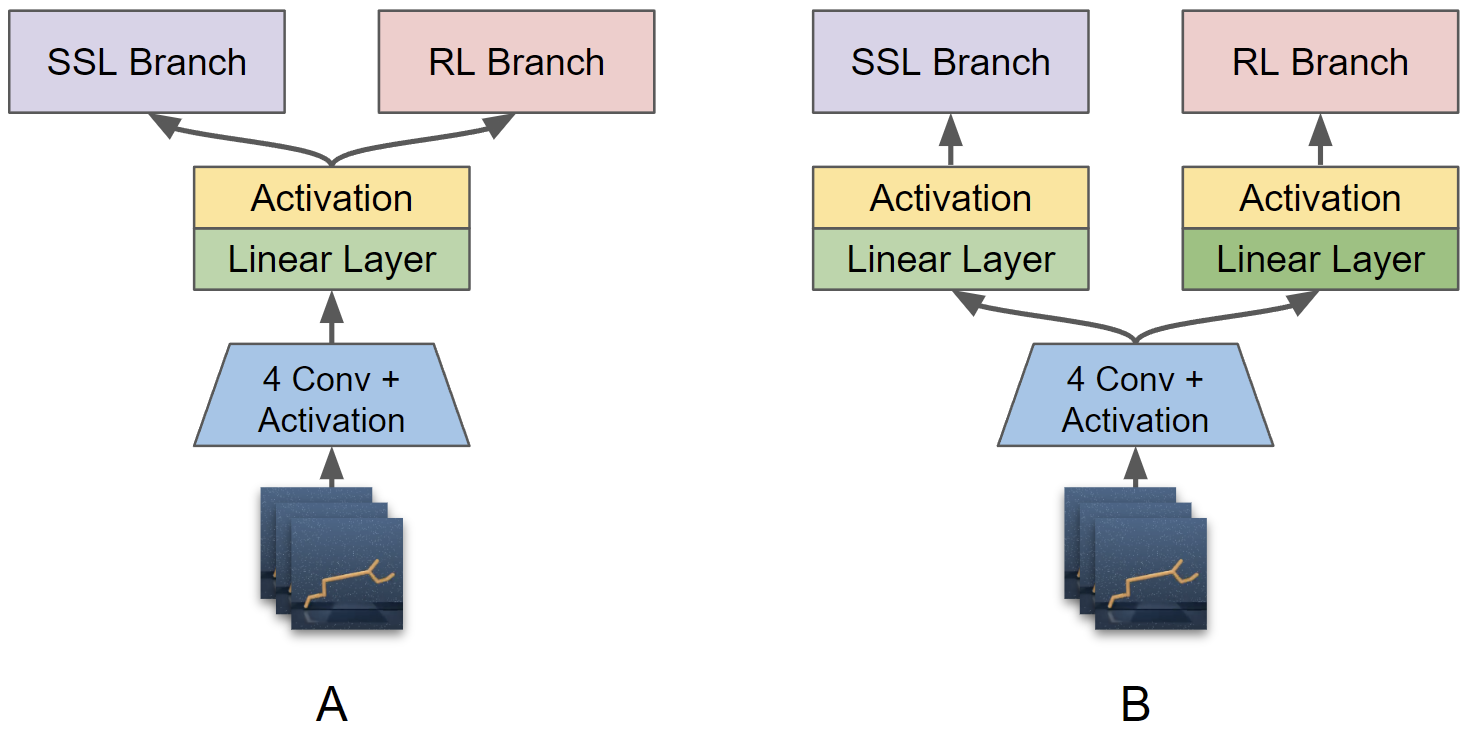}
  \caption{Comparison of two separation points.}
  \label{fig:ablationspp}
  \end{minipage}
  \hfill
  \begin{minipage}{0.49\textwidth}
        \includegraphics[width=1\linewidth]{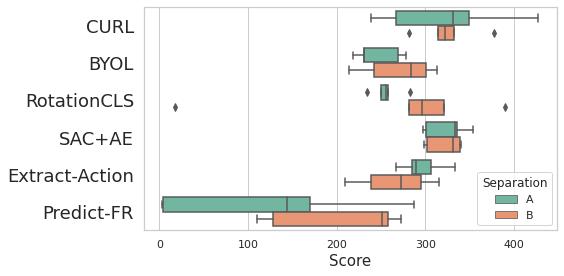}
        \caption{Ablation on separation points.}
        \label{fig:split}
    \end{minipage}
\end{figure}

Another important aspect of the encoder architecture design is how to merge or separate two branches.
As Visionary~\cite{akinola2021visionary} suggests, where and how to merge visual representation with action representation is critical when designing an efficient value network.
Similarly, we hypothesize that the point where split the representation for the SSL branch and the RL branch is also important.
Figure~\ref{fig:ablationspp} demonstrates two separation point configurations, named A and B.
Figure~\ref{fig:split} shows how the performance of different approaches changes in such two configurations.

\subsubsection{Activation Function}
\label{sec:ablationaf}
Though all the methods we tested in Table~\ref{tab:DMC} share the same visual encoder architecture, DrQ and SAC+AE apply an additional activation function $\tanh$ to the visual representation.
To make a fair comparison and study the effect of such an activation function, we conducted detailed ablations on six DMControl tasks using the hyper-parameters identical to Table~\ref{tab:hydmc}.
Results in Table~\ref{tab:tanh} confirm that the default design choice of all three methods is better than their alternatives.

\begin{table}[ht]
\caption{Ablation on Tanh activation}
\label{tab:tanh}
\centering
\resizebox{\textwidth}{!}{

\begin{tabular}{rccccccc|c}
\toprule
\textbf{Agent}&$Tanh$&ball\_in\_cup,~catch&cartpole,~swingup&cheetah,~run&finger,~spin&reacher,~easy&walker,~walk&Relative~Score\\
\midrule
\textbf{DrQ} (default)&\checkmark&\textbf{914.9 ± 21.2}&692.2 ± 222.9&360.4 ± 67.7&\textbf{935.6 ± 201.3}&\textbf{713.7 ± 147.6}&523.9 ± 182.2&\textbf{6.642}\\
\textbf{DrQ-w/o-Tanh}&&870.8 ± 177.0&\textbf{826.4 ± 44.9}&\textbf{393.7 ± 74.0}&849.6 ± 140.9&635.0 ± 155.0&\textbf{525.0 ± 163.7}&6.182\\
\midrule
\textbf{CURL-w-Tanh}&\checkmark&832.4 ± 118.4&508.5 ± 133.9&209.3 ± 24.2&676.3 ± 185.2&336.1 ± 216.5&463.7 ± 93.4&-3.266\\
\textbf{CURL} (default)&&730.0 ± 179.4&471.5 ± 89.9&215.1 ± 57.3&717.8 ± 136.5&569.8 ± 179.4&442.6 ± 87.1&-2.032\\
\midrule
\textbf{SAC+AE} (default)&\checkmark&616.1 ± 169.9&388.8 ± 130.1&291.8 ± 59.8&799.0 ± 138.9&481.3 ± 130.4&402.6 ± 161.5&-2.529\\
\textbf{SAC+AE-w/o-Tanh}&&358.9 ± 209.8&378.0 ± 96.0&289.4 ± 60.7&702.5 ± 157.3&516.8 ± 190.0&375.4 ± 136.2&-4.998\\
\bottomrule
\end{tabular}

}
\end{table}

\subsection{Detailed Results on DMControl and Atari}
\label{sec:figures}
\begin{figure}[ht]
  \centering
  \includegraphics[width=\linewidth]{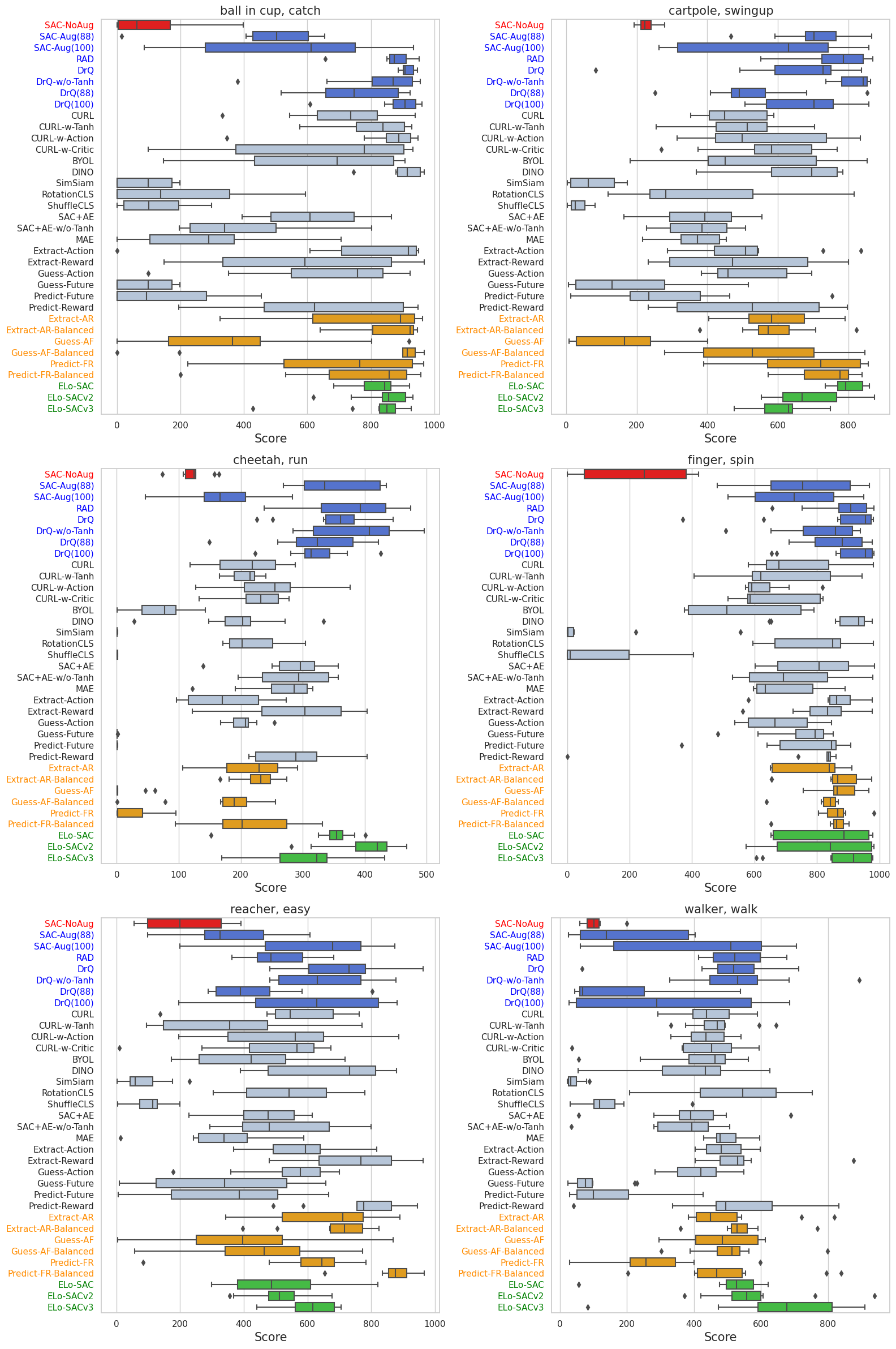}
   \caption{DMControl score distribution}
   \label{fig:fulldmc}
\end{figure}

\begin{figure}[ht]
  \centering
  \includegraphics[width=\linewidth]{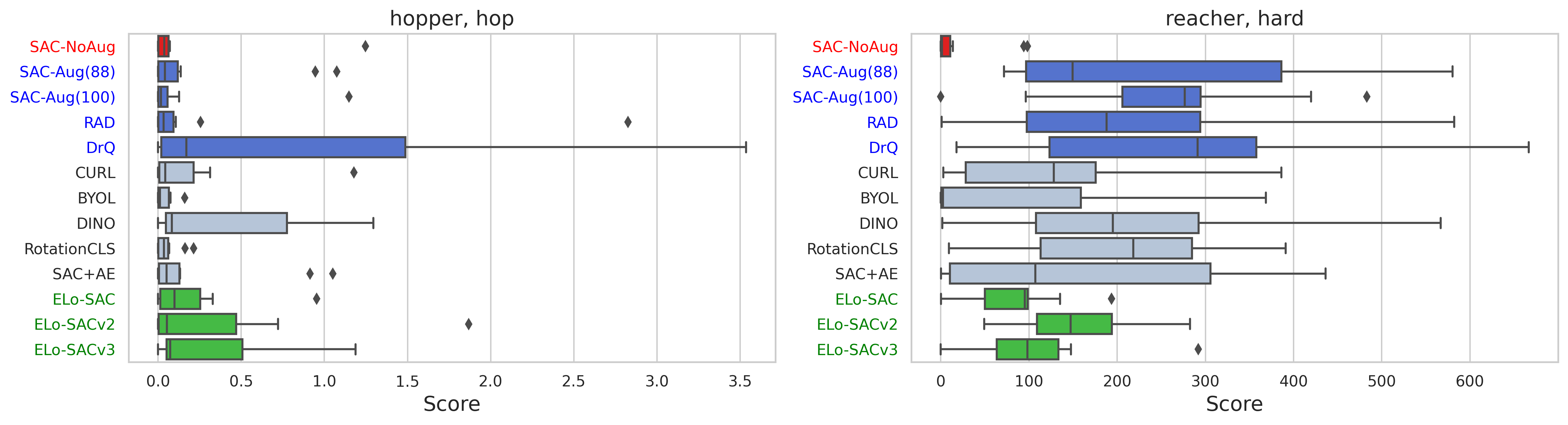}
   \caption{Hard DMControl score distribution}
   \label{fig:fulldmchard}
\end{figure}

\begin{figure}[ht]
  \centering
  \includegraphics[width=0.99\linewidth]{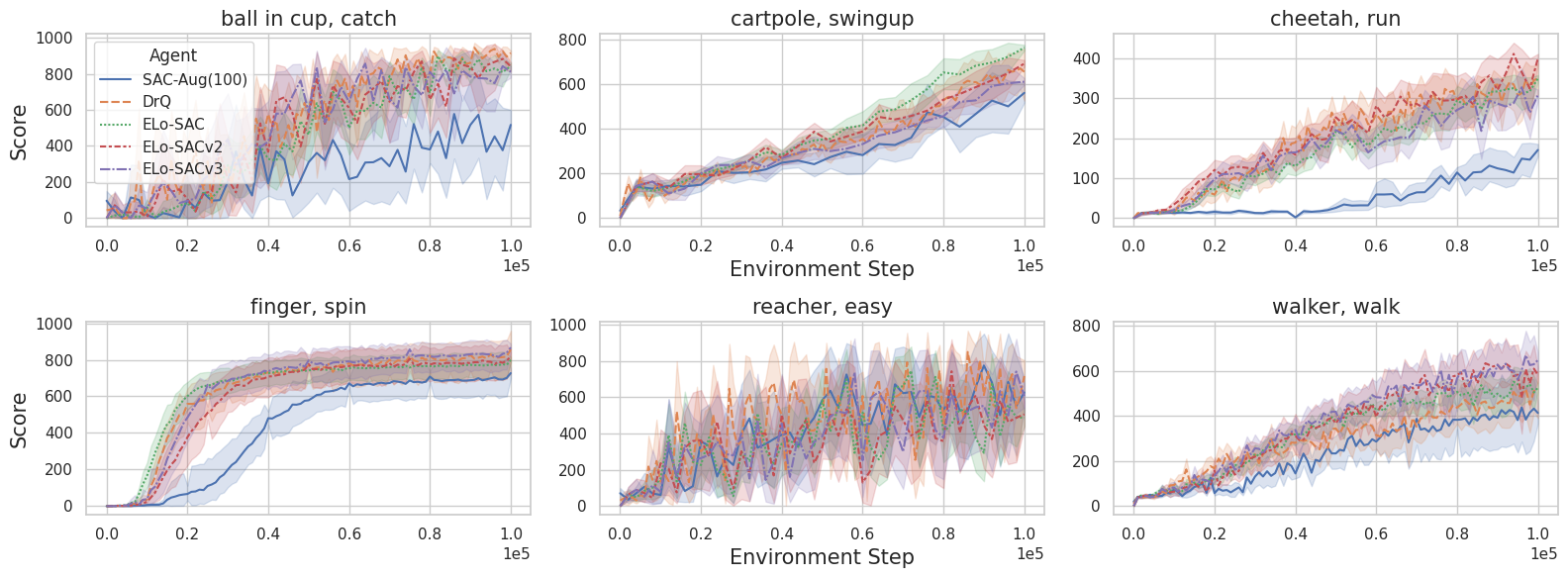}
  \caption{Step-reward curve of ELo-SAC based methods}
  \label{fig:full1}
\end{figure}

\begin{figure}[ht]
  \centering
  \includegraphics[width=0.99\linewidth]{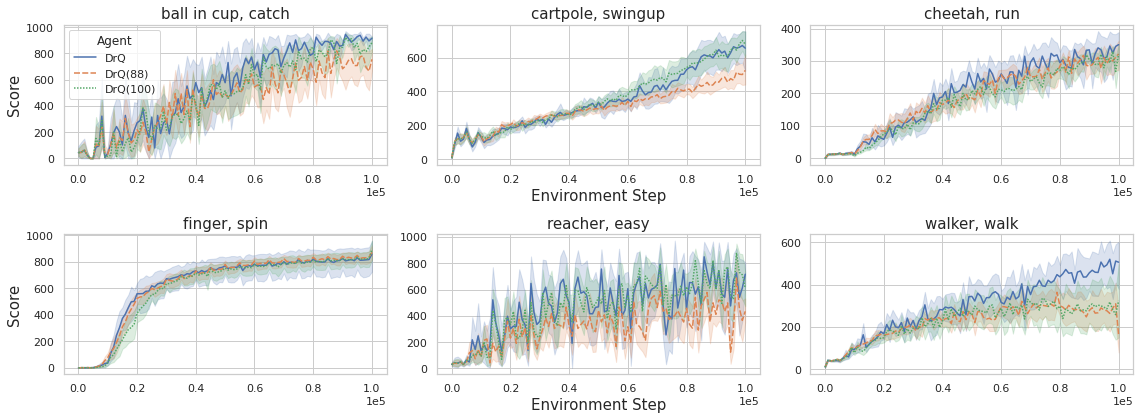}
  \caption{Step-reward curve of DrQ and its variants}
  \label{fig:full2}
\end{figure}

\begin{figure}[ht]
  \centering
  \includegraphics[width=0.99\linewidth]{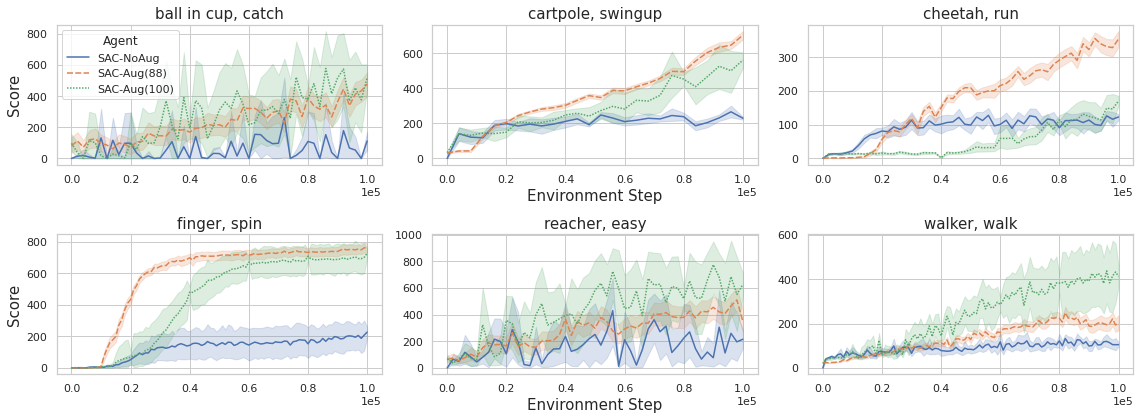}
  \caption{Step-reward curve of SAC variants with different image augmentations}
  \label{fig:full3}
\end{figure}

\begin{figure}[ht]
  \centering
  \includegraphics[width=0.99\linewidth]{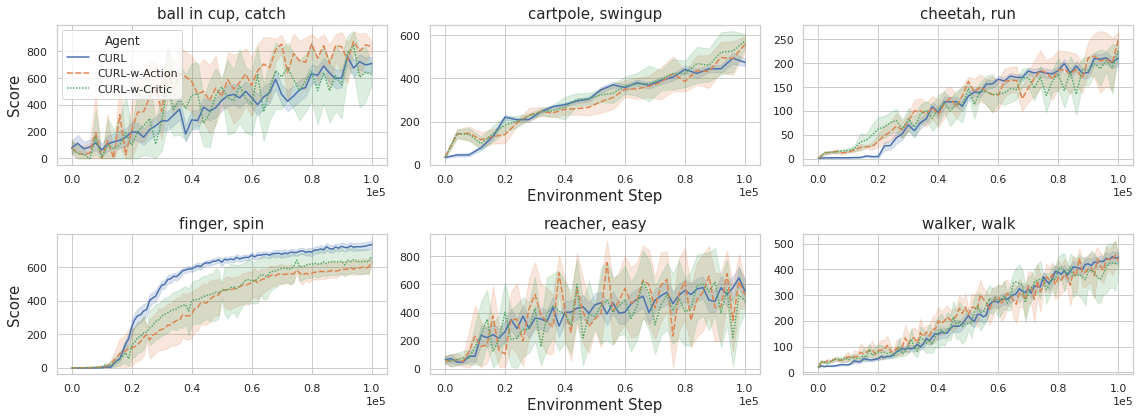}
  \caption{Step-reward curve of CURL and its variants}
  \label{fig:full4}
\end{figure}

\begin{figure}[ht]
  \centering
  \includegraphics[width=0.99\linewidth]{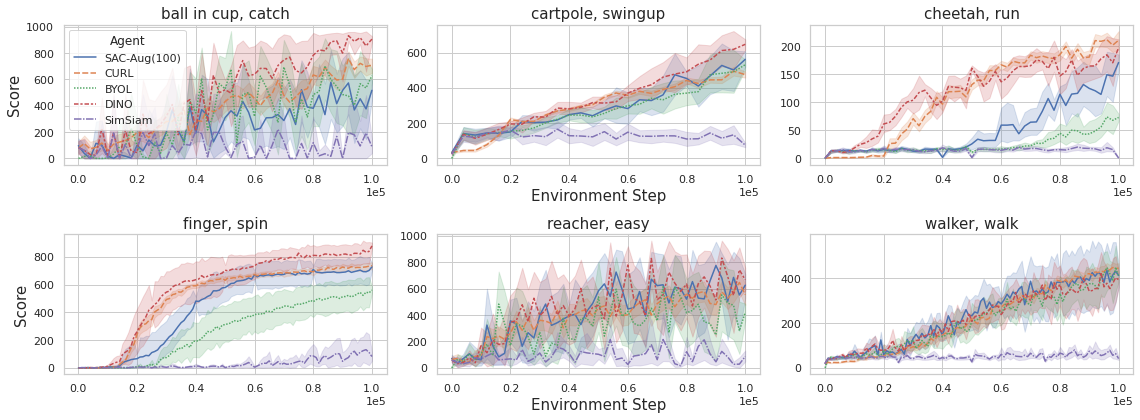}
  \caption{Step-reward curve of self-supervised learning based methods}
  \label{fig:full5}
\end{figure}

\begin{figure}[ht]
  \centering
  \includegraphics[width=0.99\linewidth]{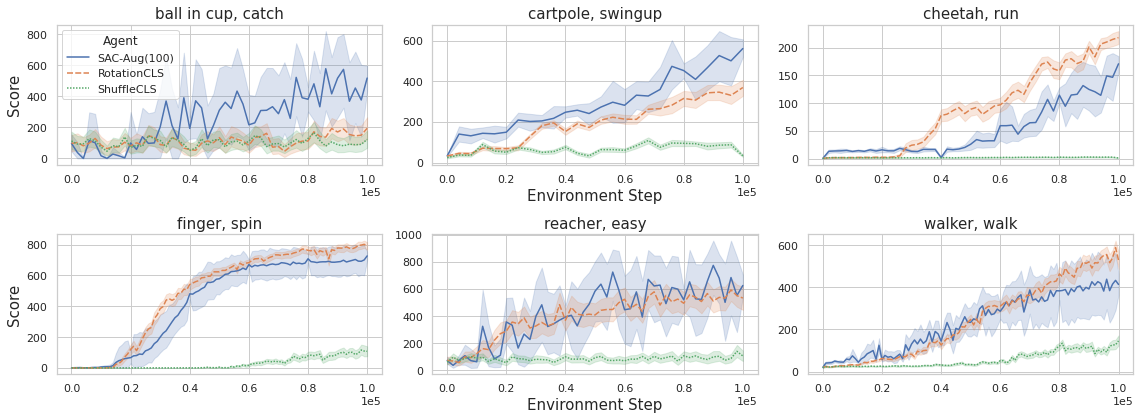}
  \caption{Step-reward curve of classification-based methods (transformation awareness)}
  \label{fig:full6}
\end{figure}

\begin{figure}[ht]
  \centering
  \includegraphics[width=0.99\linewidth]{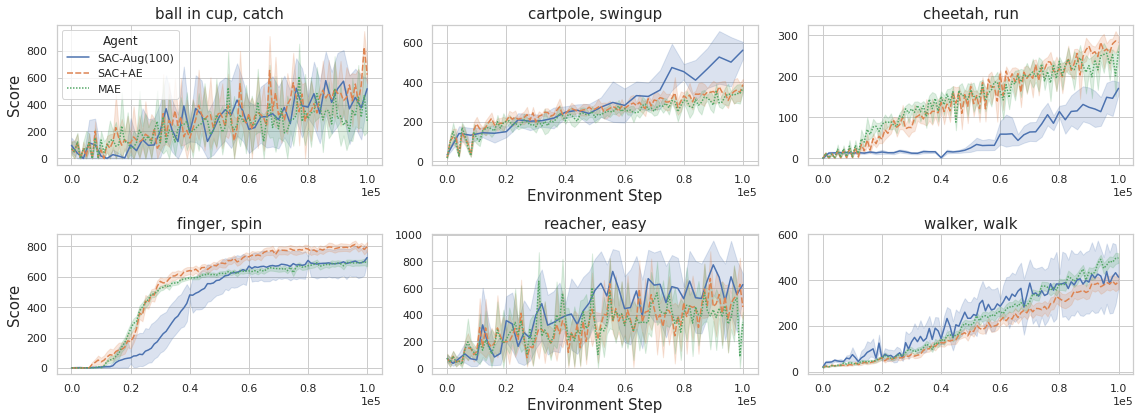}
  \caption{Step-reward curve of reconstruction methods}
  \label{fig:full7}
\end{figure}

\begin{figure}[ht]
  \centering
  \includegraphics[width=0.99\linewidth]{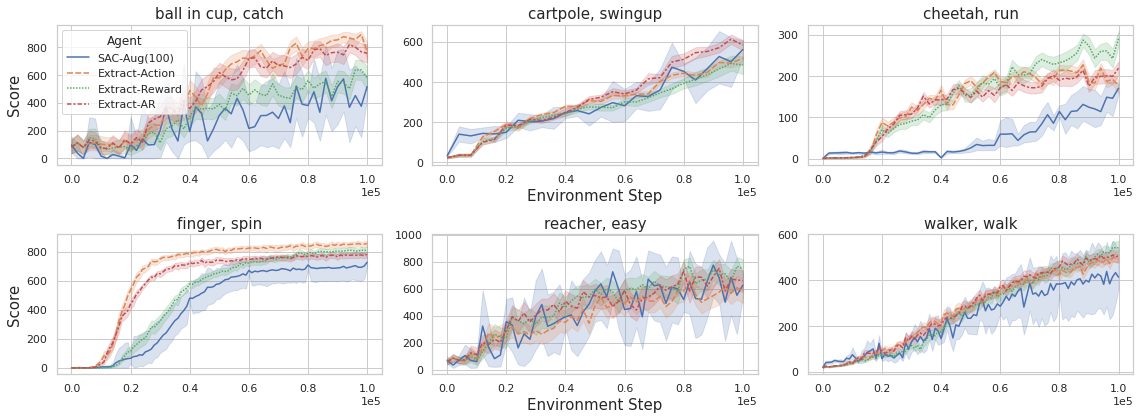}
  \caption{Step-reward curve of RL context prediction methods - 1}
  \label{fig:full8}
\end{figure}

\begin{figure}[ht]
  \centering
  \includegraphics[width=0.99\linewidth]{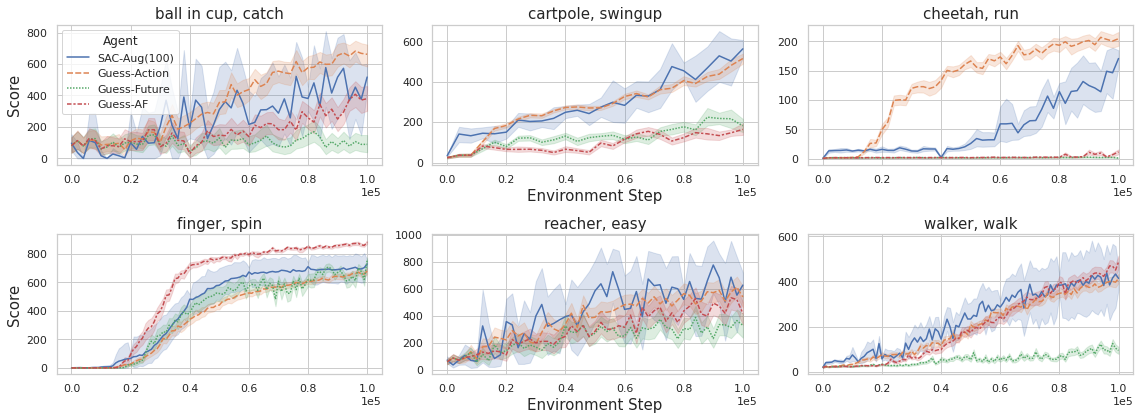}
  \caption{Step-reward curve of RL context prediction methods - 2}
  \label{fig:full9}
\end{figure}

\begin{figure}[ht]
  \centering
  \includegraphics[width=0.99\linewidth]{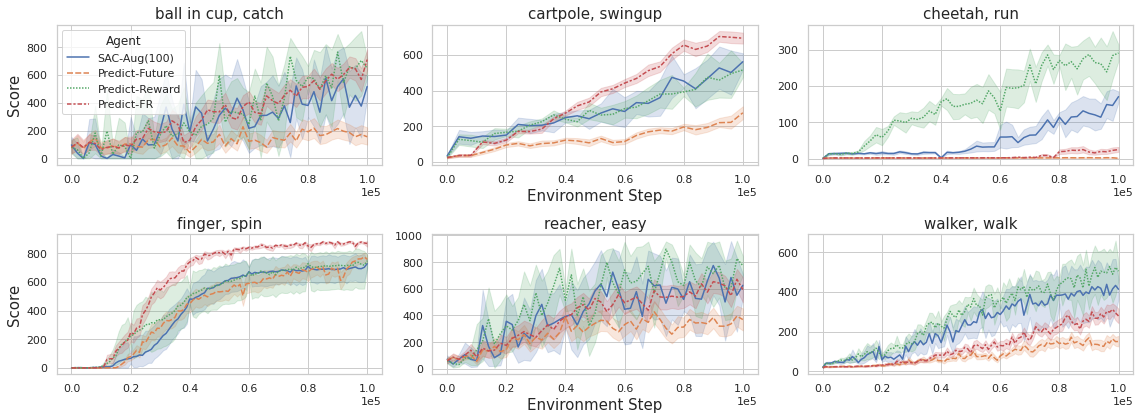}
  \caption{Step-reward curve of RL context prediction methods - 3}
  \label{fig:full10}
\end{figure}

\begin{figure}[ht]
  \centering
  \includegraphics[width=0.99\linewidth]{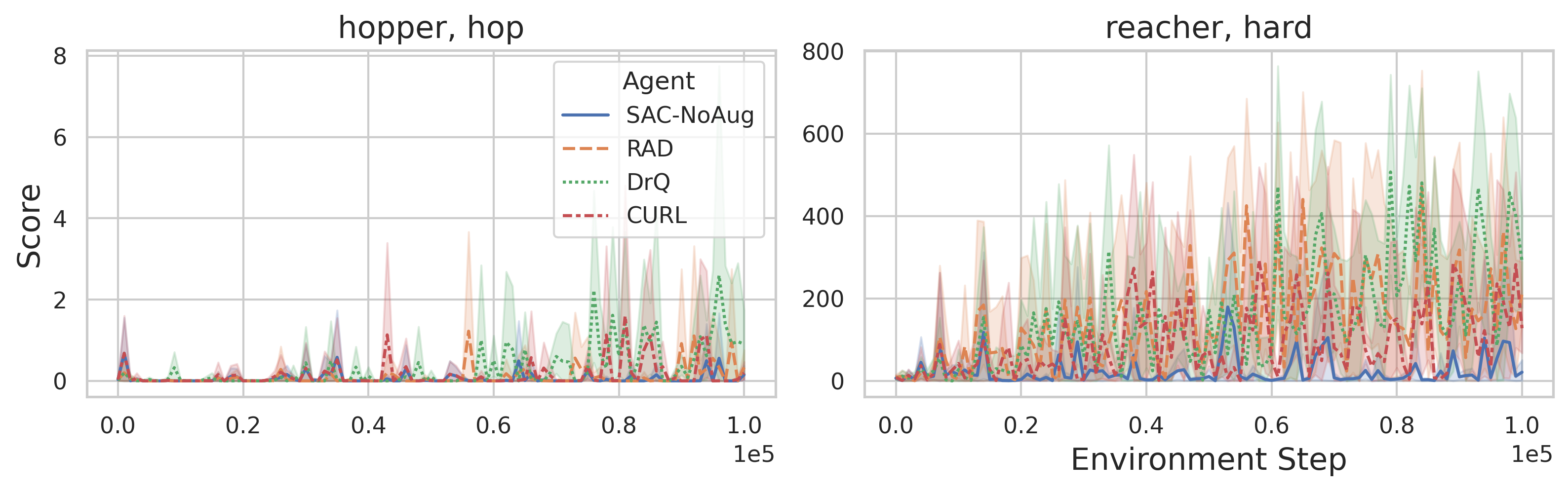}
  \caption{Step-reward curve on two harder DMControl environments - typical methods}
  \label{fig:hard1}
\end{figure}

\begin{figure}[ht]
  \centering
  \includegraphics[width=0.99\linewidth]{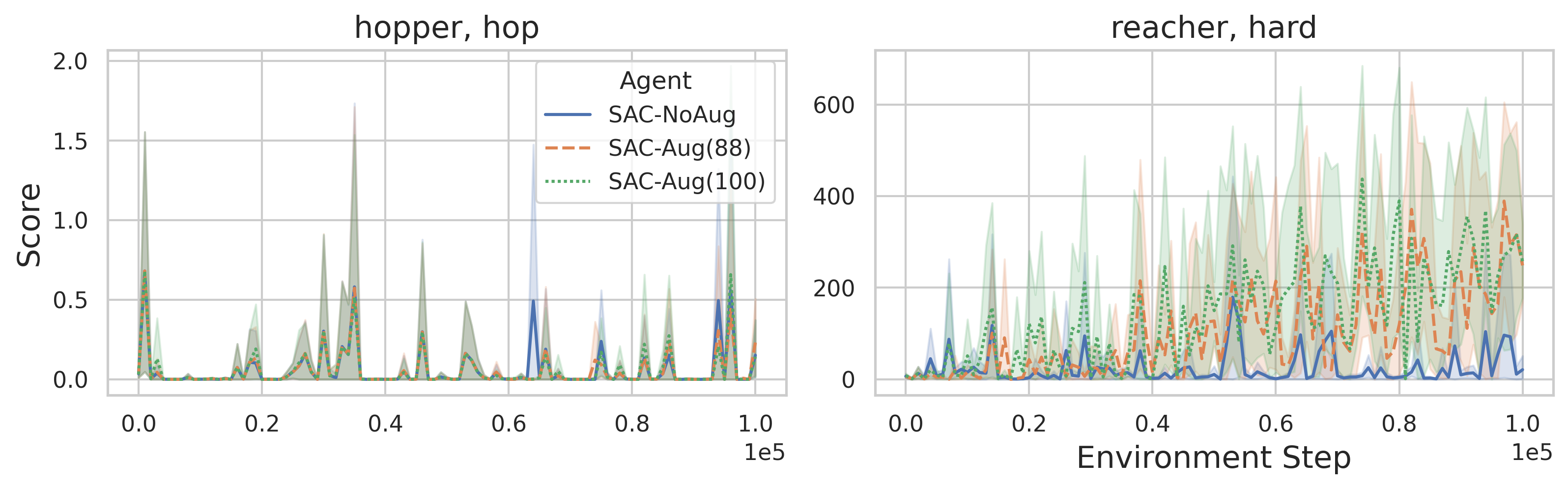}
  \caption{Step-reward curve on two harder DMControl environments - SAC with different image augmentations}
  \label{fig:hard3}
\end{figure}

\begin{figure}[ht]
  \centering
  \includegraphics[width=0.99\linewidth]{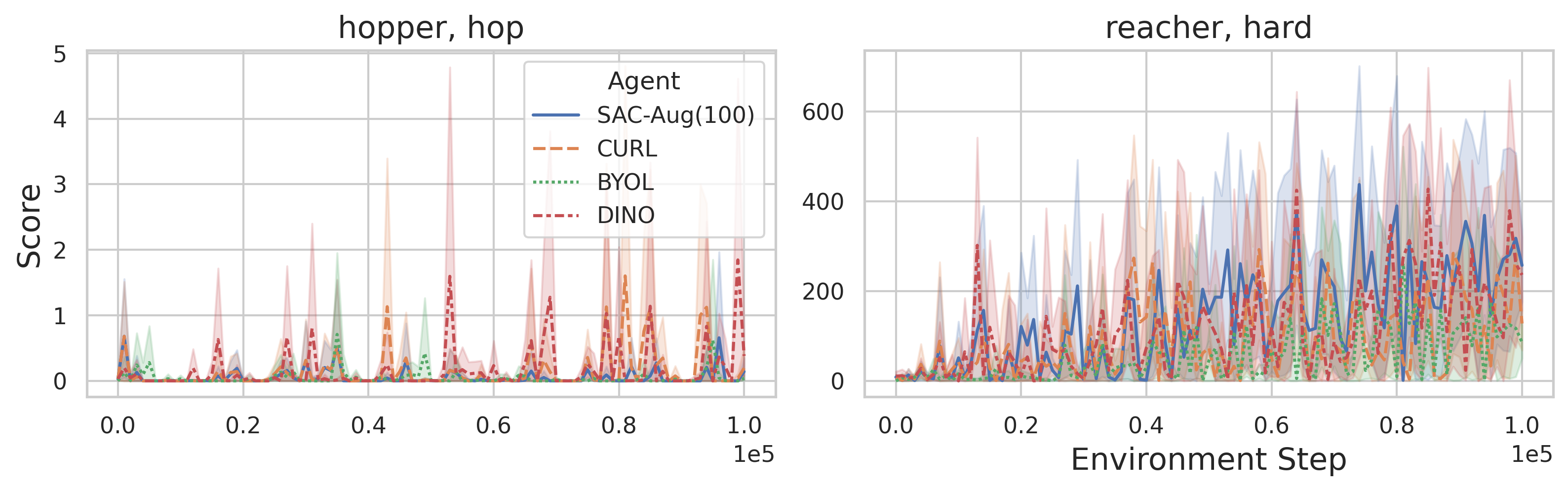}
  \caption{Step-reward curve on two harder DMControl environments - self-supervised learning based methods - 1}
  \label{fig:hard5}
\end{figure}

\begin{figure}[ht]
  \centering
  \includegraphics[width=0.99\linewidth]{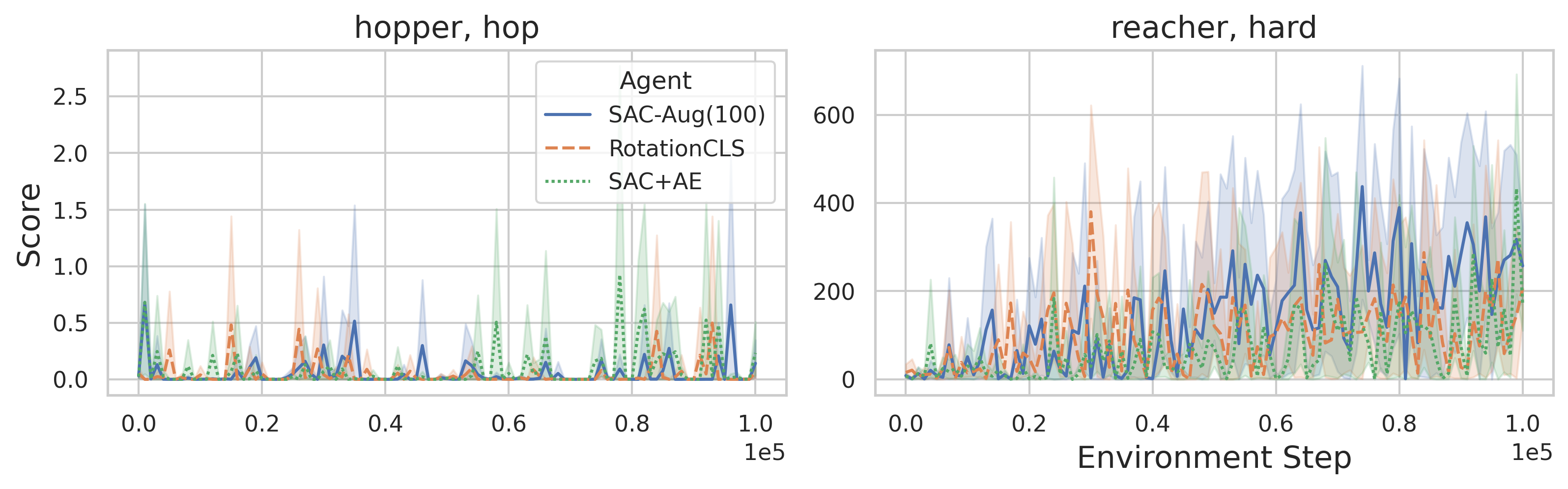}
  \caption{Step-reward curve on two harder DMControl environments - self-supervised learning based methods - 2}
  \label{fig:hard6}
\end{figure}

\begin{figure}[ht]
  \centering
  \includegraphics[width=0.99\linewidth]{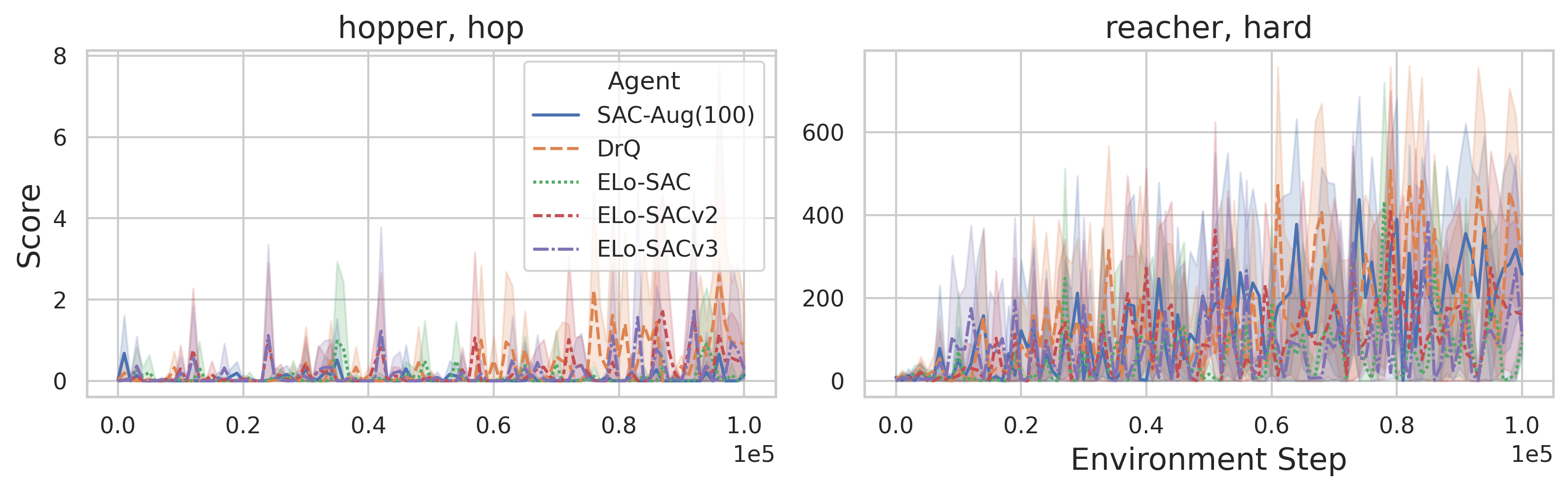}
  \caption{Step-reward curve on two harder DMControl environments - ELo-SAC based methods}
  \label{fig:hard7}
\end{figure}

\paragraph{DMControl}
Table~\ref{tab:DMC} notes the Interquartile mean, standard deviation and Relative Scores of tested algorithms in six DMControl environments.
The score distribution of tested algorithms over six environments is summarized as Figure~\ref{fig:fulldmc}.
Table~\ref{tab:harder} and Figure~\ref{fig:fulldmchard} include results of two additional harder environments in DMControl.
The figures of reward curve v.s. environment step are grouped as Figure~\ref{fig:full1}-\ref{fig:full10} and Figure~\ref{fig:hard1}-\ref{fig:hard7} by the learning method.

\begin{figure}[ht]
  \centering
  \includegraphics[width=\linewidth]{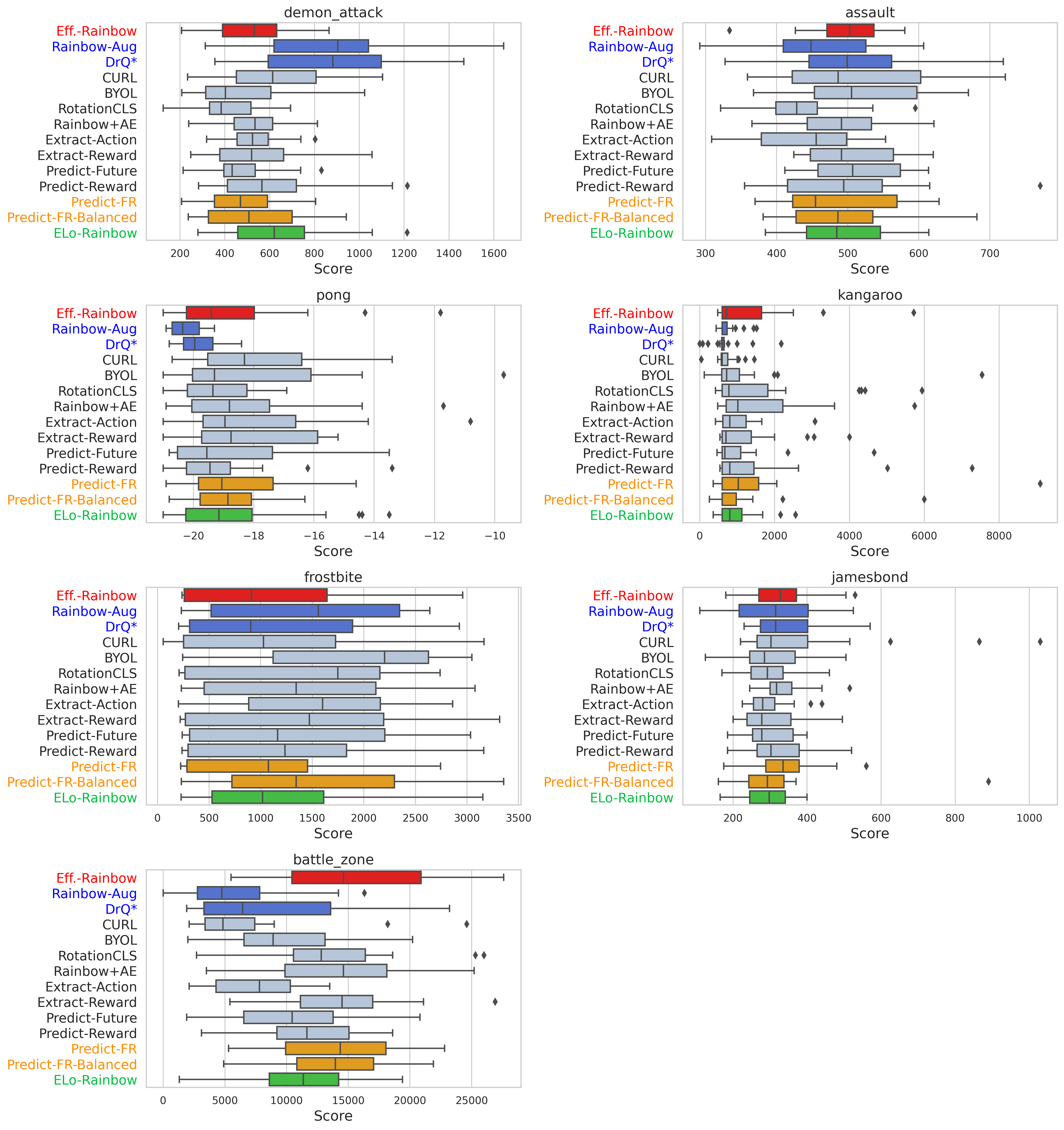}
   \caption{Atari score distribution}
   \label{fig:fullatari}
\end{figure}

\paragraph{Atari}
Similar to Table~\ref{tab:DMC} and Figure~\ref{fig:fulldmc}, Table~\ref{tab:Atari} Figure~\ref{fig:fullatari} are results in seven Atari environments.

\begin{table}[ht]
\caption{Interquartile mean and standard deviation on six DMControl tasks. The last column is colored based on the relative performance w.r.t. SAC-Aug(100), Fig.~\ref{fig:dmcbox}}
\centering
\resizebox{\textwidth}{!}{

\begin{tabular}{c|rcccccc|c}
\toprule
&\textbf{Agent}&ball\_in\_cup,~catch&cartpole,~swingup&cheetah,~run&finger,~spin&reacher,~easy&walker,~walk&Relative~Score\\
\midrule
\parbox[t]{2mm}{\multirow{7}{*}{\rotatebox[origin=c]{90}{No~SSL}}}&\textbf{SAC-NoAug}&71.4 ± 139.9&224.8 ± 28.6&120.9 ± 25.7&238.9 ± 172.6&204.8 ± 131.8&99.6 ± 38.7&\cellcolor{red!29}-8.868\\
&\textbf{SAC-Aug(88)}&510.8 ± 187.4&714.2 ± 113.9&354.5 ± 68.7&771.2 ± 175.0&347.9 ± 148.5&192.2 ± 165.0&\cellcolor{red!02}0.160\\
&\textbf{SAC-Aug(100)}&541.4 ± 306.2&563.4 ± 235.0&172.1 ± 64.0&724.6 ± 154.9&654.4 ± 222.1&422.1 ± 250.8&0.986\\
&\textbf{RAD}&879.9 ± 82.0&786.4 ± 95.1&387.9 ± 81.3&910.4 ± 104.5&508.8 ± 111.5&522.1 ± 95.5&\cellcolor{green!51}5.310\\
\cmidrule{2-9}
&\textbf{DrQ}&914.9 ± 21.2&692.2 ± 222.9&360.4 ± 67.7&935.6 ± 201.3&713.7 ± 147.6&523.9 ± 182.2&\cellcolor{green!60}\textbf{6.028}\\
&\textbf{DrQ(88)}&762.5 ± 139.4&508.2 ± 161.2&331.7 ± 80.5&877.6 ± 93.2&395.5 ± 161.0&119.2 ± 160.5&\cellcolor{red!02}0.154\\
&\textbf{DrQ(100)}&907.6 ± 102.9&675.5 ± 131.1&318.8 ± 54.2&\textbf{940.0 ± 127.2}&627.0 ± 233.0&302.9 ± 295.8&\cellcolor{green!34}3.898\\
\midrule
\parbox[t]{2mm}{\multirow{23}{*}{\rotatebox[origin=c]{90}{Self-supervised}}}&\textbf{CURL}&730.0 ± 179.4&471.5 ± 89.9&215.1 ± 57.3&717.8 ± 136.5&569.8 ± 179.4&442.6 ± 87.1&\cellcolor{green!01}1.128\\
&\textbf{CURL-w-Action}&888.4 ± 179.5&537.8 ± 189.9&247.7 ± 72.7&604.2 ± 79.3&521.3 ± 211.0&439.9 ± 67.8&\cellcolor{green!05}1.452\\
&\textbf{CURL-w-Critic}&690.9 ± 328.8&603.8 ± 156.4&233.7 ± 44.3&657.0 ± 127.1&536.3 ± 208.8&443.0 ± 157.9&\cellcolor{green!03}1.320\\
&\textbf{BYOL}&667.7 ± 281.2&507.2 ± 221.7&70.7 ± 44.3&547.3 ± 185.6&403.7 ± 183.7&449.0 ± 153.5&\cellcolor{red!07}-1.594\\
&\textbf{DINO}&\textbf{916.9 ± 65.7}&686.0 ± 152.2&198.3 ± 79.3&923.1 ± 124.4&686.2 ± 198.2&414.6 ± 162.4&\cellcolor{green!35}3.957\\
&\textbf{SimSiam}&82.6 ± 86.7&67.4 ± 68.6&0.7 ± 0.3&7.6 ± 179.4&72.3 ± 71.1&34.1 ± 24.0&\cellcolor{red!40}-12.537\\
\cmidrule{2-9}
&\textbf{RotationCLS}&157.9 ± 212.1&336.4 ± 220.1&209.7 ± 44.7&801.9 ± 139.7&540.3 ± 163.7&537.0 ± 170.3&\cellcolor{red!05}-0.718\\
&\textbf{ShuffleCLS}&112.2 ± 101.9&28.8 ± 28.4&0.9 ± 0.4&53.0 ± 162.8&108.3 ± 55.4&127.3 ± 98.9&\cellcolor{red!37}-11.701\\
\cmidrule{2-9}
&\textbf{SAC+AE}&616.1 ± 169.9&388.8 ± 130.1&291.8 ± 59.8&799.0 ± 138.9&481.3 ± 130.4&402.6 ± 161.5&\cellcolor{red!01}0.566\\
&\textbf{MAE}&251.1 ± 231.1&372.8 ± 76.1&282.0 ± 62.3&669.5 ± 112.8&336.9 ± 170.1&489.7 ± 49.4&\cellcolor{red!07}-1.635\\
\cmidrule{2-9}
&\textbf{Extract-Action}&871.0 ± 298.6&493.9 ± 162.7&172.3 ± 65.5&870.4 ± 108.1&578.3 ± 144.4&484.8 ± 70.5&\cellcolor{green!15}2.297\\
&\textbf{Extract-Reward}&598.2 ± 306.2&469.8 ± 218.7&302.1 ± 89.9&828.7 ± 115.3&753.2 ± 155.5&522.2 ± 130.5&\cellcolor{green!27}3.266\\
&\textbf{Guess-Action}&724.6 ± 265.3&495.7 ± 121.4&204.6 ± 26.2&669.9 ± 116.8&578.8 ± 161.1&410.6 ± 91.1&\cellcolor{red!00}0.813\\
&\textbf{Guess-Future}&82.4 ± 87.1&146.6 ± 178.0&0.7 ± 0.4&786.5 ± 117.8&323.4 ± 229.2&74.1 ± 73.6&\cellcolor{red!24}-7.318\\
&\textbf{Predict-Future}&121.5 ± 186.9&252.7 ± 219.9&0.7 ± 0.3&796.7 ± 166.7&365.3 ± 235.2&112.7 ± 137.4&\cellcolor{red!21}-6.201\\
&\textbf{Predict-Reward}&672.8 ± 260.3&517.8 ± 215.6&279.1 ± 71.9&837.6 ± 264.6&\textbf{796.2 ± 143.5}&520.1 ± 218.1&\cellcolor{green!33}3.826\\
\cmidrule{2-9}
&\textbf{Extract-AR}&822.2 ± 240.5&592.9 ± 124.7&225.8 ± 60.7&783.0 ± 112.0&678.7 ± 181.3&458.4 ± 148.9&\cellcolor{green!24}3.042\\
&\textbf{Guess-AF}&329.8 ± 298.4&140.7 ± 144.0&0.9 ± 22.8&880.0 ± 59.5&382.9 ± 265.0&494.7 ± 112.7&\cellcolor{red!13}-3.421\\
&\textbf{Predict-FR}&750.3 ± 256.0&723.2 ± 167.5&12.4 ± 35.7&861.5 ± 49.2&636.1 ± 201.4&270.0 ± 154.9&\cellcolor{red!00}0.821\\
\cmidrule{2-9}
&\textbf{ELo-SAC}&831.3 ± 76.2&\textbf{798.7 ± 44.4}&354.0 ± 68.9&835.7 ± 151.2&485.2 ± 171.5&532.1 ± 160.7&\cellcolor{green!42}4.567\\
&\textbf{ELo-SACv2}&864.6 ± 97.0&679.8 ± 104.7&\textbf{414.0 ± 59.8}&844.0 ± 166.4&513.9 ± 95.5&555.4 ± 163.7&\cellcolor{green!46}4.901\\
&\textbf{ELo-SACv3}&851.0 ± 143.5&612.6 ± 87.7&313.9 ± 74.6&914.7 ± 143.4&625.2 ± 94.5&\textbf{697.4 ± 238.1}&\cellcolor{green!53}5.502\\
\bottomrule
\end{tabular}

\label{tab:DMC}

}
\end{table}

\begin{table}[ht]
\caption{Scores on two harder tasks in DMControl}
\label{tab:harder}
\centering

\begin{tabular}{c|rcc|c}
\toprule
&\textbf{Agent}&hopper,~hop&reacher,~hard&Relative~Score\\
\midrule
\parbox[t]{2mm}{\multirow{5}{*}{\rotatebox[origin=c]{90}{No~SSL}}}&\textbf{SAC-NoAug}&0.033 ± 0.4&3.1 ± 39.7&\cellcolor{red!40}-2.543\\
&\textbf{SAC-Aug(88)}&0.048 ± 0.4&210.733 ± 190.9&\cellcolor{red!05}0.191\\
&\textbf{SAC-Aug(100)}&0.024 ± 0.4&\textbf{262.4 ± 140.4}&0.634\\
&\textbf{RAD}&0.038 ± 0.9&193.1 ± 186.1&\cellcolor{red!09}-0.112\\
&\textbf{DrQ}&\textbf{0.424 ± 1.4}&258.95 ± 205.6&\cellcolor{green!60}\textbf{4.017}\\
\midrule
\parbox[t]{2mm}{\multirow{9}{*}{\rotatebox[origin=c]{90}{Self-sup.}}}&\textbf{CURL}&0.076 ± 0.4&115.5 ± 116.4&\cellcolor{red!17}-0.765\\
&\textbf{BYOL}&0.025 ± 0.1&49.725 ± 126.1&\cellcolor{red!33}-2.025\\
&\textbf{DINO}&0.25 ± 0.5&200.333 ± 178.9&\cellcolor{green!20}1.789\\
&\textbf{RotationCLS}&0.031 ± 0.1&210.05 ± 117.3&\cellcolor{red!07}0.036\\
&\textbf{SAC+AE}&0.061 ± 0.4&140.567 ± 185.1&\cellcolor{red!15}-0.579\\
\cmidrule{2-5}
&\textbf{ELo-SAC}&0.116 ± 0.3&81.592 ± 53.7&\cellcolor{red!18}-0.845\\
&\textbf{ELo-SACv2}&0.147 ± 0.6&152.858 ± 70.8&\cellcolor{red!04}0.314\\
&\textbf{ELo-SACv3}&0.177 ± 0.4&98.208 ± 78.7&\cellcolor{red!09}-0.112\\
\bottomrule
\end{tabular}

\end{table}

\begin{table}[ht]
\caption{Scores on Atari, the last column is colored based on the relative performance w.r.t. Efficient Rainbow, ``*'' means using a different image augmentation method from the original paper}
\centering
\resizebox{\textwidth}{!}{

\begin{tabular}{c|rccccccc|c}
\toprule
 & \textbf{Agent              } &        assault &       battle\_zone &   demon\_attack &        frostbite &      jamesbond &         kangaroo &         pong &  Relative Score \\
\midrule
\parbox[t]{2mm}{\multirow{3}{*}{\rotatebox[origin=c]{90}{No SSL}}} & \textbf{Eff.-Rainbow       }  &    506.8  ±  59.3  & \textbf{  14840.0  ±  6681.7 } &   519.3  ±  193.1  &     873.1  ±  834.8  &    318.5  ±  92.7  &    853.0  ±  1304.8  &   -19.0  ±  2.4  &            2.065 \\
 & \textbf{Rainbow-Aug        }  &    459.7  ±  79.6  &    4770.0  ±  4379.0  &   870.3  ±  345.9  &    1469.7  ±  962.2  &   317.0  ±  110.5  &     619.0  ±  298.0  &   -20.3  ±  0.5  & \cellcolor{red!36}          -2.223 \\
\cmidrule{2-10}
 & \textbf{DrQ*               }  &    503.7  ±  89.0  &    7600.0  ±  6839.0  & \textbf{  891.2  ±  322.3 } &     943.7  ±  913.2  & \textbf{   321.0  ±  91.6 } &     605.0  ±  462.0  &   -19.9  ±  0.8  & \cellcolor{red!19}          -0.290 \\
\midrule
\parbox[t]{2mm}{\multirow{11}{*}{\rotatebox[origin=c]{90}{Self-supervised}}} & \textbf{CURL               }  &   511.6  ±  107.3  &    5100.0  ±  5530.2  &   615.3  ±  240.4  &    928.3  ±  1018.5  &   307.0  ±  219.8  &     620.0  ±  300.8  & \textbf{  -18.1  ±  2.3 } & \cellcolor{red!21}          -0.516 \\
 & \textbf{BYOL               }  & \textbf{   514.6  ±  93.4 } &    9470.0  ±  4879.6  &   418.4  ±  246.5  & \textbf{   2111.5  ±  982.6 } &    291.5  ±  90.9  &    740.0  ±  1573.6  &   -18.5  ±  2.9  & \cellcolor{red!01}           1.877 \\
\cmidrule{2-10}
 & \textbf{RotationCLS        }  &    427.1  ±  62.2  &   12950.0  ±  5742.7  &   401.0  ±  159.0  &    1591.9  ±  949.5  &    285.5  ±  70.2  &    892.0  ±  1674.2  &   -19.3  ±  1.3  & \cellcolor{red!40}          -2.647 \\
\cmidrule{2-10}
 & \textbf{Rainbow+AE         }  &    485.2  ±  74.7  &   14290.0  ±  5927.7  &   528.8  ±  158.6  &    1272.5  ±  964.3  &    320.5  ±  68.8  & \textbf{  1155.0  ±  1392.5 } &   -18.8  ±  2.3  & \cellcolor{green!60}\textbf{           4.815} \\
\cmidrule{2-10}
 & \textbf{Extract-Action     }  &    443.6  ±  72.8  &    7370.0  ±  3797.5  &   521.0  ±  126.6  &    1627.4  ±  874.5  &    282.0  ±  56.1  &     855.0  ±  612.3  &   -18.7  ±  2.5  & \cellcolor{red!36}          -2.237 \\
 & \textbf{Extract-Reward     }  &    494.8  ±  63.7  &   14420.0  ±  4901.0  &   533.4  ±  224.1  &   1286.6  ±  1109.0  &    294.5  ±  83.7  &    804.0  ±  1001.0  &   -18.4  ±  2.1  & \cellcolor{red!03}           1.692 \\
 & \textbf{Predict-Future     }  &    509.5  ±  67.7  &   10420.0  ±  5252.4  &   452.1  ±  145.6  &    1144.5  ±  988.7  &    295.0  ±  70.7  &     733.0  ±  966.0  &   -19.4  ±  2.1  & \cellcolor{red!32}          -1.796 \\
 & \textbf{Predict-Reward     }  &   485.6  ±  100.8  &   11870.0  ±  4197.2  &   547.9  ±  291.6  &    1155.9  ±  946.9  &    304.0  ±  92.7  &    908.0  ±  1718.9  &   -19.4  ±  1.7  & \cellcolor{red!16}           0.135 \\
 & \textbf{Predict-FR-Balanced }  &    485.7  ±  82.2  &   14270.0  ±  4421.5  &   495.8  ±  209.7  &   1359.1  ±  1029.2  &   293.5  ±  146.8  &    664.0  ±  1239.6  &   -18.9  ±  1.3  & \cellcolor{red!21}          -0.508 \\
\cmidrule{2-10}
 & \textbf{ELo-Rainbow        }  &    493.1  ±  67.4  &   11750.0  ±  4727.9  &   623.4  ±  249.9  &    1027.6  ±  863.8  &    297.5  ±  66.4  &     795.0  ±  593.3  &   -19.2  ±  2.3  & \cellcolor{red!20}          -0.369 \\
\bottomrule
\end{tabular}

\label{tab:Atari}

}
\end{table}

\subsection{Real-world Robot Experiments}
\label{sec:robot}
To further evaluate methods in the real world applications, we set up a continuous robot arm control environment, uArm reacher.
With the help of some simple techniques in computer vision and robotics, our environment can autonomously randomly reset and keep the agent training without any human input.

The environment requires a robotic arm with a suction cup actuator, two fixed RGB cameras, and a cube that can be picked up by the suction cup as the target, as shown in Fig.~\ref{fig:robotobs}.
The goal is to move the actuator close to the target as fast as possible.
The observation comes from two cameras with a native resolution of $640\times 480$.
The images are then resized to $100 \times 100$, stacked along channel axis, and finally randomly cropped into $84 \times 84$, resulting in an $84\times84\times(3+3)$ image observation before fed to the network.
The action space is a 3D vector ranging from -1 to 1, and it will be mapped to the actuator position movement in a 3D robot Cartesian coordinates whose original point is the center of the robot base.
The robot's motion range is manually limited for safety concerns while avoiding the actuator moving the target in one episode.
Following reacher in DMControl, we define a very simple reward function.
The reward function returns $1$ when the 3D Euclidean distance between the actuator and the target is lower than a threshold, otherwise, it returns $-1e-3$.
The length of each episode is set to 200 steps, which limits the range of the episode accumulated reward to $[-0.2, 200]$. 

To enable automatic reward generation, we make an automatic calibration framework to get the target location in 3D, and calibrate the top-down camera before any experiments.
We use AprilTag~\cite{olson2011tags} to locate the robot position in the image plane, and read the 3D robot coordinates directly from the robot.
By doing so, we can build a map between 2D image coordinates and 3D robot coordinates.
The 2D coordinates of the target is first extracted by a simple color threshold.
Then, given the constant height of the target, we can obtain 3D target location from its 2D image coordinates according to the 2D$\leftrightarrow$3D map.

The environmental \textit{reset} process is also automatic.
At the beginning of each episode, the robot arm will pick up the target cube, and randomly release the cube at a certain height like throwing dice, in order to randomly initialize the cube location.
The new location of the target cube is saved for generating rewards.
After the robot arm automatically moves to a fixed pre-assigned starting point, the environmental reset is done and then the RL agent takes over the control.
The RL agent can perform regular online training until the episode ends.
Finally, after each episode ends, the environment repeats the reset process to initialize the next episode.




\subsection{Empirical Analysis on the Learned Representations}
\label{sec:analysis}
To further understand the role of self-supervised loss and image augmentation in an online reinforcement learning system with the joint learning framework, we empirically show the properties of representations learned by different losses.

We first follow \citet{wang2022investigating} and measure the three metrics Dynamic Awareness, Diversity, and Orthogonality, extending them from discrete action space to continuous action space.

\textbf{Dynamics Awareness} means two states that are adjacent in time should have similar representations and states further apart should have a low similarity. 

\textbf{Diversity} measures a ratio between state and state-value differences. 
High diversity means two states have two different representations to be distinguished even when they have similar state values.

\textbf{Orthogonality} reflects the linear independence of the representation, in another word, the higher the orthogonality, the lower the redundancy in the representations.

Assume an image observation $x_i$ is taken when the intrinsic system state is $s_i$.
Denoting the visual representation of $x_i$ generated by the encoder from the critic networks as $\phi_i$, and  $\text{Critic}(\phi_i, \cdot)$ is the learned critic network output.
Eq.~\ref{eq:repdist} shows how to compute the three representation metrics.

\begin{equation}
\begin{split}
  \text{Dynamic Awareness} &= \frac{\sum_i^N\left\Vert\phi_i-\phi_{j\sim U(1, N)}\right\Vert_2 - \sum_i^N\left\Vert\phi_i-\phi_i'\right\Vert_2}{\sum_i^N\left\Vert\phi_i-\phi_{j\sim U(1, N)}\right\Vert_2} \\
  \text{Diversity} &= 1 - \frac{1}{N^2}\sum_{i, j}^N\min\left(\frac{d_{v,i,j} / \max_{i,j}d_{v,i,j}}{d_{s,i,j} / \max_{i,j}d_{s,i,j} + 10^{-2}}, 1 \right) \\
  \text{Orthogonality} &= 1 - \frac{2}{N(N-1)}\sum_{i,j,i<j}^N{\frac{\lvert \left<\phi_i, \phi_j\right> \rvert}{\left\Vert\phi_i\right\Vert_2\left\Vert\phi_j\right\Vert_2}}
  \end{split}
  \label{eq:repdist}
\end{equation}
where $N$ is the total number of samples, $U(1,N)$ means uniformly sample from $[1, N]$, $d_{s,i,j} = \left\Vert \phi_i - \phi_j \right\Vert_2$ and $d_{v,i,j} = \lvert\max_a{\text{Critic}(\phi_i, a)} - \max_a{\text{Critic}(\phi_j, a)}\rvert$.

\textbf{Predict State from Visual Representation} Besides the three metrics on visual representations and state-values, we further measure the quality of visual representation $\phi_i$ by predicting the system state $s_i$ only using $\phi_i$.
The intuition is that a better visual representation should be able to capture the intrinsic system state more precisely.
We utilize a two-layer MLP to regress the system state $s_i$ on its corresponding visual representation $\phi_i$.
Mean squared error is applied to supervise the network as well as to evaluate the network on the test set.

To properly measure all these metrics, We first collect a dataset in cartpole swingup from DMControl using state-based SAC, which is different from any methods we'll benchmark to avoid bias.
We run state-based SAC with five random seeds, and take the replay buffer of each run to form a dataset.
The whole dataset has $12500 \times 5 = 62500$ state transitions.
We measure Dynamics Awareness and Orthogonality on the full dataset, while Diversity is calculated for one run due to computational cost.
For state prediction, we use the first four runs as the training set and the last run is held for the evaluation.

\begin{figure}[ht]
  \centering
  \includegraphics[width=0.99\linewidth]{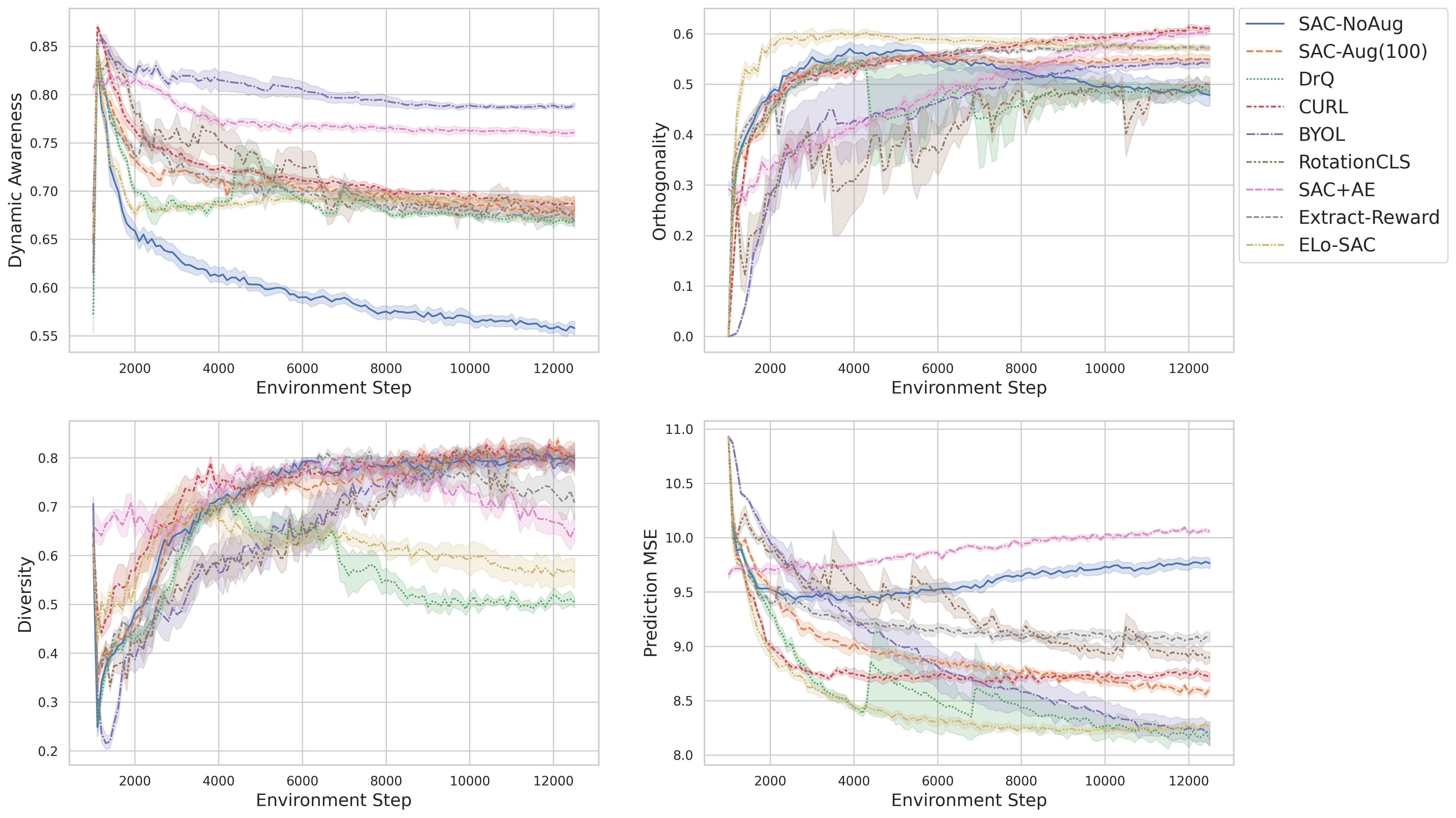}
   \caption{Scores versus representation metric values}
   \label{fig:metric_step}
\end{figure}

\begin{figure}[ht]
  \centering
  \includegraphics[width=0.99\linewidth]{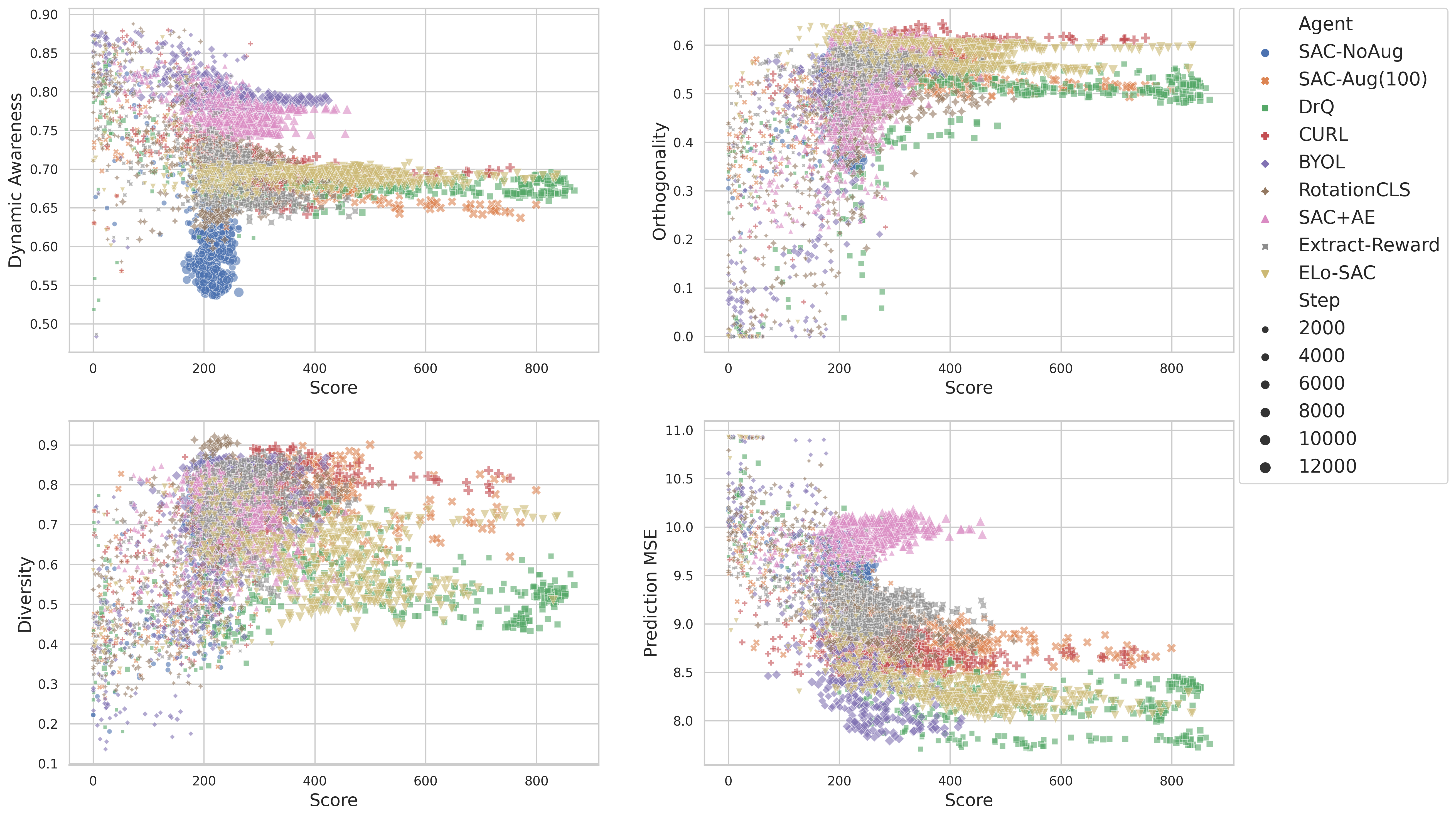}
   \caption{Scores versus representation metric values}
   \label{fig:metric_reward}
\end{figure}

Finally, we benchmark selected methods with five different random seeds on cartpole swingup, and reports the metrics above every $100$ model update step.
We demonstrate how the four metrics correlate to the environment step and agent performance as Fig.~\ref{fig:metric_step} and Fig.~\ref{fig:metric_reward}.


Fig.~\ref{fig:metric_step} shows how metrics change as training. 
Most of the methods converge on a similar Orthogonality, Dynamic Awareness, and Diversity value.
SAC-NoAug has a low Dynamic Awareness measure which could be used to explain its low performance.
While a higher Dynamic Awareness measure does not bring extra scores for BYOL and SAC+AE.
Similarly, the lower Diversity value of DrQ and ELo-SAC do not hurt their performance either.
Meanwhile, most of the metrics become relatively stable after the first 4000 steps.
Therefore, we confirm that the shallower layers of the neural networks in visual reinforcement learning converge faster as observed by \citet{chen2021improving}.

Fig.~\ref{fig:metric_reward} shows the correlation between metrics and the agent performance.
We report the Pearson correlation coefficient as Table~\ref{tab:pearson}.
As \citet{wang2022investigating} suggested, these metrics only measure certain properties of the visual representation, and they do not suggest that a property is necessary for better policy learning.
However, we find that the state prediction error is correlated to the agent performance to some extent, which may be valuable in some cases.

\begin{table}[ht]
\caption{Pearson correlation coefficient between scores and representation metrics}
\label{tab:pearson}
\centering

\begin{tabular}{cccc}
\toprule
 Dynamic Awareness&  Orthogonality&   Diversity&    Prediction MSE\\
\midrule
-0.284 & 0.435 & 0.111 & -0.625 \\
\bottomrule
\end{tabular}
\end{table}


\subsection{Observation on Pretraining Framework}
\label{sec:pretraining}

Besides the joint learning framework used in CURL and SAC+AE, \citet{shelhamer2016loss} investigate a pretraining framework to combine SSL with RL and use the self-supervised loss as an intrinsic reward to further boost performance during online learning.
Recent works on policy learning (e.g., \cite{zhan2020framework, shah2021rrl, wang2022vrl3, xiao2022masked, parisi2022unsurprising}) also take advantage of the self-supervised learning in a multi-step framework and show its great potential in solving challenging visual-based problems.

This pretraining framework is similar to how self-supervision has been benefiting supervised Computer Vision tasks~(\cite{chen2020simple, he2020momentum, caron2021emerging, chen2021exploring, kahatapitiya2021self, das2021viewclr, piergiovanni2019evolving}): pretrain with self-supervised losses, and then finetune with the downstream task loss. 
Motivated by them, in this section, we design and benchmark the two-stage pretraining framework, replacing the joint learning framework used in CURL and SAC+AE.

\begin{figure}[ht]
  \centering
  \includegraphics[width=0.99\linewidth]{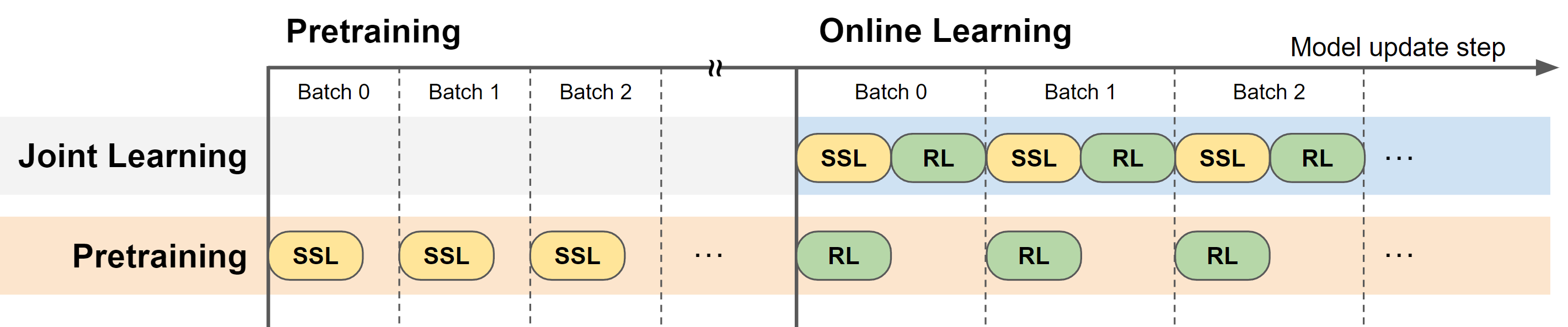}
   \caption{Two learning frameworks for SSL + RL, the rounded rectangle means to update the model with the labeled loss for one step.}
   \label{fig:framework}
\end{figure}

In the first stage, we use data collected by training a SAC-Aug(100) agent on the same task and update the visual encoder only using self-supervised loss.
We name this stage pretraining which means to use self-supervised losses to update the model and to be downstream task agnostic.
Then in the second stage i.e., the online training stage, we only keep the trained encoder from the first stage and train an agent using SAC-Aug(100).
The only difference between this stage and training an agent from scratch is that here the visual encoder has been ``initialized'' with the pre-trained weights while it is randomly initialized in SAC-Aug(100).
This also means that the image encoder can be tuned by RL loss in the online training stage to match the online sample distribution.
Fig.~\ref{fig:framework} compares two training frameworks, in which the rounded rectangle means to update the model with the labeled loss for one step.

\begin{figure}[ht]
\centering
\includegraphics[width=0.9\linewidth]{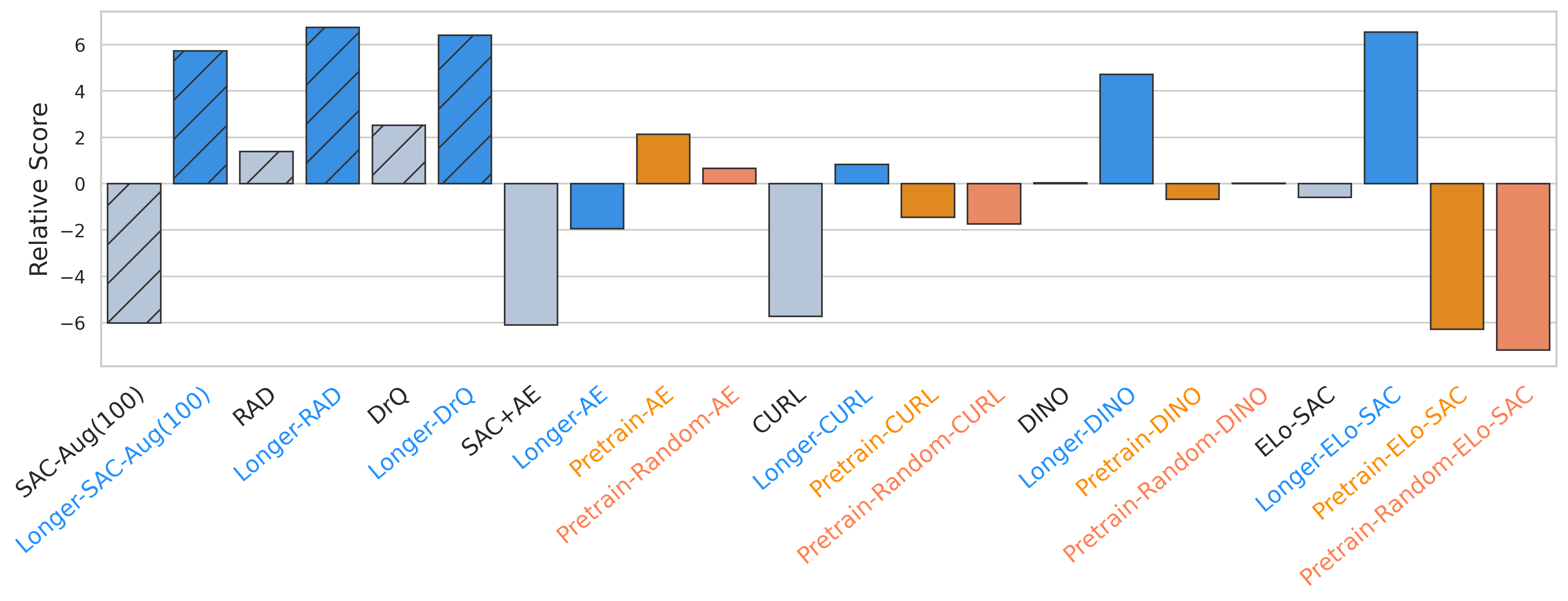}
\caption{Relative Score of two learning frameworks for combining SSL to RL. The bars with diagonal lines stand for the methods that only use image augmentation without any self-supervised losses.}
\label{fig:pretrainrs}
\end{figure}

The methods using the pretraining framework have the prefix `Pretrain'.
`Pretrain-Random' means the data used for pretraining is collected by a random policy.
In both cases, the pretraining framework has the same model update steps as the joint learning framework baseline.
But note that the pretraining model has access to extra data collected by other policies, which makes it an unfair comparison.
To this end, we test another joint learning configuration named with the prefix `Longer'.
Here we match the total number of environment steps (or collected data) to its pretraining variants.
Similarly, three methods without any self-supervised learning are benchmarked with 'Longer' configuration.
We compare two frameworks in six DMControl environments, Relative Scores are reported as Fig.~\ref{fig:pretrainrs} and the full results are shown as Table~\ref{tab:pretrain}.


\begin{table}[ht]
\caption{Comparison of the two frameworks. Methods in gray are without a self-supervised loss for reference. The total amount of data/environment step at each stage is listed in the second and the third column.}
\label{tab:pretrain}
\centering
\resizebox{\textwidth}{!}{

\begin{tabular}{r|rr|lll}
\toprule
\textbf{Agent}&\thead{Pretraining\\env.step}&\thead{Online\\env.step}&ball\_in\_cup,~catch&cartpole,~swingup&cheetah,~run\\
\midrule
\textcolor{gray}{\textbf{SAC-Aug(100)}}&\textcolor{gray}{0}&\textcolor{gray}{100k}&\textcolor{gray}{541.4 ± 306.2}&\textcolor{gray}{563.4 ± 235.0}&\textcolor{gray}{172.1 ± 64.0}\\
\textcolor{gray}{\textbf{Longer-SAC-Aug(100)}}&\textcolor{gray}{0}&\textcolor{gray}{200k}&\textcolor{gray}{944.9 ± 75.3(\textcolor{mygreen}{↑403.5})}&\textcolor{gray}{851.0 ± 36.4(\textcolor{mygreen}{↑287.6})}&\textcolor{gray}{424.0 ± 66.8(\textcolor{mygreen}{↑251.9})}\\
\midrule
\textcolor{gray}{\textbf{RAD}}&\textcolor{gray}{0}&\textcolor{gray}{100k}&\textcolor{gray}{879.9 ± 82.0}&\textcolor{gray}{786.4 ± 95.1}&\textcolor{gray}{387.9 ± 81.3}\\
\textcolor{gray}{\textbf{Longer-RAD}}&\textcolor{gray}{0}&\textcolor{gray}{200k}&\textcolor{gray}{932.6 ± 52.6(\textcolor{mygreen}{↑52.7})}&\textcolor{gray}{846.2 ± 34.1(\textcolor{mygreen}{↑59.8})}&\textcolor{gray}{551.4 ± 176.1(\textcolor{mygreen}{↑163.5})}\\
\midrule
\textcolor{gray}{\textbf{DrQ}}&\textcolor{gray}{0}&\textcolor{gray}{100k}&\textcolor{gray}{914.9 ± 21.2}&\textcolor{gray}{692.2 ± 222.9}&\textcolor{gray}{360.4 ± 67.7}\\
\textcolor{gray}{\textbf{Longer-DrQ}}&\textcolor{gray}{0}&\textcolor{gray}{200k}&\textcolor{gray}{952.0 ± 301.5(\textcolor{mygreen}{↑37.1})}&\textcolor{gray}{857.4 ± 31.4(\textcolor{mygreen}{↑165.2})}&\textcolor{gray}{475.9 ± 78.7(\textcolor{mygreen}{↑115.5})}\\
\midrule
\textbf{SAC+AE}&0&100k&616.1 ± 169.9&388.8 ± 130.1&291.8 ± 59.8\\
\textbf{Longer-AE}&0&200k&579.4 ± 274.0(\textcolor{red}{↓-36.7})&467.1 ± 196.9(\textcolor{mygreen}{↑78.3})&359.0 ± 57.6(\textcolor{mygreen}{↑67.2})\\
\textbf{Pretrain-AE}&100k&100k&914.7 ± 129.0(\textcolor{mygreen}{↑298.6})&759.1 ± 99.3(\textcolor{mygreen}{↑370.3})&419.8 ± 41.1(\textcolor{mygreen}{↑128.0})\\
\textbf{Pretrain-Random-AE}&100k&100k&903.1 ± 219.9(\textcolor{mygreen}{↑287.0})&736.0 ± 97.4(\textcolor{mygreen}{↑347.2})&405.3 ± 55.5(\textcolor{mygreen}{↑113.5})\\
\midrule
\textbf{CURL}&0&100k&730.0 ± 179.4&471.5 ± 89.9&215.1 ± 57.3\\
\textbf{Longer-CURL}&0&200k&935.0 ± 26.5(\textcolor{mygreen}{↑205.0})&776.2 ± 82.2(\textcolor{mygreen}{↑304.7})&307.6 ± 57.3(\textcolor{mygreen}{↑92.5})\\
\textbf{Pretrain-CURL}&100k&100k&921.0 ± 25.5(\textcolor{mygreen}{↑191.0})&705.4 ± 138.3(\textcolor{mygreen}{↑233.9})&213.0 ± 56.7(\textcolor{red}{↓-2.1})\\
\textbf{Pretrain-Random-CURL}&100k&100k&874.5 ± 298.3(\textcolor{mygreen}{↑144.5})&745.3 ± 124.5(\textcolor{mygreen}{↑273.8})&224.3 ± 60.6(\textcolor{mygreen}{↑9.2})\\
\midrule
\textbf{DINO}&0&100k&916.9 ± 65.7&686.0 ± 152.2&198.3 ± 79.3\\
\textbf{Longer-DINO}&0&200k&952.6 ± 48.9(\textcolor{mygreen}{↑35.7})&858.1 ± 21.4(\textcolor{mygreen}{↑172.1})&248.6 ± 49.3(\textcolor{mygreen}{↑50.3})\\
\textbf{Pretrain-DINO}&100k&100k&748.1 ± 164.7(\textcolor{red}{↓-168.8})&759.3 ± 110.8(\textcolor{mygreen}{↑73.3})&344.7 ± 56.5(\textcolor{mygreen}{↑146.4})\\
\textbf{Pretrain-Random-DINO}&100k&100k&904.6 ± 266.6(\textcolor{red}{↓-12.3})&758.6 ± 86.1(\textcolor{mygreen}{↑72.6})&355.6 ± 77.5(\textcolor{mygreen}{↑157.3})\\
\midrule
\textbf{ELo-SAC}&0&100k&888.3 ± 90.6&772.8 ± 167.3&359.7 ± 69.7\\
\textbf{Longer-ELo-SAC}&0&200k&949.5 ± 29.8(\textcolor{mygreen}{↑61.2})&866.6 ± 30.0(\textcolor{mygreen}{↑93.8})&489.6 ± 149.7(\textcolor{mygreen}{↑129.9})\\
\textbf{Pretrain-ELo-SAC}&100k&100k&505.8 ± 301.3(\textcolor{red}{↓-382.5})&617.9 ± 147.1(\textcolor{red}{↓-154.9})&400.2 ± 63.6(\textcolor{mygreen}{↑40.5})\\
\textbf{Pretrain-Random-ELo-SAC}&100k&100k&466.2 ± 200.3(\textcolor{red}{↓-422.1})&519.8 ± 175.4(\textcolor{red}{↓-253.0})&302.9 ± 126.4(\textcolor{red}{↓-56.8})\\
\bottomrule
\end{tabular}

}
\bigskip
\resizebox{\textwidth}{!}{

\begin{tabular}{r|rr|lll}
\toprule
\textbf{Agent}&\thead{Pretraining\\env.step}&\thead{Online\\env.step}&finger,~spin&reacher,~easy&walker,~walk\\
\midrule
\textcolor{gray}{\textbf{SAC-Aug(100)}}&\textcolor{gray}{0}&\textcolor{gray}{100k}&\textcolor{gray}{724.6 ± 154.9}&\textcolor{gray}{654.4 ± 222.1}&\textcolor{gray}{422.1 ± 250.8}\\
\textcolor{gray}{\textbf{Longer-SAC-Aug(100)}}&\textcolor{gray}{0}&\textcolor{gray}{200k}&\textcolor{gray}{868.6 ± 140.8(\textcolor{mygreen}{↑144.0})}&\textcolor{gray}{911.6 ± 92.3(\textcolor{mygreen}{↑257.2})}&\textcolor{gray}{658.3 ± 378.2(\textcolor{mygreen}{↑236.2})}\\
\midrule
\textcolor{gray}{\textbf{RAD}}&\textcolor{gray}{0}&\textcolor{gray}{100k}&\textcolor{gray}{910.4 ± 104.5}&\textcolor{gray}{508.8 ± 111.5}&\textcolor{gray}{522.1 ± 95.5}\\
\textcolor{gray}{\textbf{Longer-RAD}}&\textcolor{gray}{0}&\textcolor{gray}{200k}&\textcolor{gray}{874.6 ± 150.8(\textcolor{red}{↓-35.8})}&\textcolor{gray}{819.2 ± 115.7(\textcolor{mygreen}{↑310.4})}&\textcolor{gray}{765.5 ± 337.8(\textcolor{mygreen}{↑243.4})}\\
\midrule
\textcolor{gray}{\textbf{DrQ}}&\textcolor{gray}{0}&\textcolor{gray}{100k}&\textcolor{gray}{935.6 ± 201.3}&\textcolor{gray}{713.7 ± 147.6}&\textcolor{gray}{523.9 ± 182.2}\\
\textcolor{gray}{\textbf{Longer-DrQ}}&\textcolor{gray}{0}&\textcolor{gray}{200k}&\textcolor{gray}{906.9 ± 155.7(\textcolor{red}{↓-28.7})}&\textcolor{gray}{809.2 ± 102.0(\textcolor{mygreen}{↑95.5})}&\textcolor{gray}{740.0 ± 314.3(\textcolor{mygreen}{↑216.1})}\\
\midrule
\textbf{SAC+AE}&0&100k&799.0 ± 138.9&481.3 ± 130.4&402.6 ± 161.5\\
\textbf{Longer-AE}&0&200k&887.8 ± 127.4(\textcolor{mygreen}{↑88.8})&578.6 ± 160.7(\textcolor{mygreen}{↑97.3})&700.3 ± 232.7(\textcolor{mygreen}{↑297.7})\\
\textbf{Pretrain-AE}&100k&100k&869.8 ± 150.6(\textcolor{mygreen}{↑70.8})&757.8 ± 174.0(\textcolor{mygreen}{↑276.5})&308.0 ± 243.1(\textcolor{red}{↓-94.6})\\
\textbf{Pretrain-Random-AE}&100k&100k&793.6 ± 175.9(\textcolor{red}{↓-5.4})&858.0 ± 155.0(\textcolor{mygreen}{↑376.7})&107.8 ± 216.1(\textcolor{red}{↓-294.8})\\
\midrule
\textbf{CURL}&0&100k&717.8 ± 136.5&569.8 ± 179.4&442.6 ± 87.1\\
\textbf{Longer-CURL}&0&200k&732.1 ± 146.2(\textcolor{mygreen}{↑14.3})&688.8 ± 229.8(\textcolor{mygreen}{↑119.0})&701.5 ± 148.0(\textcolor{mygreen}{↑258.9})\\
\textbf{Pretrain-CURL}&100k&100k&785.8 ± 134.1(\textcolor{mygreen}{↑68.0})&754.5 ± 106.2(\textcolor{mygreen}{↑184.7})&277.7 ± 152.0(\textcolor{red}{↓-164.9})\\
\textbf{Pretrain-Random-CURL}&100k&100k&693.8 ± 178.2(\textcolor{red}{↓-24.0})&804.8 ± 205.8(\textcolor{mygreen}{↑235.0})&356.2 ± 131.8(\textcolor{red}{↓-86.4})\\
\midrule
\textbf{DINO}&0&100k&923.1 ± 124.4&686.2 ± 198.2&414.6 ± 162.4\\
\textbf{Longer-DINO}&0&200k&926.0 ± 128.3(\textcolor{mygreen}{↑2.9})&861.4 ± 131.8(\textcolor{mygreen}{↑175.2})&722.6 ± 251.4(\textcolor{mygreen}{↑308.0})\\
\textbf{Pretrain-DINO}&100k&100k&877.5 ± 123.8(\textcolor{red}{↓-45.6})&635.0 ± 172.0(\textcolor{red}{↓-51.2})&260.1 ± 145.8(\textcolor{red}{↓-154.5})\\
\textbf{Pretrain-Random-DINO}&100k&100k&823.4 ± 75.4(\textcolor{red}{↓-99.7})&712.6 ± 126.6(\textcolor{mygreen}{↑26.4})&197.5 ± 147.2(\textcolor{red}{↓-217.1})\\
\midrule
\textbf{ELo-SAC}&0&100k&789.3 ± 198.2&478.3 ± 159.9&537.5 ± 164.5\\
\textbf{Longer-ELo-SAC}&0&200k&919.2 ± 154.1(\textcolor{mygreen}{↑129.9})&753.8 ± 159.9(\textcolor{mygreen}{↑275.5})&789.9 ± 335.3(\textcolor{mygreen}{↑252.4})\\
\textbf{Pretrain-ELo-SAC}&100k&100k&711.5 ± 161.7(\textcolor{red}{↓-77.8})&503.6 ± 220.1(\textcolor{mygreen}{↑25.3})&115.5 ± 152.4(\textcolor{red}{↓-422.0})\\
\textbf{Pretrain-Random-ELo-SAC}&100k&100k&742.3 ± 142.4(\textcolor{red}{↓-47.0})&548.0 ± 136.4(\textcolor{mygreen}{↑69.7})&179.7 ± 189.5(\textcolor{red}{↓-357.8})\\
\bottomrule
\end{tabular}

}

\end{table}

In general, given the same amount of model updates, the pretraining framework performs better than the joint learning framework except ELo-SAC (we believe this is because ELo-SAC search was done only under the joint learning framework).
But such an advantage of the pretraining framework may come from the extra data used in the pretraining stage.
When the same amount of data is given, the longer joint-learning configuration usually performs better than the pretraining methods except when AutoEncoder is the self-supervised loss.
Such observations imply that the learning framework has different impacts on policy learning even if the same self-supervised loss is applied.
It might not be the best practice to directly use the existing self-supervised losses designed for joint learning framework with the pretraining framework.
However, only Longer-ELo-SAC achieves comparable results compared to image augmentation based methods with `Longer' configuration.
On the other hand, we argue that on DMControl, the advantages of the pretraining framework come from the access to extra data instead of the framework itself.
When the total environment step is limited and no previous data has been collected, the joint learning framework can better solve DMControl problems.


\subsection{Limitations}
\label{sec:limitation}
In this paper, we focus on SAC for environments with continuous action space and Rainbow for environments with discrete action space. 
Though both methods are generic, it will be interesting to see how self-supervised losses work with other RL methods and image augmentations in more challenging environments.
Meanwhile, both RAD~\cite{laskin2020reinforcement} and DrQ~\cite{yarats2020image} investigate many image augmentation approaches for their learning methods.
We only focus on random crop and translate because of their positive effects, and more combinations of image augmentation and self-supervised learning methods worth further investigation.
In addition, the search space of ELo-based methods are relatively limited.
They may achieve better scores with a larger search space (more losses) and a more representative searching environment.

\subsection{Computation Information}
\label{sec:computation}
Training one DMControl agent for 50k model update steps usually takes 3 hours on one NVIDIA A5000 GPU.
It takes around 1.5 hours to train five Atari agents in parallel using an Apple M1 Max CPU.


\end{document}